\documentclass{article}
\usepackage{frExamplee}
\usepackage{graphicx}
\usepackage{apalike}
\usepackage{setspace}
\usepackage[ruled,linesnumbered]{algorithm2e}
\usepackage{multirow}
\usepackage{makecell}

\usepackage{algpseudocode}
\makeatletter
\newcommand{\removelatexerror}{\let\@latex@error\@gobble}
\makeatother
\algnewcommand\algorithmicforeach{\textbf{for each}}
\algdef{S}[FOR]{ForEach}[1]{\algorithmicforeach\ #1\ \algorithmicdo}
\usepackage{amsmath}
\usepackage{booktabs}
\usepackage{cite}

\usepackage{threeparttable}
\usepackage{color}

  \usepackage[caption=false,font=footnotesize]{subfig}

\begin{document}


\title{LiDAR Road-Atlas: An Efficient Map Representation for General 3D Urban Environment}


\author{
	Banghe Wu\thanks{ School of Computer Science and Engineering, Nanjing University of Science and Technology, Nanjing, Jiangsu, China (e-mail: banghe\_wu@njust.edu.cn).}\And
	Chengzhong Xu\thanks{ The State Key Laboratory of Internet of Things for Smart City (SKL-IOTSC), Department of Computer Science, University of Macau, Macau (e-mail: czxu@um.edu.mo).}\And
	Hui Kong\thanks{ The State Key Laboratory of Internet of Things for Smart City (SKL-IOTSC), Department of Electromechanical Engineering (EME), University of Macau, Macau (e-mail: huikong@um.edu.mo). Corresponding author.}
}


\maketitle

\begin{abstract}

In this work, we propose the LiDAR Road-Atlas, a compactable and efficient 3D map representation, for autonomous robot or vehicle navigation in general urban environment. The LiDAR Road-Atlas can be generated by an online mapping framework which incrementally merges local 2D occupancy grid maps (2D-OGM). Specifically, the contributions of our method are threefold. First, we solve the challenging problem of creating local 2D-OGM in non-structured urban scenes based on a real-time delimitation of traversable and curb regions in LiDAR point cloud. Second, we achieve accurate 3D mapping in multiple-layer urban road scenarios by a probabilistic fusion scheme. Third, we achieve very efficient 3D map representation of general environment thanks to the automatic local-OGM induced traversable-region labeling
and a sparse probabilistic local point-cloud encoding. 
Given the LiDAR Road-Atlas, one can achieve accurate vehicle localization, path planning and some other tasks. Our map representation is insensitive to dynamic objects which can be filtered out in the resulting map based on a probabilistic fusion. Empirically, we compare our map representation with a couple of popular map representations in robotics society, and our map representation is more favorable in terms of efficiency, scalability and compactness. Additionally, we also evaluate localization performance given the LiDAR Road-Atlas representations on two public datasets. With a 16-channel LiDAR sensor, our method achieves an average global localization errors of 0.26m (translation) and 1.07 degrees (rotation) on the Apollo dataset, and 0.89m (translation) and 1.29 degrees (rotation) on the MulRan dataset, respectively, at 10Hz, which validates its promising performance. 


\end{abstract}
  


\section{Introduction}

Now
days, 
autonomous online 3D mapping plays a key role on robot exploration of unknown environment, and can also acts as the "teaching" role in most ``teach-and-repeat'' paradigms for robot navigation or autonomous driving applications. As exploration goes on, the map size (either runtime memory usage or hard-drive storage) can scale up quickly. Likewise, 3D mapping of large general urban environment is also important for commercialized autonomous vehicle, where the resulting 3D high-definition (HD) map can be re-used for vehicle localization and path planning. When applied at large scale, the map's storage size is one key factor that has to be considered as well.
Therefore, a parsimonious map representation is highly desirable. 
In this paper, we propose the LiDAR Road-Atlas, a compactable and efficient map representation of general 3D urban environment, for autonomous robot or vehicle navigation. It will be generated by an online mapping framework  based on incrementally merging 2D local occupancy grid maps.

\begin{figure}
\centering
\includegraphics[width=0.95\linewidth]{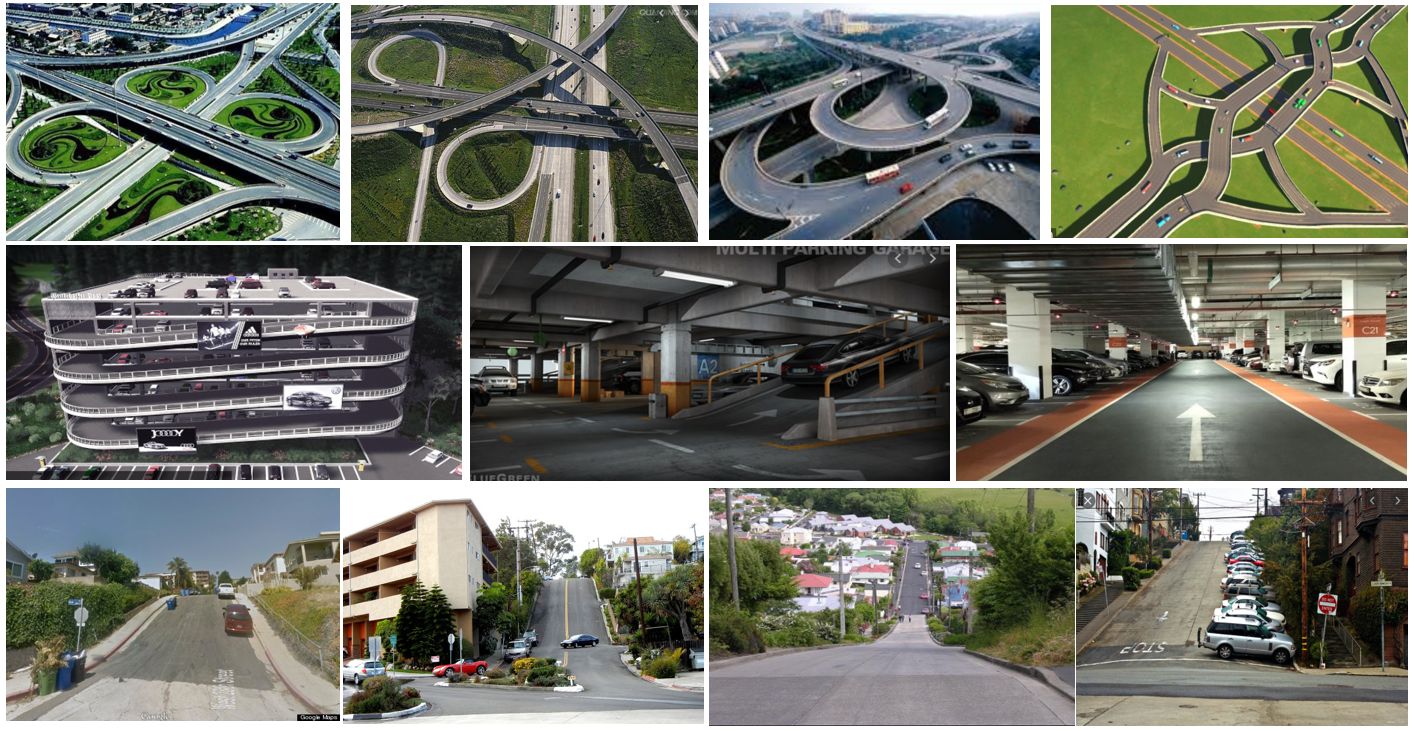}
\caption{Examples of nonflat urban infrastructures including underpass, overpass, multiple-layer garage, and steep road scenes etc.}
\label{fig:nonflat-structured-road}
\end{figure}

In the robotics society, one classical and efficient way to represent the environment is the 2D occupancy grid map (2D-OGM) \cite{Elfes:ogm} where the cells in the grid are labeled as drivable, unknown or occupied. A 2D occupancy map usually suffices for robot navigation in structured outdoor environment. For navigation in non-structured outdoor  environments, where ground is not purely flat (Fig.\ref{fig:nonflat-structured-road}) or where overhanging objects (e.g. tree branches, tunnels,
low roofs, etc.) are very common, a 2D representation of the environment is insufficient because 2D-OGM cannot fully represent the actual environment. In such outdoor scenarios, 2D-OGM cannot be accurately created in general because it is not easy to reliably find the boundaries between known (traversable region) and unknown (occluded region) areas due to the existence of low obstacle regions.
Using an underrepresented map can be totally erroneous in path planning, let alone vehicle localization.
Therefore, a more suitable alternative
for outdoor navigation would be a map representation where the environment could be well described. Over the years, a few more expressive 3D map representation methods than 2D-OGM have been proposed for autonomous vehicle navigation \cite{Triebel:IROS06, Souza:ICAR13, Souza:Robotica16, Pfaff:IJRR07, Steinbrucker:ICRA14,Hornung:AutoRobot13,Yue:SMC16,Funk:RAL21}, and these methods can be broadly categorized as elevation-maps or full 3D point-cloud maps. The map's scale in robotics society is not necessarily very large considering that robots only operate in an area of relatively limited size.

In the autonomous driving society, unlike the map representations in robotics field,  HD-maps can be extremely large, which can be up to a city-scale. Therefore, high efficient and compactable map representations are desirable for autonomous vehicle navigation. The recent road-based mapping methods for autonomous driving in urban environment \cite{Schreiber:IV13, Ranganathan:IROS13, Lu:IV17, Rehder:IV15, Jeong:IV17, Qin:IROS20, Herb:IROS19, Qin:2021} belong to this line of research, where varieties of semantic 2D and 3D features are segmented from road region. The 2D elements are basically on road surface, including lane markers and curb-lines, zebra-crossings, and all kinds of ground arrow signs etc. The 3D structures are mostly regular vertical objects, such as traffic lights, traffic signs and light poles etc. Generally, deep learning methods are used to segment these semantic image regions. Then, high-precision GPS/IMU and multi-ray LiDAR sensors are combined to build a prior urban map. Thereafter, the locations of the extracted semantic-feature in images are transformed into the world coordinates and anchored to the created prior map. 
Given the created HD map, one can apply it to accurately localize a vehicle with a single camera (aided by a low-cost GPS receiver and wheel-odometers for coarse localization in general), where the image from the camera is segmented by a deep-learning model to find the road-scene semantic features, and thereafter a fine localization is achieved by feature association and PnP (perspective-n-point) algorithms. 

By comparing the above two types of HD-maps in robotics domain and autonomous-driving industries, respectively, we find that 
neither of them can well meet the requirement for creating HD-maps of general urban environment, where semantic road markers or regular man-made road-scene infrastructures (e.g., traffic light poles) might be absent, and there might also exist multiple-level road surface (e.g., underpass and multiple overpass). In robotics field, 
the elevation-based methods cannot fully represent complex scenes and the full 3D point-cloud based maps generally consume too much memory or storage space.In autonomous driving society, the existing road-based mapping methods heavily rely on semantic road features. Although these features are salient and common in high-way road scenes, they are not necessarily present universally in general urban environment, e.g., various residential communities, university campuses or industrial parks etc. Thus, one has to collect new images of semantic road-scene features which are specific to each individual scene before learning a new semantic segmentation deep learning model, for which it is a heavy workload.
In addition, these methods cannot guarantee to build maps in an online mode, and some of them also need expensive high-precision GPS/IMU  sensors. 

Given the problems of the above map representations, it is necessary to think of an efficient map which is parsimonious in memory consumption and hard-drive storage, and highly descriptive for complex environment as well. 
Summing up all the above thoughts, the main contributions of our mapping work can be as follows. (1) We propose an efficient online mapping method which can be applied to general 3D urban environments to create an HD-map for autonomous vehicle navigation. The result of mapping is a parsimonious map representation by embedding
sparse local point-cloud encodings and local-OGM induced traversable-region labeling. (2) We can handle the challenging problem of creating local 2D-OGM in non-structured urban scenes based on a real-time delimitation of traversable and curb regions in LiDAR point cloud.
(3) We can achieve accurate 3D mapping in multiple-layer urban road scenarios and solve 
the map update problem of the existing multiple-level elevation-maps caused by dynamic objects present in the mapping process 
by a probabilistic fusion scheme.

We have compared the proposed map representation with the popular OctoMap map representation \cite{Hornung:AutoRobot13}. Empirically,  the resulting map by the proposed map representation is much smaller than that of the OctoMap representation. Based on the runtime memory coverage, our map representation can afford the online large-scale map creation while the OctoMap cannot.
We have also conducted extensive experiment on online map update and map-based localization, by which we have shown the accuracy and efficiency of the proposed map representation.


In contrast to the existing mapping methods, our method belongs to the more broad category of traversable-region based mapping algorithms, where the traversable region is not limited to structured urban road where the man-made semantic road elements (road markings/signs, and traffic lights etc.) are available. 
Actually, our method can be applied to more general natural scenes instead of just urban environment. Our method only uses a multiple-channel LiDAR sensor (VLP-16) to generate the LiDAR Road-Atlas. Noted that we do not use cameras in this work although we occasionally show images in the paper, and the purpose is to tell what the evaluation scenarios look like. 
Because our mapping is an online process, it can also be easily extended to some robotic exploratory tasks in an unknown environment.

\begin{figure*}
\centering
\includegraphics[width=0.95\linewidth]{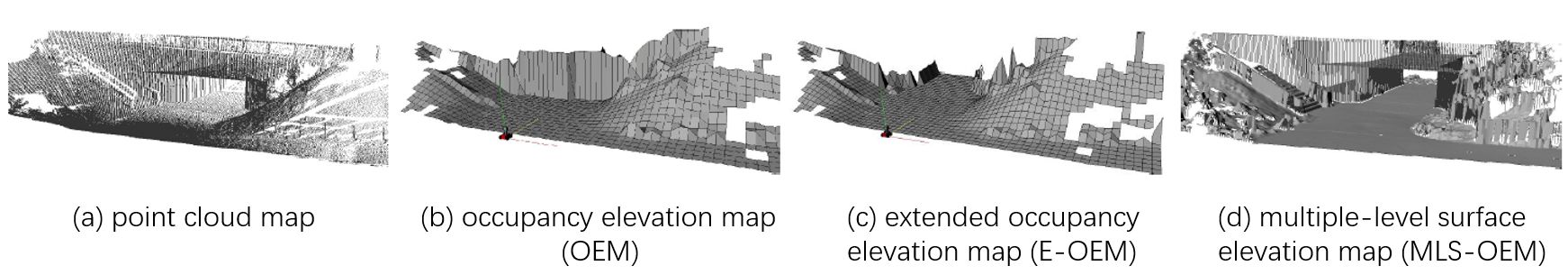}
\caption{Four types of 3D occupancy grid map representations (courtesy of Triebel et al. \cite{Triebel:IROS06}). Refer to our introduction part for more details of these methods.}
\label{fig:OEM}
\end{figure*}

\section{Related works}

\textbf{Representation of traversable region -} 
In representing traversable regions, some researches \cite{Rehder:IV15, Jeong:IV17, Qin:IROS20, Herb:IROS19, Qin:2021} focused on the process of building
road maps. Regder et al. \cite{Rehder:IV15} detected lanes in image and exploit odometry information to generate local grid maps, and pose between local grid maps is further optimized and stitched.
Jeong et al. \cite{Jeong:IV17} first differentiated road markings, and more informative class information can be used to avoid ambiguity on matching. Pose graph
optimization based on loop closure is performed to eliminate drift for
global pose consistency. Qin et al. \cite{Qin:IROS20} utilized road markers to build a semantic map for underground parking lots with application to autonomous parking based on only cameras.
Herb et al. \cite{Herb:IROS19} proposed a crowd-sourced solution to 
build a semantic map. However, it was difficult to be
applied because its computational complexity is generally high in matching inter-session features. Very recently, Qin et al. \cite{Qin:2021} proposed a light-weight semantic map representation for assisting camera-based vehicle localization in high-way structured environment. 
In utilizing the created road HD-map, Schreiber et al. \cite{Schreiber:IV13} proposed to  detect curbs and lanes, and then match the structure of these elements with an HD-map. Ranganathan et al.
\cite{Ranganathan:IROS13} utilized detected corners on road markers for localization. Lu
et al. \cite{Lu:IV17} formulated a non-linear optimization to match
road markers with the map. The vehicle odometry
and epipolar geometry were also taken into consideration
to estimate the 6-DoF camera pose.
Morales et al. \cite{Morales:JFR09} proposed an autonomous system for robot
navigation in outdoor cluttered pedestrian walkways. As a part of this system, a module of driving path detection is given for safe navigation of robot. Different from our method, this method applied a cascaded height filter to find traversable region in each individual scan, and then detected curb edge point to construct a map for navigation. The navigation map is still two-dimensional because the planned way-points lie in a planar map.

\textbf{Representation of 3D space occupancy -} Occupancy elevation map (OEM) is also a frequently used map representation method to deal with the problem of under-representing 3D environment in 2D-OGM 
\cite{Bares:elm, Triebel:IROS06, Hebert:elm, Souza:ICAR13, Souza:Robotica16}, as shown in Fig.\ref{fig:OEM}.
In OEM, the height of the surface is stored in the corresponding cell of the discretized grid. While the OEM provides an efficient map representation, it lacks the
ability to fully represent vertical structures or even multiple levels of horizontal planes (e.g., underpass/overpass bridges in Fig.\ref{fig:OEM}(b)).
The extended occupancy elevation map (E-OEM)\cite{Pfaff:elmEX} can alleviate the problem of OEM to some extent, and allows to handle overhanging
objects (e.g., tree branches) or represent the underpass. But it cannot fully represent the environments where there are more than one level of horizontal surface (e.g., the overpass bridges in Fig.\ref{fig:OEM}(c) or multiple-level parking garages). 
To address this issue, Triebel et al. propose multi-level surface occupancy elevation map (MLS-OEM) data structures \cite{Triebel:IROS06}, where a list of surfaces are stored in each cell of the discrete grid (e.g., Fig.\ref{fig:OEM}(d)). Additionally,
they use intervals to represent vertical structures. The MLS-OEM roughly provides a 3D representation without the complexity of a full 3D point-cloud map, enabling a mobile robot to model complex environments. 
However, the MLS-OEM representation is designed for mapping static environment, and it is not capable of handling the influence of dynamic objects during the mapping process. In constrast, our method can not only represent the multiple-level surface elevation but also deal with the influence of dynamic objects in the mapping process. 

Another widely used map representation is the OctoMap \cite{Hornung:AutoRobot13}, where each point can be stored and indexed in an Octree data structure. The indexing of each point is fast due to the tree structure. 
The OctoMap map is able to model almost any kinds of environments without prior assumptions on it, and it models occupied areas as well as free space. Unknown areas of the environment can also be implicitly encoded in the map. Although OctoMap has high representative capability, its memory usage and hard-drive storage is also quite high. 
The normal distribution transform traversability mapping (NDT-TM) representation \cite{Ahtiainen:JFR17} for outdoor environments is able to classify sparsely vegetated areas as traversable without compromising accuracy on other terrain types. The NDT-TM representation is proposed by exploiting 3D LIDAR sensor data to incrementally expand normal distributions
transform occupancy (NDT-OM) maps \cite{saarinen20133d}, which is a 3D spatial model based
on a regular grid that concurrently estimates both the
occupancy and the shape distribution in each cell. In addition to geometrical information, the NDT-OM representation is augmented with statistical data of the permeability and reflectivity of each cell of NDT-OM map. Using these
additional features, a support-vector machine classifier is trained to discriminate between traversable and nondrivable
areas of the NDT-TM maps.
Despite of advantages over 2D grid mapping, the construction of full 3D occupancy grid models \cite{Hornung:AutoRobot13,Ahtiainen:JFR17,saarinen20133d} typically has excessively high demands for run-time memory and hard-drive storage during its online application to a mobile robot in large-scale outdoor scenarios or when fine map resolution is necessary.


\begin{figure}[h]
    \centering
    \includegraphics[width=0.7\textwidth]{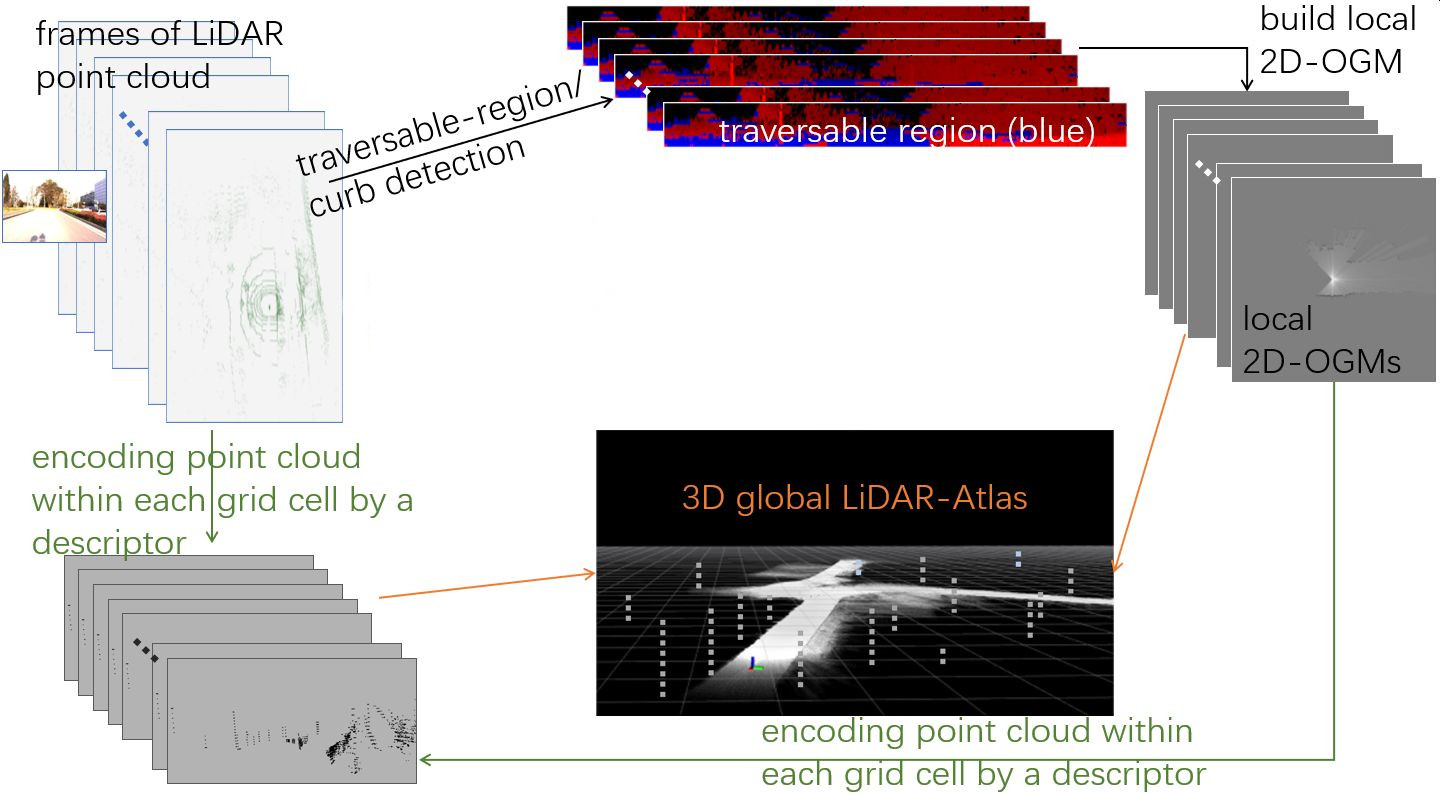}
    \caption{Overview of the process generating 3D global LiDAR-Atlas. Basically, it consists of three major components to extract intermediate results that are aggregated to form the 3D LiDAR-Atlas, i.e., travserable-region (curb) detection for building local 2D-OGM, encoding of the point-cloud's geometric information within each cell of the local 2D-OGM.
    }
    \label{fig:overview}
\end{figure}

The remainder of the paper is organized as follows. We introduce our method in Section III, and the implementation details and experimental results are given in Section IV. We discuss our method in terms of its scalable, extendable, and applicable properties in Section V. The conclusions are drawn in the last section. Four videos which demonstrate the creation of LiDAR Road-Atlas for four different scenes are available at https://youtu.be/9wXiZzGau3w, https://youtu.be/wjkf8snJzRA,  
https://youtu.be/4tfKbb47Ajc and https://youtu.be/UMZLE8CMGtk.

\section{Our method}
Our work aims at building a 3D global sparse occupancy grid map with labeled traversable regions and static obstacles (e.g., trees and buildings) codings embedded inside. 
To do that, we estimate the LiDAR's 6D pose based on a LiDAR-based SLAM or odometry method. At each LiDAR's key-frame location, we build a local occupancy grid map and align it to the estimated 6D pose (Fig.\ref{fig:local-occ-grid}). A stitching operation is given to fuse two overlapping local occupancy grid maps when necessary. We consider not only common outdoor scenes but also the scenarios where it is necessary to build multiple layers of local occupancy grid map, e.g., the crossovers and multiple-layer parking lots shown in Fig.\ref{fig:nonflat-structured-road}. 

Our method takes in point cloud data from a 16-channel LiDAR sensor and estimates relative pose between two consecutive frames. An overview of our method is shown in Figure \ref{fig:overview}, where the point cloud is processed and automatically labeled, with each point being classified as traversable regions or obstacles. Thereafter, the processed point cloud is used to construct a local grid map. Meanwhile, for each frame of point cloud, spare descriptors of local geometric structures are extracted and embeded into the global 3D LiDAR Road-Atlas along with the labeled traversable region in each local grid map. 

\subsection{Detecting traversable area and obstacles in each LiDAR point-cloud keyframe}\label{section3.1}
Accurate detection of traversable regions is crucial for path-planning and route-following in the ``repeat'' stage of the ``teach-and-repeat'' navigation paradigm. Especially when localization in the map is not robust enough, a reliable traversable road extraction method can be able to keep the robot stay and run in the path. Traversable region labelling is strongly relevant to detection of obstacles, ranging from nontrivial ones such as vehicles, walls, overhanging tree branches and overpass bridges to trivial ones such as curbs or ditches. 

Our algorithm is designed to automatically detect traversable region during map building process. The online detection is also valuable to planning paths for robotics exploration task. 
In general, one usually uses the Random Sampling and Consensus  (RANSAC) algorithm \cite{ransac} to find a dominant plane of the scene. However, there usually exist multiple planes of similar size in a scene. When applying a 
single RANSAC, small obstacles (e.g., curbs or pits on road) are usually overlooked if a large tolerance threshold is adopted.
As a result, it is neither reasonable nor robust to apply a single RANSAC model to detect traversable regions. 

Although a single RANSAC does not work well in a whole point-cloud frame, it can do a good job locally. Therefore, one could
apply multiple RANSAC models sequentially to the whole point cloud, where a partial traversable region is detected and removed from the point cloud by a different RANSAC model in an iterative way. Although this multiple-RANSAC way could generally get better results than a global RANSAC model, it is not trivial to determine the exact number of iterations necessary for a good segmentation. 

\begin{figure}
    \centering
    \includegraphics[width=0.7\textwidth]{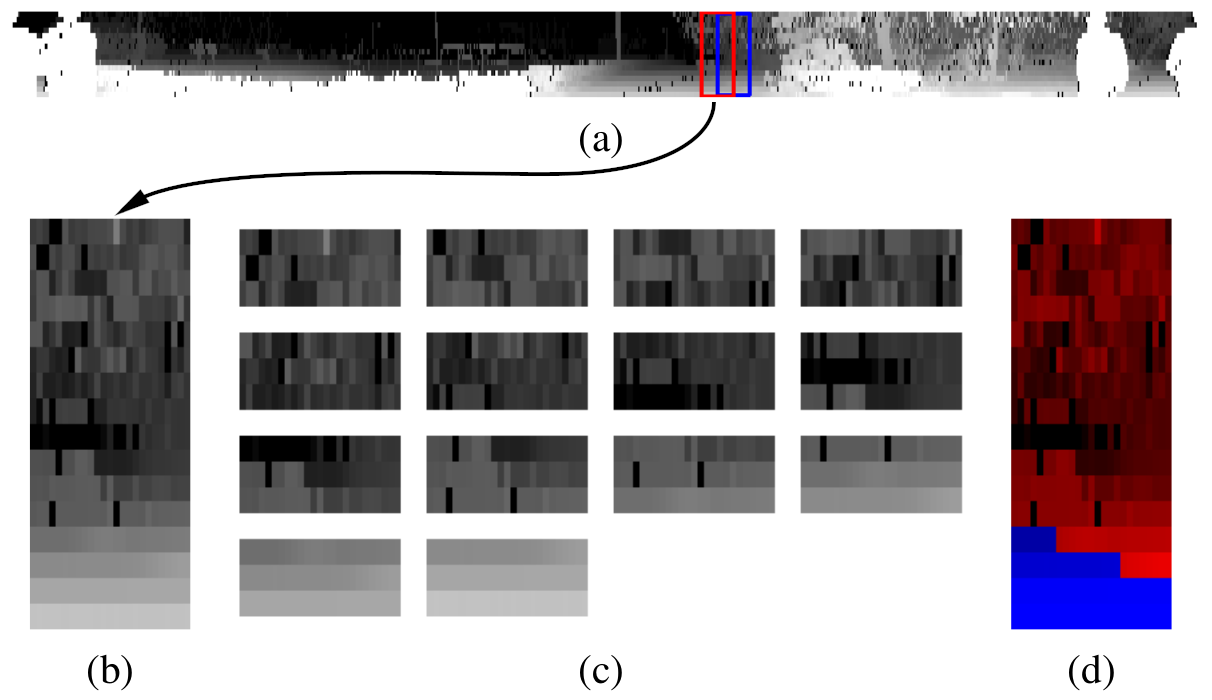}
    \caption{Splitting a whole LiDAR frame into many overlapping regions. (a) One VLP-16 LiDAR image (16x1800) is shown. Note that we have adjusted the scale of LiDAR image for a better view. The LiDAR image is splitted into many overlapping local regions horizontally and vertically, where two neighboring horizontal overlapping sectors (red and blue boxes) have a size of 16x50 and a step size of 25. Likewise, let us forget about the disproportionate ratio of the box's dimension.
    (b) the enlarged red box. (c) All overlapping local areas extracted from (b) have a step size of 1, and each local area has a size of 3x50. (d) the detected travserable region (blue) and obstacle area (red) based on the proposed method.
    }
    \label{fig:area_extract}
\end{figure}

We propose a new multiple-RANSAC based method to detect traversable and obstacle region from a frame of point cloud. Specifically,
we divide 
the point cloud $P$ into overlapping local sectors $A_i$, and perform multiple RANSAC models on these sectors, respectively, to separate traversable areas from obstacle regions. Note that the smallest obstacles by our method can range from curbs to small ditches or stones on road.
Figure \ref{fig:area_extract} illustrates the way of splitting a whole LiDAR point-cloud frame into many overlapping regions. Based on the splitting, the multiple-RANSAC traversable region and obstacle detection method is given in the Algorithm \ref{algo:1}.

Since RANSAC cannot distinguish between the ground and the wall, we define $check\_normal$ to check the normal vector of the plane obtained by RANSAC to remove too steep planes. The return value of $check\_normal$ is only true when the angle between the normal vector and the z-axis is less than the threshold. The threshold depends on the vehicle. We set it to $0.4 rad$ in our experiments.

\begin{figure}[!h]
    \removelatexerror
    \begin{algorithm}[H]
        \caption{Detecting traversable area and obstacles}
        \label{algo:1}
        \LinesNumbered
        \KwIn{whole point cloud $P$, local point cloud $A_i$}
        \KwOut{traversable-region $G$, obstacles $O$}
        \ForEach{$A_i$}{
            $G_{local}$ = RANSAC($A_i$)
            \algorithmiccomment{$N$ is the normal vector of the plane} \\
            \If{size($G_{local}$) $\geq$ size($A_i$)$/2$ $\&\&$ check\_normal($N$)}{
                $G$ = $G \cup G_{local}$ \\
            }
        }
        $O$ = $P - G$ \\
    \end{algorithm}
\end{figure}

\begin{figure}[h]
    \centering
    \subfloat[]{ \includegraphics[width=0.32\textwidth]{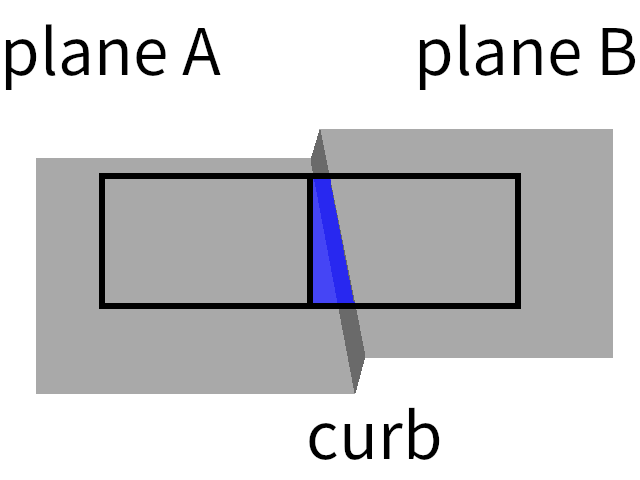} \label{area_b}}
    \subfloat[]{ \includegraphics[width=0.32\textwidth]{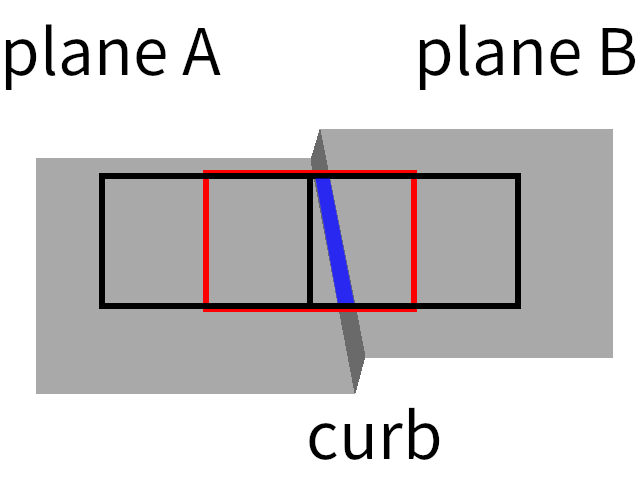} \label{area_c}}
    \caption{A demonstration of the role of our ``'splitting with overlap'' scheme on segmentation of a curb-like region. (a)Two split regions without ``overlap'' between them. (b) Three split regions with ``overlap'' between two neighboring local regions. Based on Algorithm \ref{algo:1}, the total traversable region in a point-cloud frame is composed of many smaller traversable planes which are detected in the local regions. The rest corresponds to obstacle area. In (a), part of road plane (blue region) is detected as the curb obstacle. In (b), we can deal with this issue.}
    \label{fig:area_overlap}
\end{figure}

Generally, there are two possibilities in fitting a plane in a split local region. One is that the whole local region can be perfectly fitted by a plane model, e.g., the left dark rectangle shown in Fig.\ref{fig:area_overlap}(a)(b). 
The other is that partial of the split local area can be  fitted by a plane model, e.g., the red rectangle and the right dark rectangle shown in Fig.\ref{fig:area_overlap}(b). For such local regions, the plane is detected as the part which covers more than 50\% of the whole local region. In Fig.\ref{fig:area_overlap}(a), we show an alternative local-region splitting scheme where there is no overlap between two neighboring local region, for which the segmentation of small obstacle is worse than that in the splitting case with overlap.
Therefore, the local-region splitting way with overlap is our choice. It deserves pointing out that the splitting with overlap is crucial for reliably detecting small objects such as curbs, which can guarantee creating an accurate local occupancy grid map, as shown in Fig.\ref{fig:local_grid}.


In our implementation, we split a whole point cloud frame into local areas using the LiDAR image representation \cite{Chen:ICRA17}.
The raw output data of a multiple-channel LiDAR are actually based on spherical coordinate system,
including the yaw angle $\theta$ of each laser beam and measurement distance $d$. The pitch angle $\phi$ of each laser beam can be loaded from the calibration file. Actually, there are a couple of ways to divide a point cloud to meet the overlap condition. However, the raw point cloud representation is sparse, and it is not straightforward to directly perform neighborhood calculation and extract the neighborhood that meets the overlap condition. That is why we choose the LiDAR image representation in traversable region detection. 


\subsection{Creating 2D local occupancy grid map from each LiDAR point-cloud keyframe}\label{section3.2}
Conventionally, the 2D-OGM is obtained by a 2D LiDAR sensor. It can produce a clear boundary representing dynamic or static obstacles in indoor environment or 2D structured outdoor environment, and thus a 2D-OGM can well represent the environment. In such environment, vehicle localization and path planning can be fulfilled with ease in a 2D-OGM generally. 
In a more general 3D urban environment, 2D-OGM cannot be reliably created if obstacles are not well labeled. In our work, we can deal with this issue by the traversable and curb region labeling procedure as described in the last section. Meantime, we can also take the advantage of 2D-OGM over its 3D counterpart, i.e., efficiency and small memory usage. 

\begin{figure}[h]
    \centering
  \includegraphics[width=0.7\textwidth]{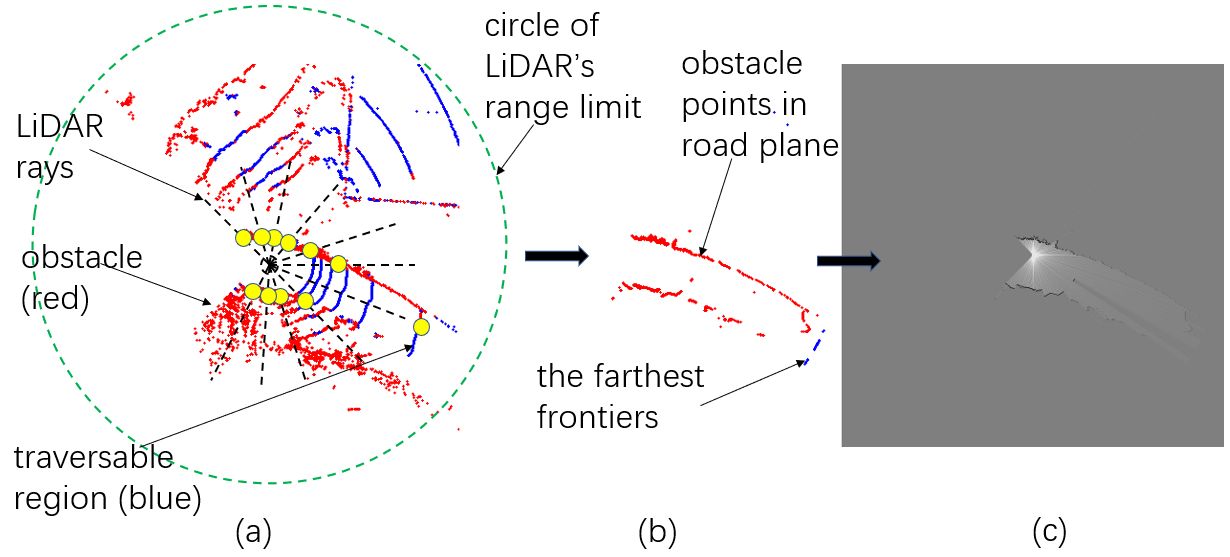}
    \caption{(a) Projecting the automatically labeled 3D points (based on the method proposed in the last section) onto
the 2D ground plane, and applying ray-tracing to find the points corresponding to small obstacles (curb points). (b) the obtained obstacle points (red) corresponding to curbs and farthest frontier points (blue). (c) the obtained 2D occupancy grid map by using the points in (b) based on the method in \cite{Thrun:05,Hess:ICRA16}.}
    \label{fig:local_grid}
\end{figure}

Specifically, we assume that the local ground area around the vehicle can be approximated as a 2D plane. After removing points with significant height, 
each frame of point cloud is projected vertically onto this 2D plane. Figure \ref{fig:local_grid} (a) shows such a projection. However, the resulting points are too dirty to apply the 2D occupancy grid mapping algorithm \cite{Thrun:05,Hess:ICRA16} to construct a local occupancy grid effectively. In addition, directly using the projected points as the result of 2D-OGM will be very sparse when the resolution is small and cannot be filled effectively, which is unfavorable for the subsequent application of the map (such as active exploration).
Therefore, we hope to convert all the projected 2D points to obtain a virtual scan which is more convenient for generating a 2D-OGM before applying the method in \cite{Thrun:05,Hess:ICRA16}. To do that, we adopt the ray-tracing technique to find the closest obstacle point along each ray direction, e.g., the yellow points in Fig.\ref{fig:local_grid}(a). In addition, if there is no 
obstacle point along a certain ray, the farthest point, which has been labeled as traverable (e.g., the blue points in Fig.\ref{fig:local_grid}(b)), is used to speed up the filling of the map to differentiate known area from unknown one. Figure \ref{fig:local_grid} shows a created 2D-OGM. In our implementation, if the resolution of our 2D-OGM is set to 0.1m/cell-width and the size of the local region that LiDAR can sense is 40mx40m, the size of our 2D-OGM is 400 pixels by 400 pixels. 

\begin{figure}[h]
    \centering
    \includegraphics[width=0.7\textwidth]{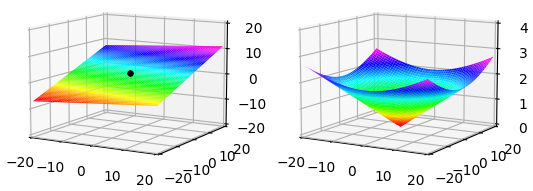}
    \caption{An example of the distribution of the height of a local grid. The left image shows the mean values $\mu$ at each cell location, and the black point in the center corresponds to the location of the robot. The right image shows the standard deviation $\sigma$ at each cell location, where $\sigma = \tan5^{\circ}\cdot d+0.1$.}
    \label{fig:guassian}
\end{figure}

Due to measurement noise, it can be assumed that the altitude of each local grid of the segmented traversable region in each LiDAR frame is subject to a Gaussian distribution.
Specifically, the mean of each Gaussian distribution is the height of the local grid that is represented in the global coordinate system.
In general, the standard deviation of each Gaussian distribution increases with the distance, $d$, to the LiDAR sensor. 
For efficiency purpose, we simply set the standard deviation as a combination of two parts, namely $\sigma = \alpha_d + \beta$. The $ \alpha_d$ is a variable standard deviation which becomes larger with the increased distance $d$, and $\beta$ is a constant used to ensure that the standard deviation of the cell under the robot is nonzero.
Figure \ref{fig:guassian} shows an example of the distribution.

\subsection{Building 3D global occupancy grid map}
\begin{figure}
\centering
\includegraphics[width=0.8\linewidth]{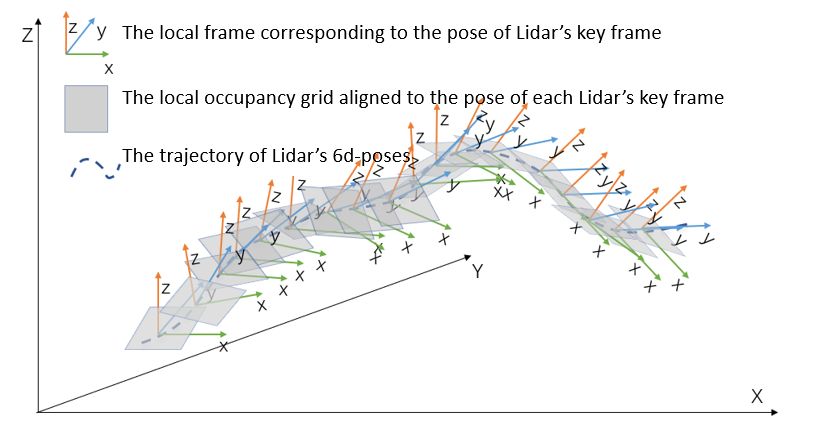}
\caption{Alignment of local occupancy grid maps to the estimated 6D pose at each key-frame location of the pose trajectory. }
\label{fig:local-occ-grid}
\end{figure}

With a LiDAR-based odometry method, we can obtain the pose trajectory of the vehicle. For each key-frame pose along the trajectory, we estimate a local 2D-OGM and make it aligned with the pose represented in the global coordinate system. To construct a 3D global occupancy grid, our solution is to merge all local occupancy grid maps by stitching the neighboring ones so that the transition between two local 2D-OGM is as smooth as possible, as shown in Fig.\ref{fig:local-occ-grid}. 

\begin{figure}[t]
    \centering
    \subfloat[]{ \includegraphics[width=0.48\textwidth]{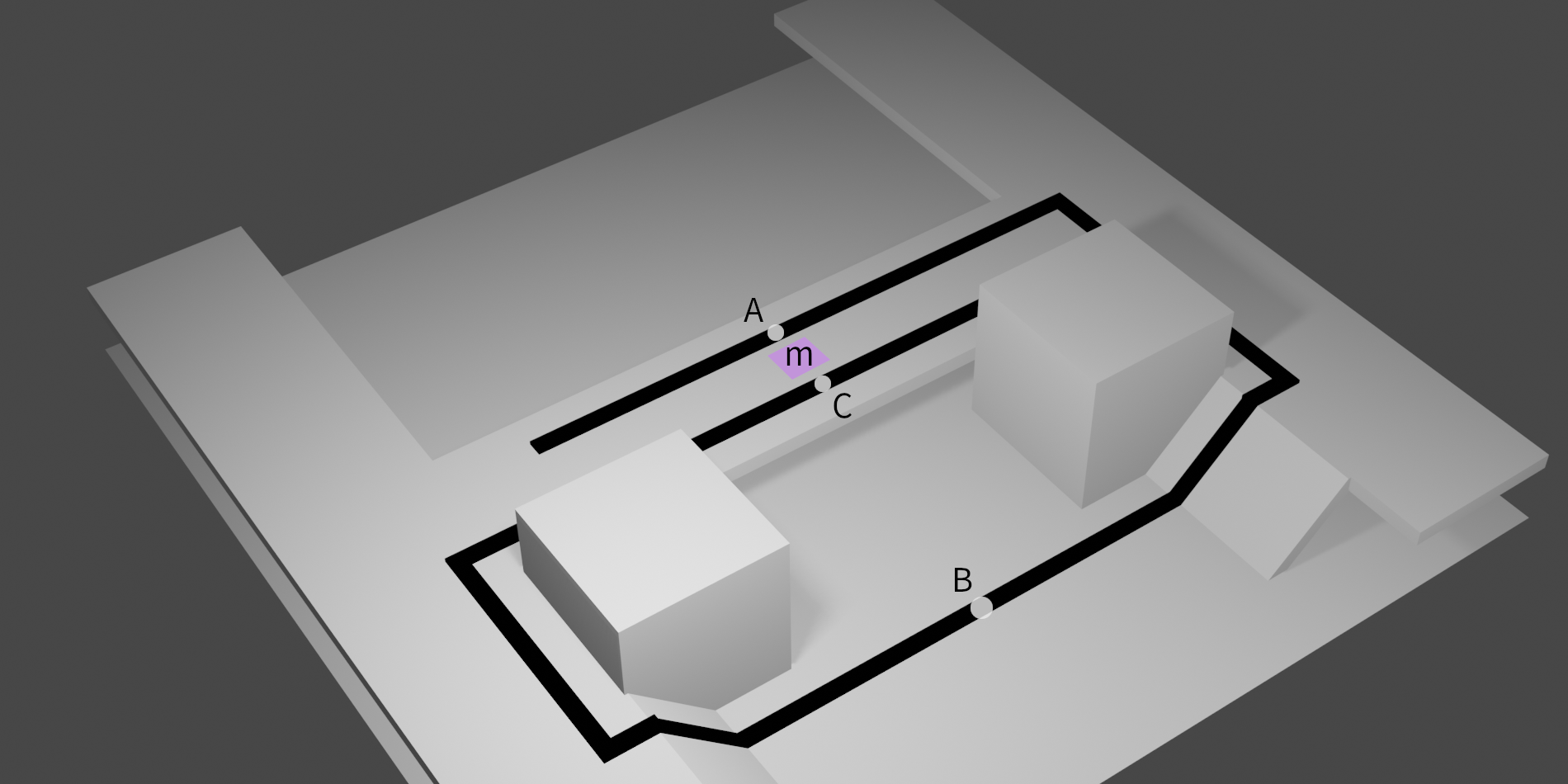}}
    \subfloat[]{ \includegraphics[width=0.24\textwidth]{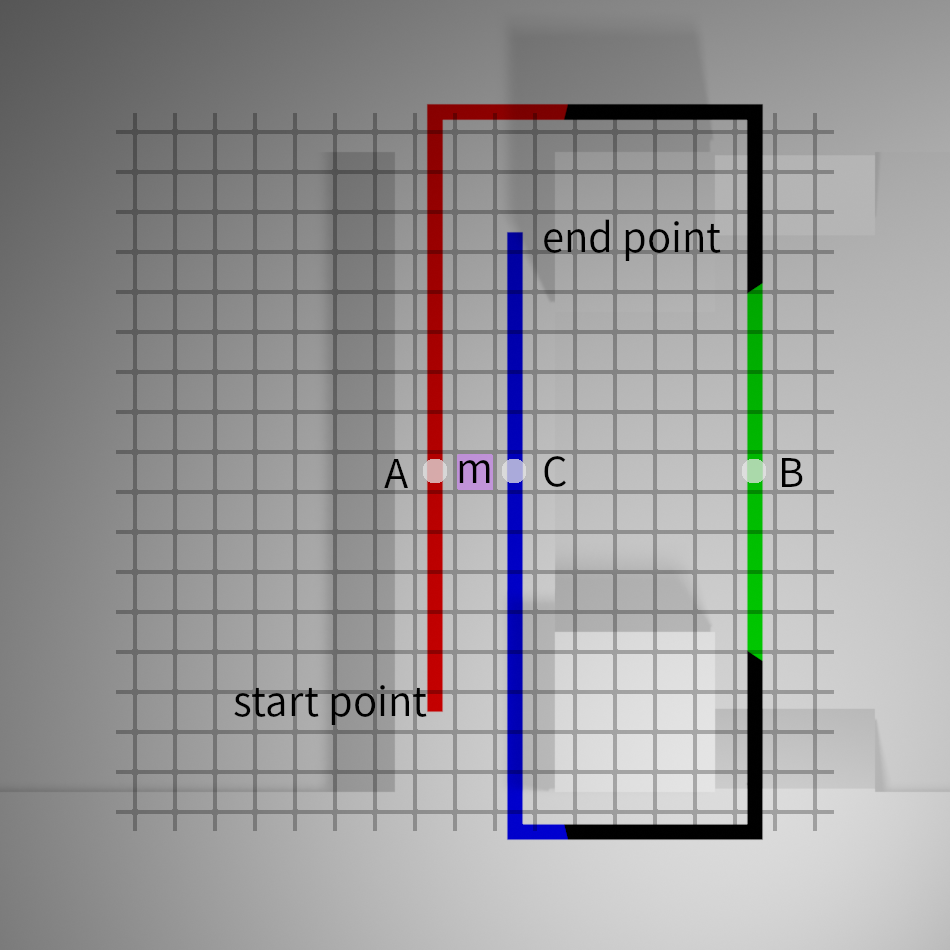}}
    \caption{A conceptual illustration of fusing multiple observations made at a  multiple-level local cell. (a) A typical scene and vehicle's trajectory, 
    where $m$ is a multi-layer cell in the global map. 
    (b) 
    Top view. Each little grid is a cell of the global map. 
    According to the $OLR$ criterion, the representative observations of the red and blue segments (made at $A$ and $C$) are redundant and are fused into the same layer. The representative observation of the green trajectory (made at $B$) is categorized into another layer.}
    \label{fig:fuse_example}
\end{figure}

To make our explanation easier to understand, let us use Fig.\ref{fig:fuse_example} as a conceptual example although the real scenarios should be different from it. 
As shown in Fig.\ref{fig:fuse_example}, the scene includes an underpass and an overpass. When a robot carrying a LiDAR sensor moves along the dark line in Fig.\ref{fig:fuse_example}(a), correspondingly, the whole route can be split into five segments  (red$\rightarrow$dark$\rightarrow$green$\rightarrow$dark$\rightarrow$blue, shown in (b)) based on the visibility of the cell $m$ to the LiDAR sensor. For example, the red, green and blue segments are the first, third and fifth one, respectively, where the cell $m$ is visible to the LiDAR sensor. The two dark segments represent the two courses where the cell $m$ is  non-visible to the LiDAR sensor, 
and will not affect the result at cell $m$, so these parts do not need to be discussed in the fusion process. 
Note that there are two levels at the cell $m$ location, and the top level of the cell $m$ is visible only to the red and blue segments, and the bottom level of $m$ is visible only to the green segment.
To efficiently generate the 3D-OGM, only the representative observations of the red, green and blue segments are utilized during fusion.  
Let us take the representative observations of the red segment as an example. When the robot moves from the starting point to the end of the red segment, the distance between the robot and the cell $m$ first decreases and then increases, with the minimal distance achieved at location $A$. Generally, the variance of measuring the ground cell $m$ at location $A$ is the smallest. We set the observation made to the cell $m$ at location $A$ is denoted as the representative one of the red segment. Likewise, the observations made to the cell $m$ at locations $B$ and $C$ are the representative ones of the green and blue segments, respectively. 

We first give some notions before the detailed description. A certain cell in the real world (the global map) can be denoted by $c_i$, and there might be multiple-level planes at $c_i$. The distance between the moving robot and $c_i$ over time is calculated as $d_{c_i}(t)$, where $t$ is a time variable. 
The observations made to $c_i$ by a robot over time in a segment $s_j$ (e.g., the red one in Fig.\ref{fig:fuse_example}) are denoted as $\mathbf{m}_{c_i}^{s_j}(t), t\in[the \; period \; when \; robot \; was \; running \; in \;s_j]$, where the representative observation is represented by $\mathbf{m}_{c_i}^{s_j}(t^{*})$. That means the robot is closest to $c_i$ in the segment $s_j$ at time $t^{*}$. 

As illustrated in Fig.\ref{fig:fuse_example}, if assuming the whole course includes $N$ segments, $s_j,j\in[1,N]$, where the cell $c_i$ is totally visible to the LiDAR sensor, we thus have $N$ representative observations made to $c_i$, $\mathbf{m}_{c_i}^{s_j}(t^{*}),j\in[1,N]$. Without losing generality, each of the representative observations $\mathbf{m}_{c_i}^{s_j}(t^{*}),j\in[1,N]$ can initially be assumed to be made on each individual ground plane of different altitude (if assuming that there might exist more than one level of ground planes at the location of cell $m$). 
From a probabilistic perspective, $\mathbf{m}_{c_i}^{s_j}(t^{*}),j\in[1,N]$ is subject to a Gaussian Mixture Model (GMM) in theory. Similarly, this applies to the representative observations made to the other cells. 

Before we estimate the GMM representation of the altitude at each cell $c_i$, we would first apply a merging step to remove the redundant ground planes. 
For example, the representative observations made at locations $A$ and $C$ to the cell $m$  in Fig.\ref{fig:fuse_example} should be merged because they are at the same altitude. In contrast, the representative observations made at $A$ and $B$, or $C$ and $B$, respectively, are not at the same altitude and should not be merged. 

\begin{figure*}
    \centering
    \subfloat[]{ \includegraphics[width=0.33\textwidth]{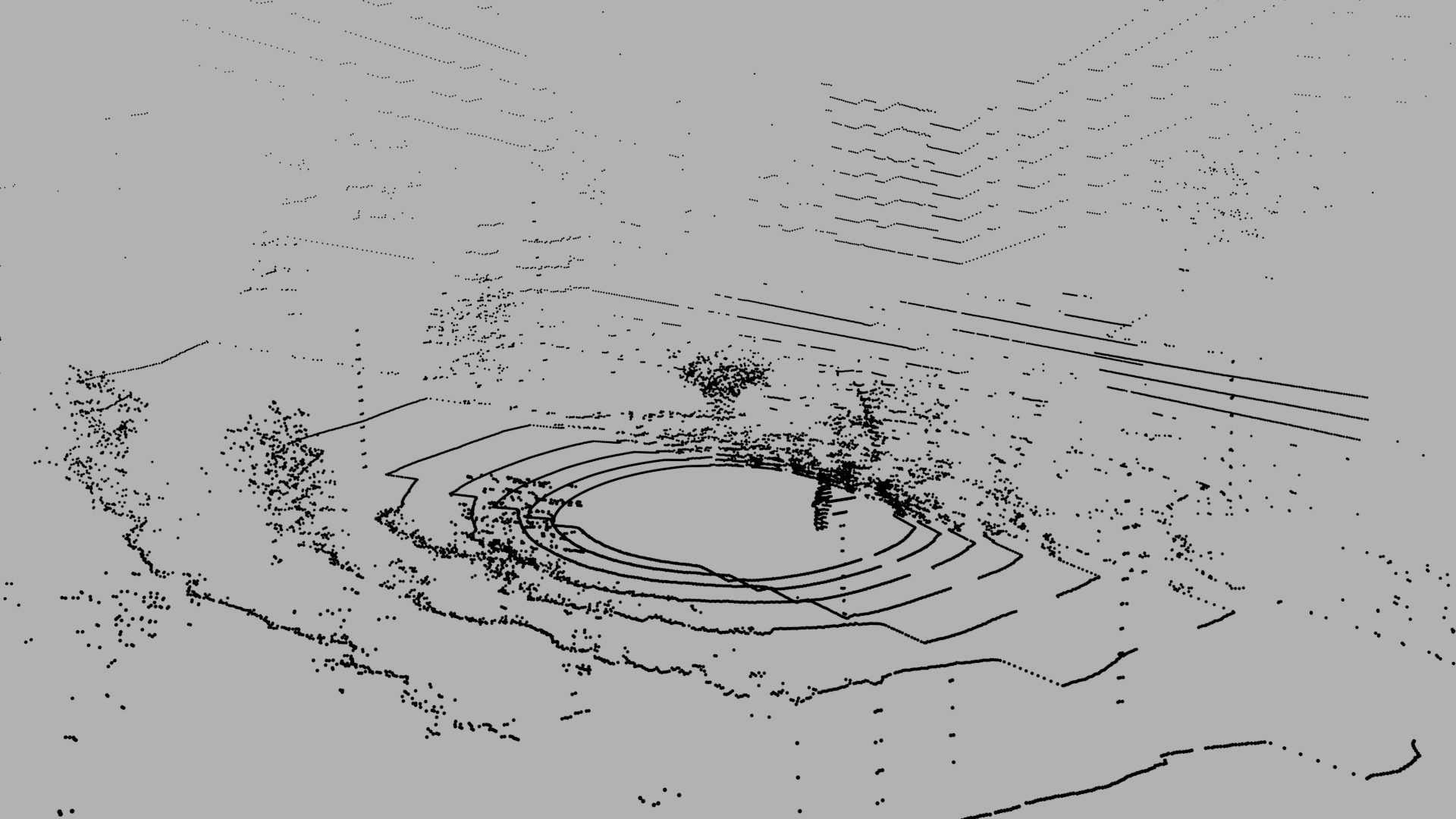}}
    \subfloat[]{ \includegraphics[width=0.33\textwidth]{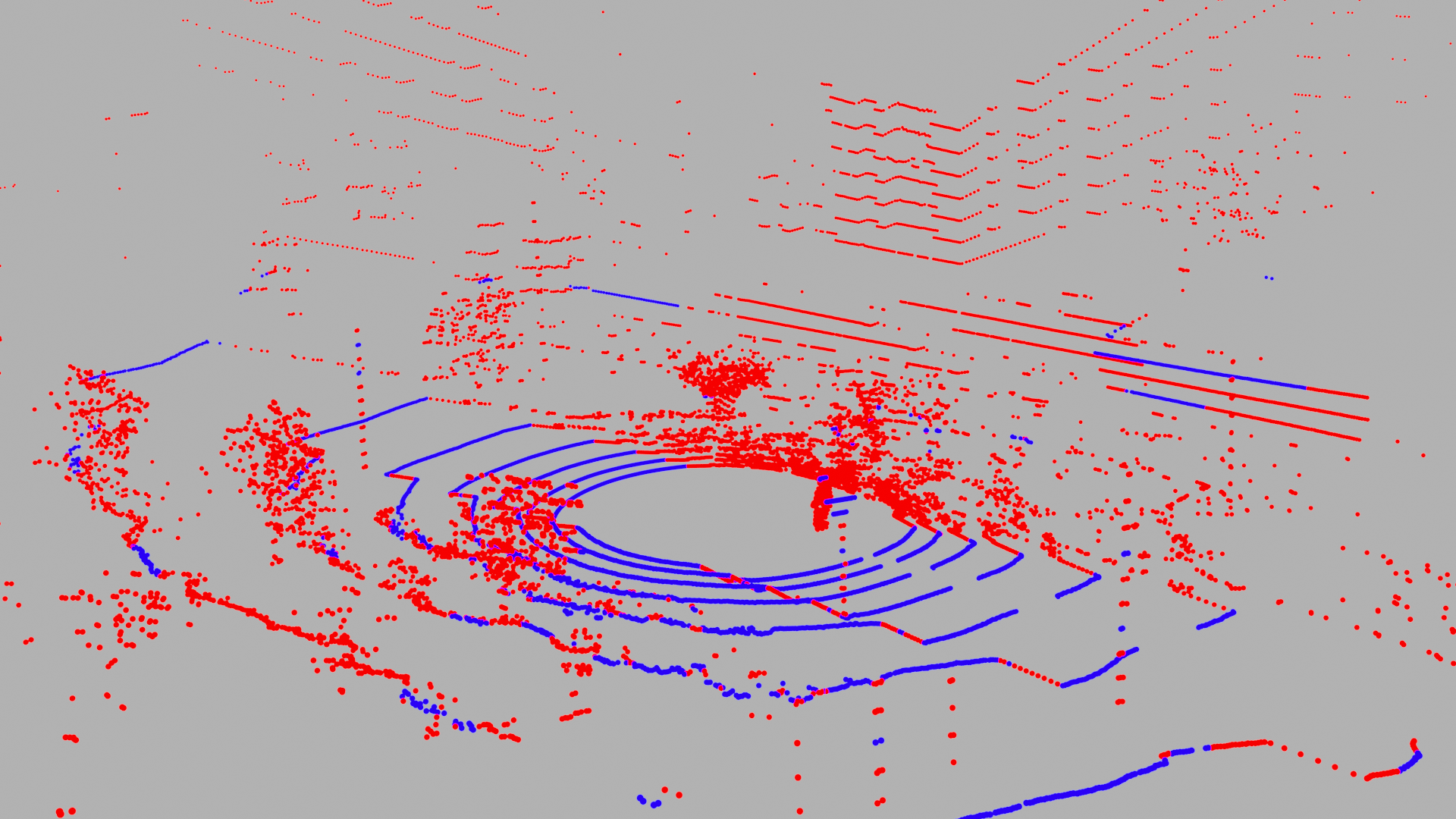}}
    \subfloat[]{ \includegraphics[width=0.33\textwidth]{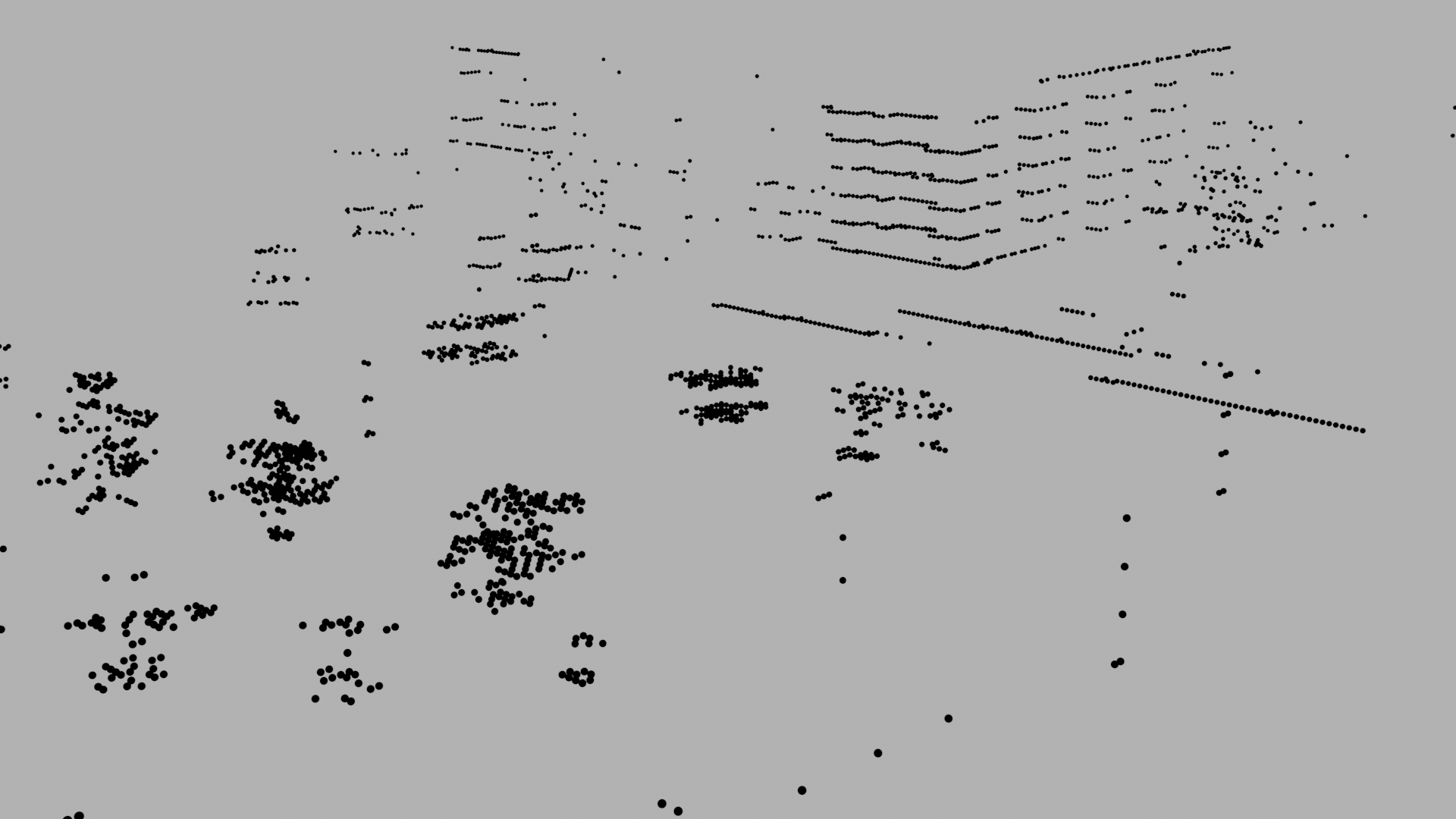}}
    \caption{(a) is a frame of raw point cloud data. (b) is the result after applying the traversable region detection method introduced in Section \ref{section3.1}, where the red points belong to obstacle and the blue points belong to ground. (c) the reconstructed point cloud based on the encoded descriptor of each cell as described in Section \ref{section3.4}.}
    \label{fig:obs_step}
\end{figure*}

According to the method in \cite{Sun:measuring11}, we use the \textit{overlaprate} ($OLR$) to evaluate whether these independent observations should be merged or not.
To do this, the set of 2D local occupancy grids that correspond to the representative observations $\mathbf{m}_{c_i}^{s_j}(t^{*}),j\in[1,N]$ are used, represented by
$\mathbf{ogm}_{c_i}^{s_j}(t^{*}),j\in[1,N]$. Note that the altitude of each of these occupancy grids can also be represented by a Gaussian mixture model in the global coordinate system (refer to Section III.B). Thus, we first sort them based on the means of their altitudes. 

The lowest one with the smallest mean value of altitude,  e.g., $\mathbf{ogm}_{c_i}^{s_q}(t^{*})$, is initially assigned the label ``1''. The second lowest, e.g., $\mathbf{ogm}_{c_i}^{s_k}(t^{*})$, is compared with $\mathbf{ogm}_{c_i}^{s_q}(t^{*})$ based on altitude difference, and is assigned a label by checking whether the altitude difference is larger than a preset threshold. 
Generally, given the label of   $\mathbf{ogm}_{c_i}^{s_q}(t^{*})$, $f_{c_i}^{s_q}$, the label of its adjacent neighbor, $f_{c_i}^{s_k}$, can be updated based on (\ref{eq:cjadd1}),

\begin{equation}\label{eq:cjadd1}
f_{c_i}^{s_k} = \left\{
\begin{aligned}
&f_{c_i}^{s_q}     ,& & \text{if } OLR(\mathbf{ogm}_{c_i}^{s_q}(t^{*}), \mathbf{ogm}_{c_i}^{s_k}(t^{*})) > \epsilon \\
&f_{c_i}^{s_q} + 1 ,& & \text{else}
\end{aligned}
\right.
\end{equation}

where $OLR(\cdot,\cdot) = p_{saddle} / p_{sub\_max}$. The $p_{saddle}$ is the saddle-point location of the joint distribution of the altitudes of $\mathbf{ogm}_{c_i}^{s_q}(t^{*})$ and $\mathbf{ogm}_{c_i}^{s_k}(t^{*})$, and $p_{sub\_max}$ is the location of the smaller peak of the two modes. The $\epsilon$ is an threshold and can be 0.6 as described in \cite{Sun:measuring11}.

In the end, we get $Z$ independent labels which 
means there are $Z$ layers of ground surface at $c_i$.
Then each observation through the whole course can be assigned one of the $Z$ labels based on (\ref{eq:cjadd2}). 


\begin{equation}\label{eq:cjadd2}
f_{c_i}(t) \!=\! \left\{
\begin{aligned}
&f_{c_i}^{s_1} ,&\!\!&\!\!t < t_1^{*} \\
&f_{c_i}^{s_N} ,&\!\!&\!\!t \ge t_N^{*} \\
&f_{c_i}^{s_j} ,&\!\!&\!\!t_j^{*} \le t < t_{j+1}^{*}, |\mu_j^* - \mu_t| \le |\mu_{j+1}^* - \mu_t| \\
&f_{c_i}^{s_{j+1}} ,&\!\!&\!\!t_j^{*} \le t < t_{j+1}^{*}, |\mu_j^* - \mu_t| > |\mu_{j+1}^* - \mu_t|
\end{aligned}
\right.
\end{equation}

where $\mu_t$ is the altitude of the observation made at time $t$, and the variables with $*$ are associated with the representative observations. For example, the $t^{*}$ is the time instance at which a representative observation is made, and $\mu^{*}$ is the mean altitude value of a certain representative occupancy grid.

Finally, we will get $Z$ sequences that can be used for fusion. According to the general elevation map fusion method \cite{Triebel:IROS06}, the $Z$ sequences are fused sequentially to generate the final result. Note that because our framework is an online mapping process, it is not necessary to wait until all observations thought the whole course are made. The update in (\ref{eq:cjadd3}) is an online process as well in real implementation.

\begin{equation}\label{eq:cjadd3}
\begin{split}
\mu_{f,0:n_f} &= \frac{\sigma_{f,n_f}^2 \mu_{f,0:n_f-1} + \sigma_{f,0:n_f-1}^2 \mu_{f,n_f}}{\sigma_{f,0:n_f-1}^2 + \sigma_{f,n_f}^2} \\
\sigma_{f,0:n_f}^2 &= \frac{\sigma_{f,0:n_f-1}^2\sigma_{f,n_f}^2}{\sigma_{f,0:n_f-1}^2 + \sigma_{f,n_f}^2} \\
\mu_{f,k} &\in \{\mu_t|f_{c_i}(t) = f\} ,\; \sigma_{f,k} \in \{\sigma_t|f_{c_i}(t) = f\}
\end{split}
\end{equation}

Where $\mu_{f,0:n_f}$ and $\sigma_{f,0:n_f}$ represent the fusion result from the first observation to the last observation that has been assigned the label $f$, and $n_f$ is the number of observations labeled as $f$.

The whole merging algorithm is based on the observations for the current cell and the corresponding pose. Therefore, when the poses of local maps change, it is only necessary to update the area in the global map which is covered by the local map whose pose changes. 

\subsection{Embedding geometric information for vehicle localization}\label{section3.4}
The above steps can be used to create a 3D road-map or traversable-region map, which is mainly for vehicle path planning purposes. Sometimes it is also useful to keep vehicle run in the track when vehicle localization is not accurate or totally gets lost. 
But for accurate vehicle localization, the above created map cannot be reliable generally although it should be applicable to 2D structured scenes. When vehicle is running in a non-structured 3D scene, we have to incorporate 3D vertical geometric structures into the created road map. Inserting vertical structures into the road map is in analogy to growing trees along road which stretches away into the distance. 



To do that, vertical structures should be 
detected in each LiDAR point-cloud frame and then inserted into the generated road-map. Based on our scheme presented in Section \ref{section3.1}, 
we initially get the point-cloud region pc$_{obst}$ which corresponds to obstacles (e.g., trees, bushes, light poles or buildings). However, Dynamic objects (moving pedestrian or vehicles) can also be detected as obstacles because this scheme cannot distinguish dynamic objects from static obstacles. Dynamic objects are not expected to be reserved as vertical structures in the final LiDAR-Atlas because they can affect adversely  localization performance when the LiDAR-Atlas is used for navigation. 
In addition, the static obstacle points are somewhat redundant for localization and should be sparsified for efficiency purposes.

To deal with these issues on vertical geometric structures, we propose a probabilistic way to achieve the aim of reducing redundant 3D points and removing dynamic objects as much as possible. Specifically, 
an efficient 3D probabilistic grid map, 3D-PGM, in $x-y-z$ dimensions ($z$ representing the vertical dimension) is created to reserve sparse vertical static structures, where the resolution of $x-y$ plane is set to be 0.1m/cell-width (the same as the one adopted in creating the 2D-OGM in Section \ref{section3.2}). 
Vehicles actually drive on the ground, and there is no significant change in height for dynamic objects in vertical dimension locally. So we set a sparser resolution in the vertical dimension. 
The lowest grid in the 3D-PGM is slightly higher than the ground so as to effectively ignore the noisy obstacle points that may exist near the ground. 

\begin{figure*}[h]
    \centering
    \subfloat[]{ \includegraphics[width=0.2\textwidth]{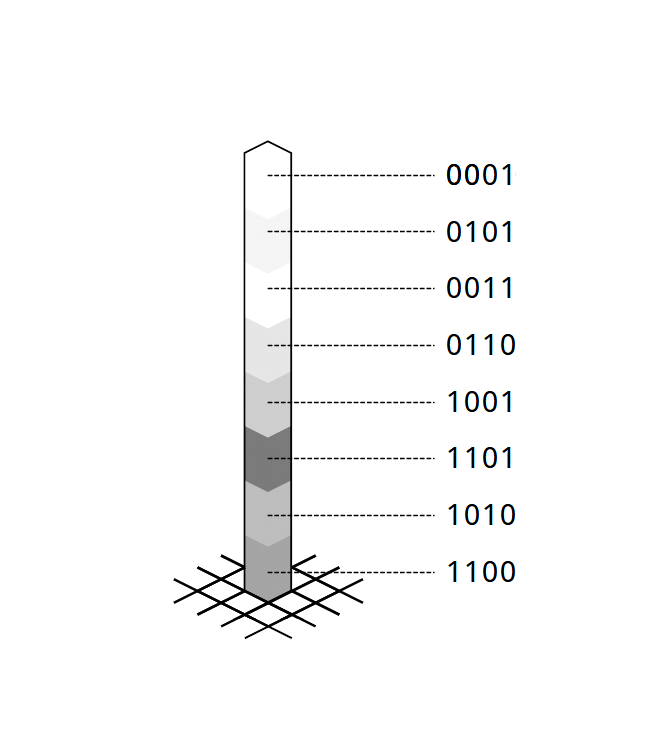}}
    \subfloat[]{ \includegraphics[width=0.2\textwidth]{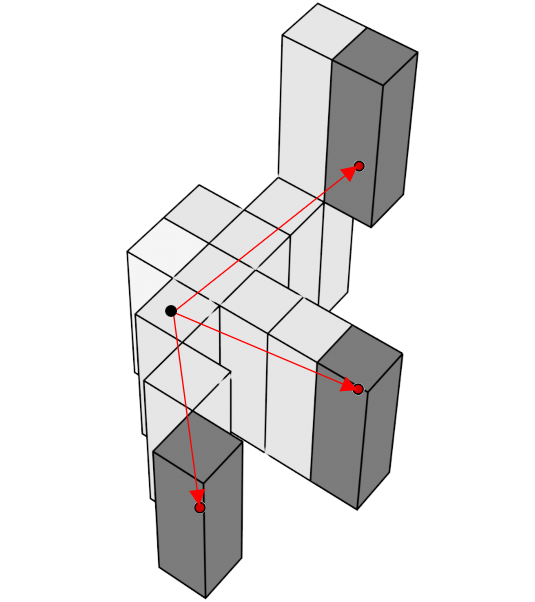}}
    \caption{Grey-scale level shows the occupancy probability of a segment. The darker the grey-scale level, the larger the probability of being occupied. (a) is an illustration of probabilistic coding, where multiple segments corresponding to the same 2D location are shown. In this example, we encode the probability with 4 bits and label it on the right. (b) shows the process of converting an obstacle point cloud to a 3D probability grid map for a LiDAR frame. The dark dot is the position of the LiDAR sensor and the three red dots represent three obstacle-grid locations.}
    \label{fig:pgm-cell}
\end{figure*}

To remove dynamic obstacles through the 3D probabilistic grid map, we find that it is not necessary to use high-precision numbers to represent probability values. Specifically, for each 3D probability cell, we can use fewer bits (4 bits) to store a probability value to reduce memory cost during runtime, shown in Fig.\ref{fig:pgm-cell} (a). The encoding process is a simple linear mapping of probability values to the range 1 to 15. Exclusion of 0 is to ensure that there are an odd number of states, ensuring that 0.5 (unknown state) can be represented exactly. So 1 to 7 means non-obstacle, 8 meaning unknown, and 9 to 15 meaning obstacle. 
In contrast, 2-bit representation cannot express uncertainty under the condition of odd state (e.g., the part greater than 0.5 or less than 0.5 contains multiple possible values), and 3-bit representation cannot fill all bytes when the number of segments is not a multiple of 8. 4-bit representation is the most saving and appropriate option.
Then, we encode the occupancy probability for all segments at the same 2D position based on the 4-bit representation. In implementation,  the probability update is by mapping back to the range of 0 to 1 from the 4-bit representation based on efficient look-up table. Considering the result of this encoding as a descriptor, the representation is ultimately consistent with a 2D map.

\begin{figure*}[h]
    \centering
    \subfloat[]{ \includegraphics[width=0.30\textwidth]{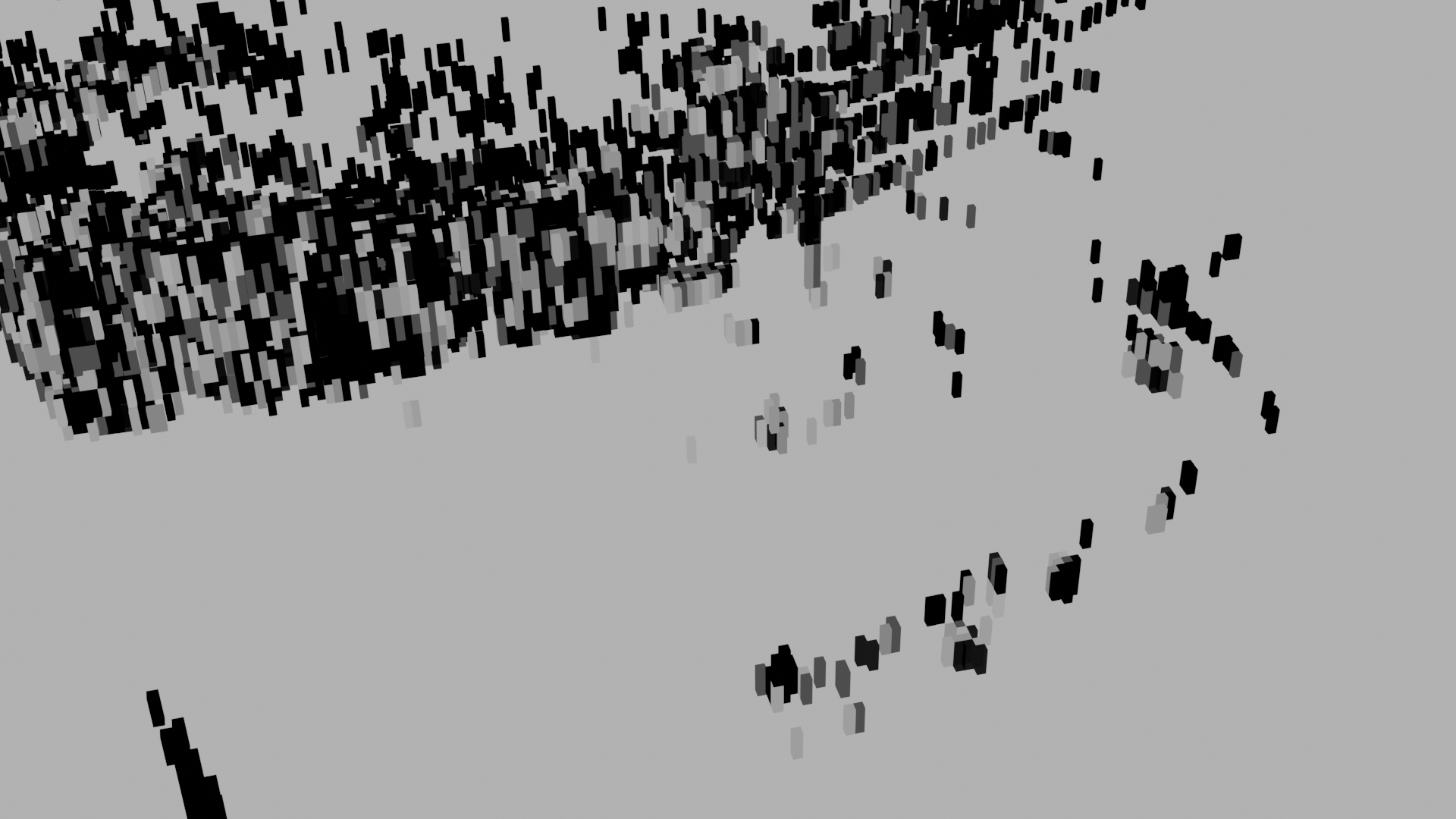}}
    \subfloat[]{ \includegraphics[width=0.30\textwidth]{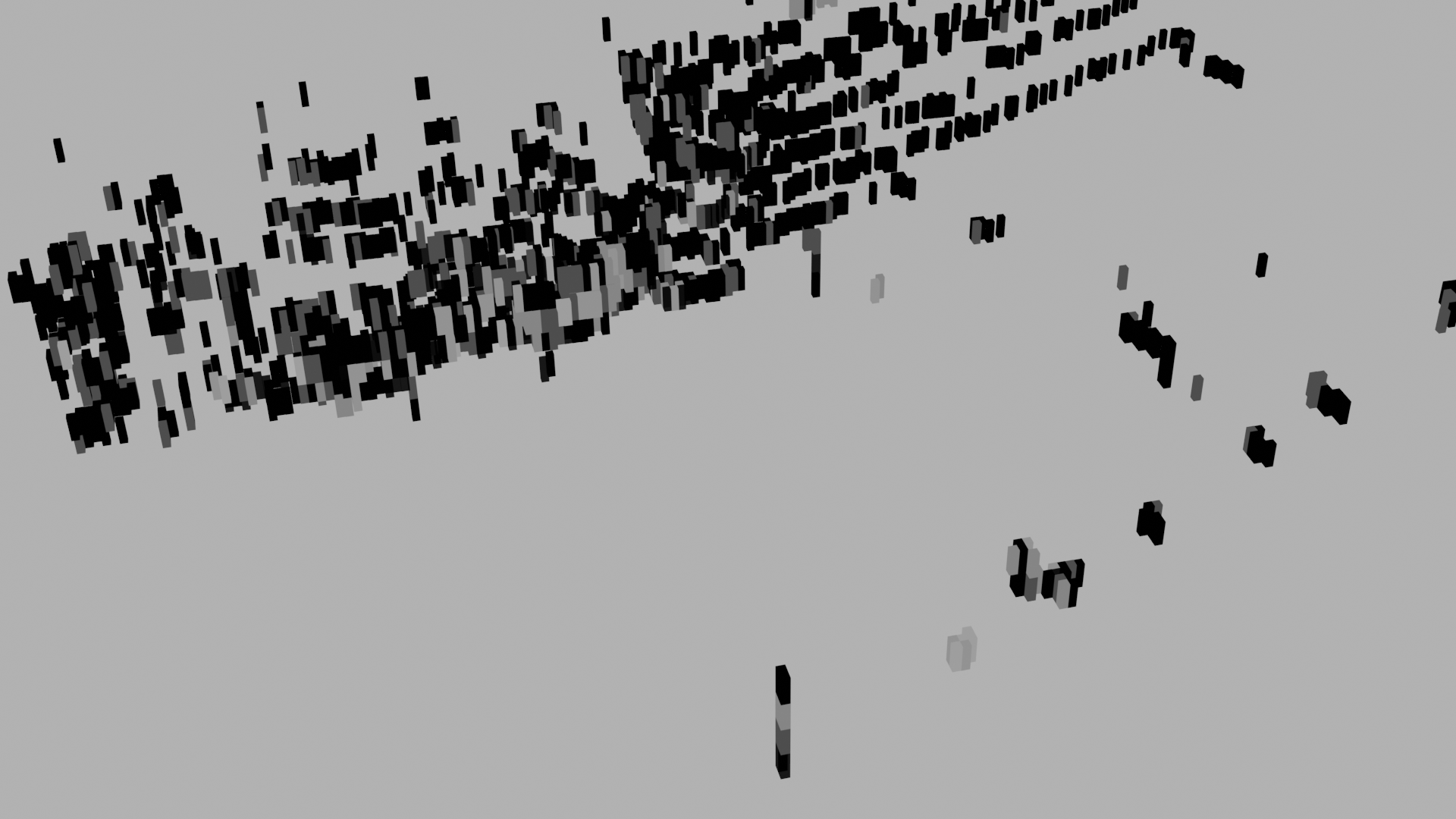}}
    \subfloat[]{ \includegraphics[width=0.30\textwidth]{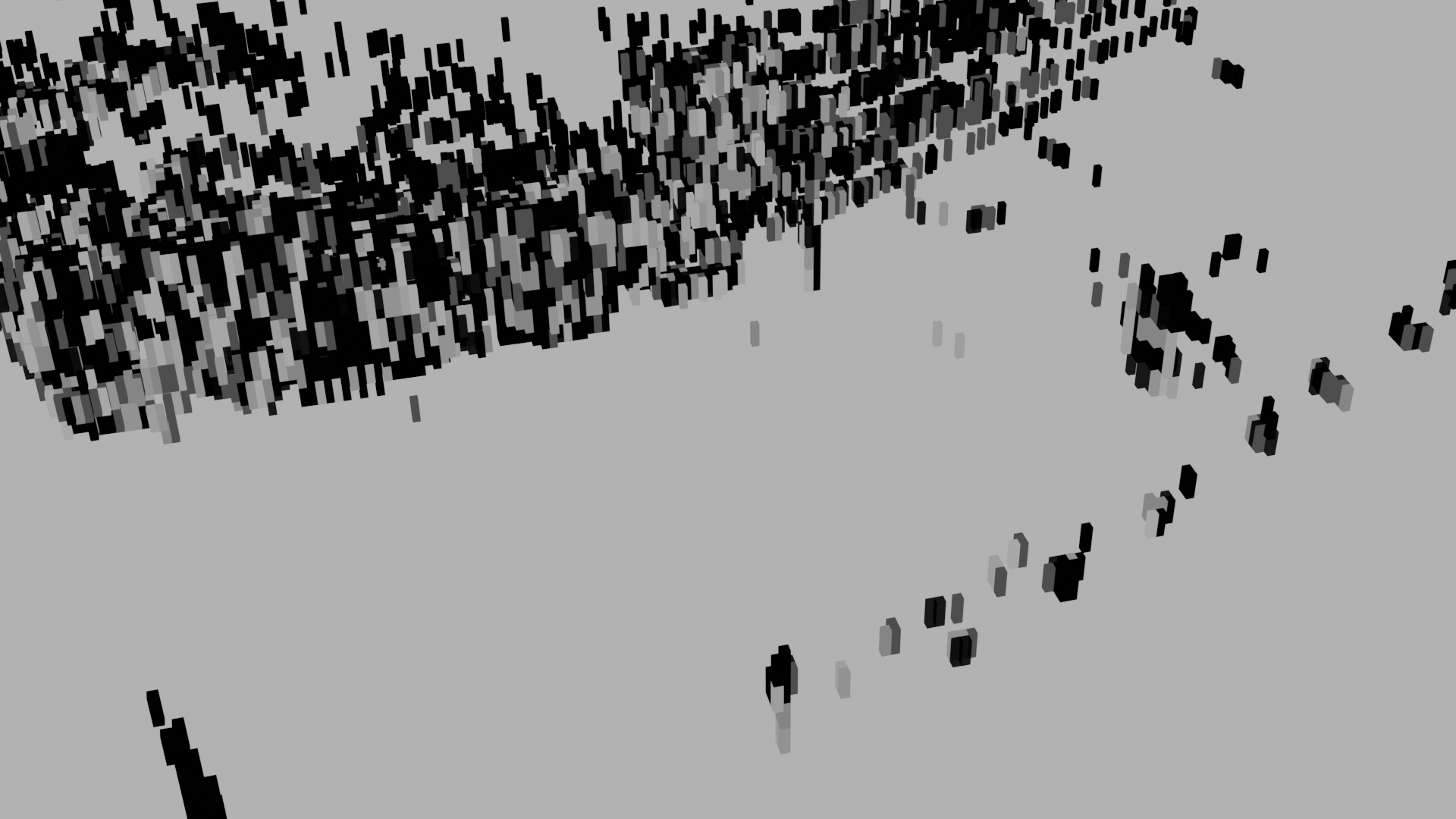}}
    \caption{An illustration of the process of fusing two neighboring probabilistic grid maps. The darker the intensity, the higher the probability of being an obstacle (vertical structure). Segments corresponding to traversable road region are not drawn.
(a) is the global map before adding a new frame. (b) is a new frame. (c) is after adding the new frame.
It can be seen that the probabilities of vertical structures (including dynamic obstacles) are updated and the dynamic obstacles (in the central part of the image) are removed.}
    \label{fig:pgm}
\end{figure*}

Once we have the economical representation of probability values, given a LiDAR frame, the point cloud $pc_{obst}$ (e.g., the red region in the Fig.\ref{fig:obs_step} (b)) corresponding to the vertical structures (including dynamic objects) can be represented with the 3D probability grid map. 
This procedure is similar to 
the one used for constructing a general 2D occupancy grid map, where a miss probability $p_{miss}$ is applied to all segments passed by the ray from the LiDAR to the obstacle point, and a hit probability $p_{hit}$ is applied to the segment where a obstacle point is located. For ease of calculation, the sensor model we use takes fixed values for both $p_{miss}$ and $p_{hit}$. The process of adding obstacle points within a single LiDAR frame is shown in Fig.\ref{fig:pgm-cell} (b).

\begin{figure}
    \centering
    \subfloat[Before the car leaves.]{ \includegraphics[width=0.35\textwidth]{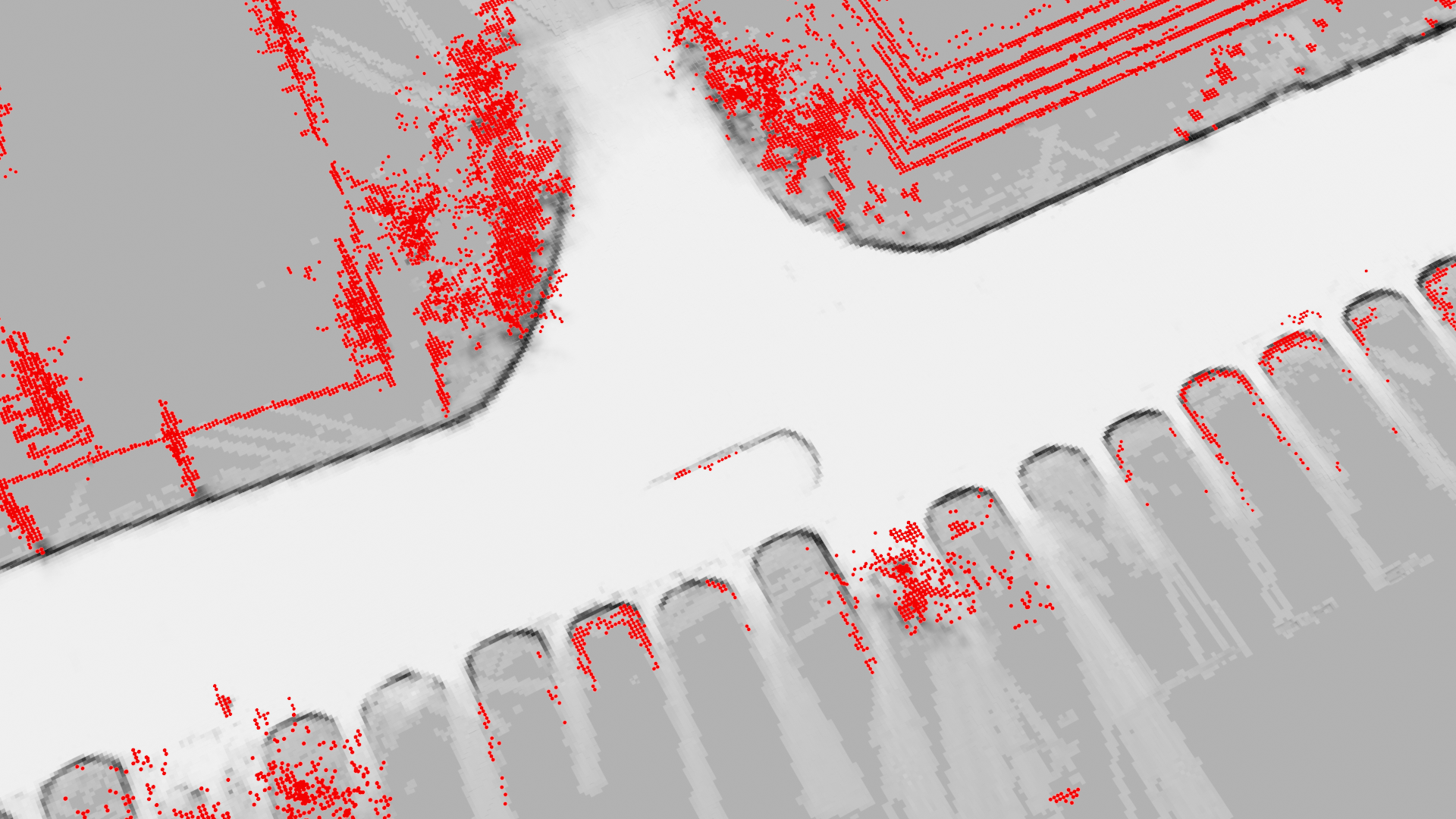} }
    \subfloat[After the car leaves.]{ \includegraphics[width=0.35\textwidth]{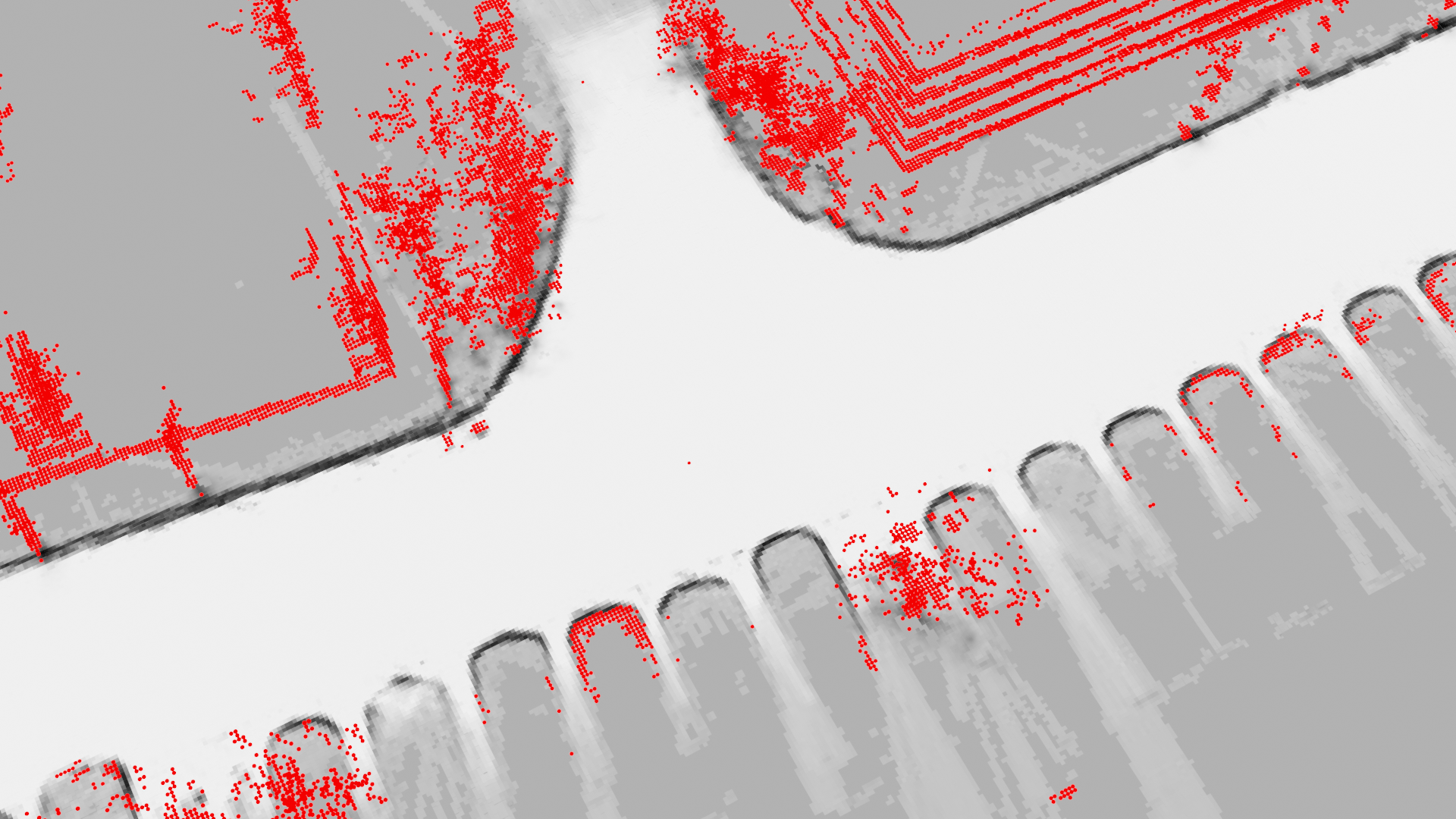} }
    \caption{Illustrating dynamic object filtering by probabilistic fusion: A car is leaving the parking-lot.}
    \label{fig:dynamicScene}
\end{figure}

Reversely, we can decode the probability of each segment from the descriptors of each 2D cell, and based on these probabilities, we can approximately reconstruct the obstacle point cloud with the obstacle segments. The reconstruction result of one frame is shown in Fig.\ref{fig:obs_step} (c).
Over time, we can construct a global map from each consecutive local frames by applying a probabilistic mixture model to fuse two 3D probabilistic grid maps. Fig.\ref{fig:pgm} shows a update of the probabilistic grid map by adding a new frame (each key frame as a local map) to the existing global map. We also show one example of filtering out dynamic objects with the probabilistic mixture in Fig.\ref{fig:dynamicScene}.

To apply the map to robot localization, one can reconstruct a point cloud from the 4-bit descriptors of a LiDAR frame, and utilize ICP-like registration method for vehicle localization in the map. In our localization implementation, we first encode the frame to be localized, then decode the point cloud for ICP, and then do ICP matching with the point cloud decoded from the global map. The purpose of encoding and then decoding the LiDAR frame to be localized is to make it consistent with the sparsity of the global point cloud.

\begin{figure*}[h]
    \centering
    \subfloat{ \includegraphics[width=0.12\textwidth]{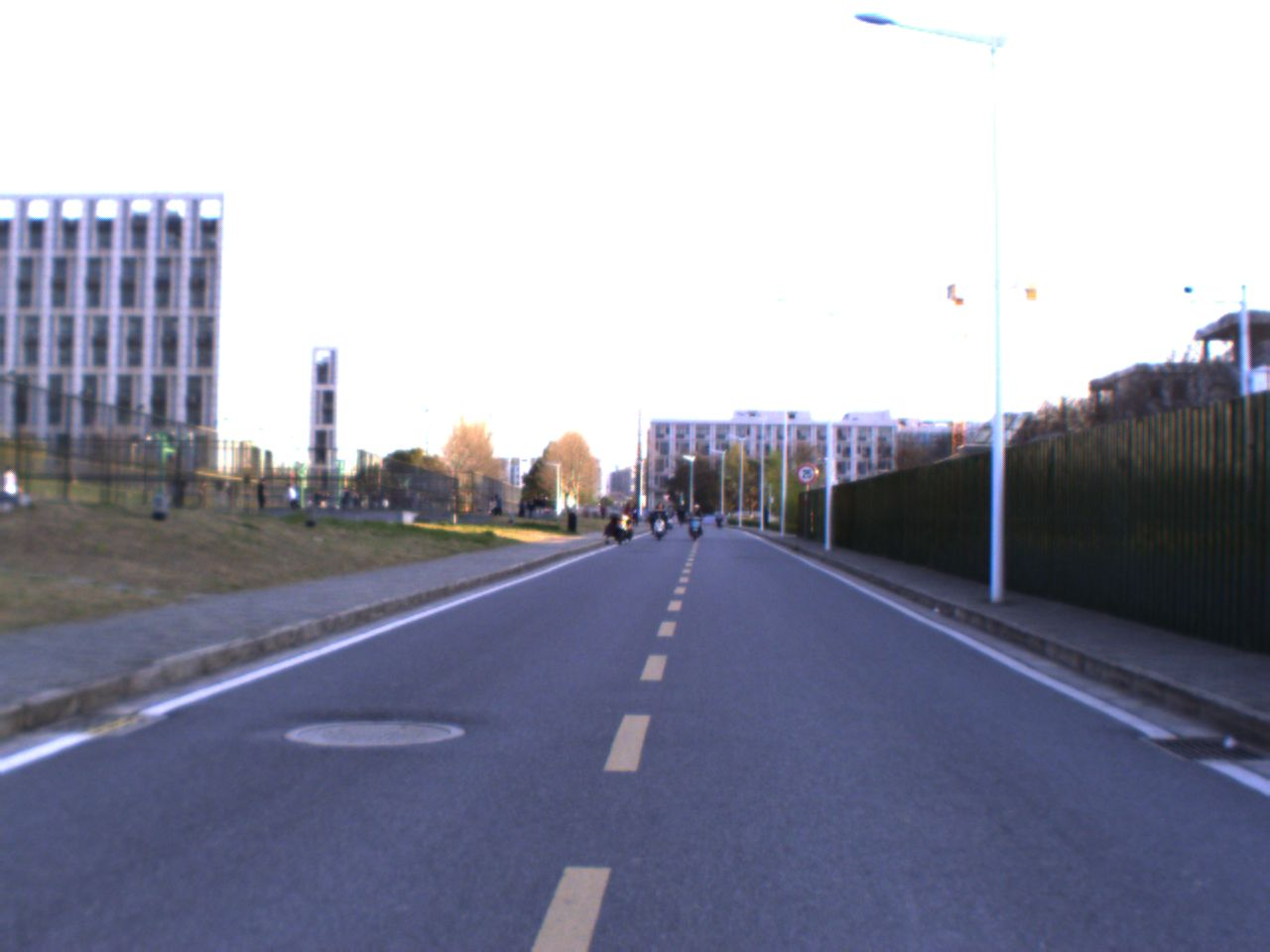}}
    \subfloat{ \includegraphics[width=0.36\textwidth]{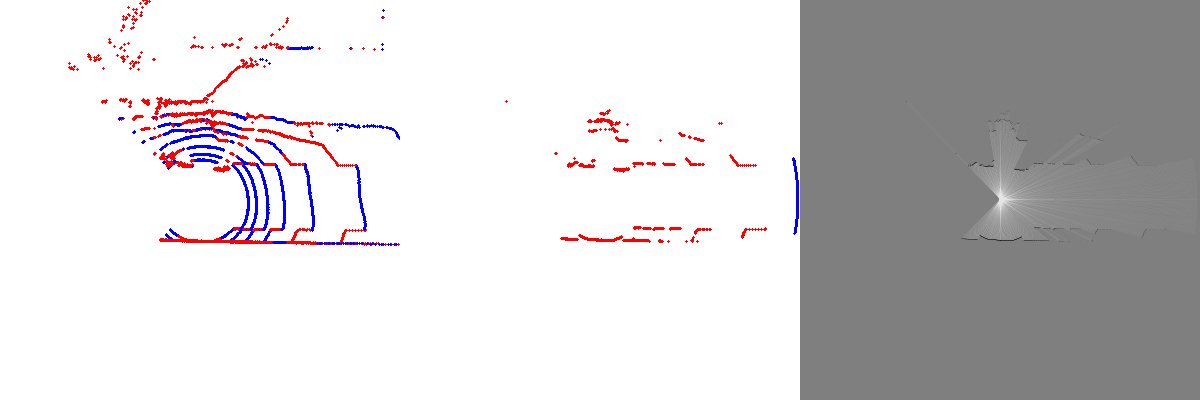}}
    \subfloat{ \includegraphics[width=0.12\textwidth]{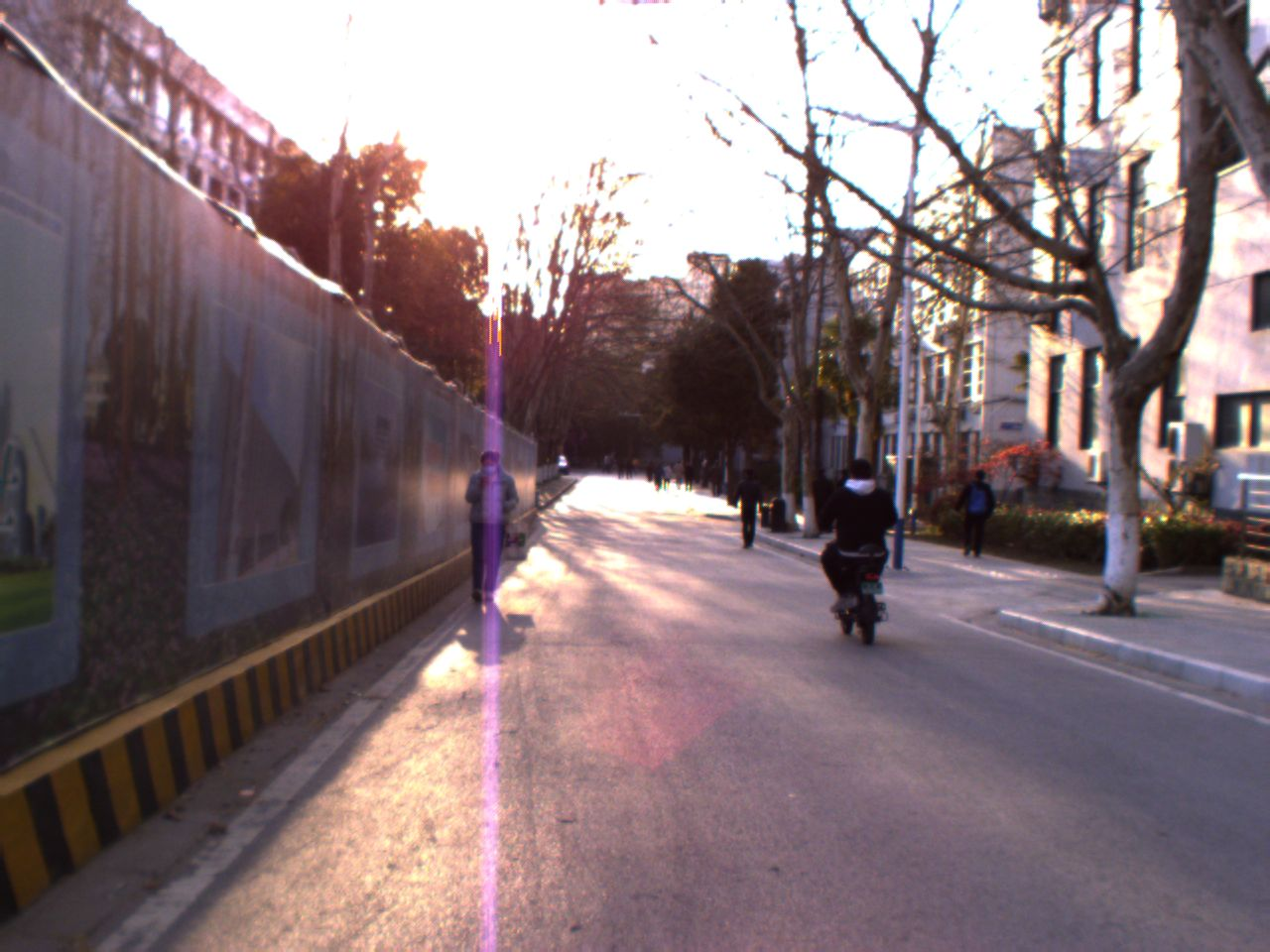}}
    \subfloat{ \includegraphics[width=0.36\textwidth]{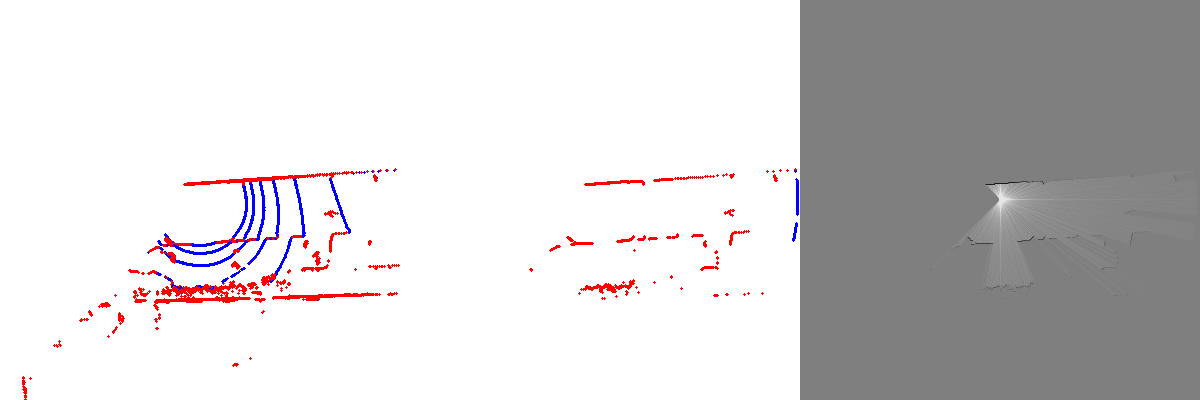}}
    \vskip -8pt
    \subfloat{ \includegraphics[width=0.12\textwidth]{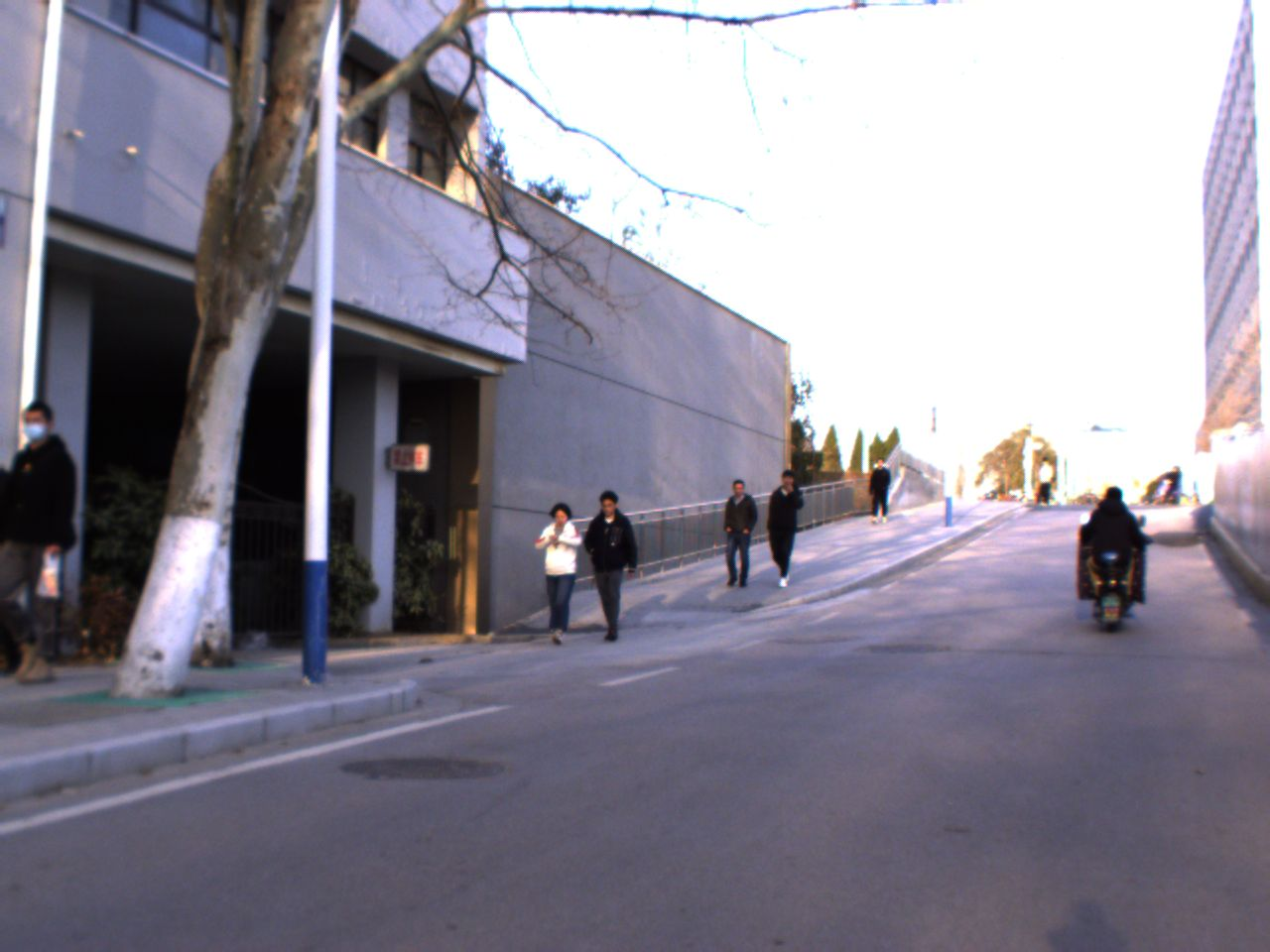}}
    \subfloat{ \includegraphics[width=0.36\textwidth]{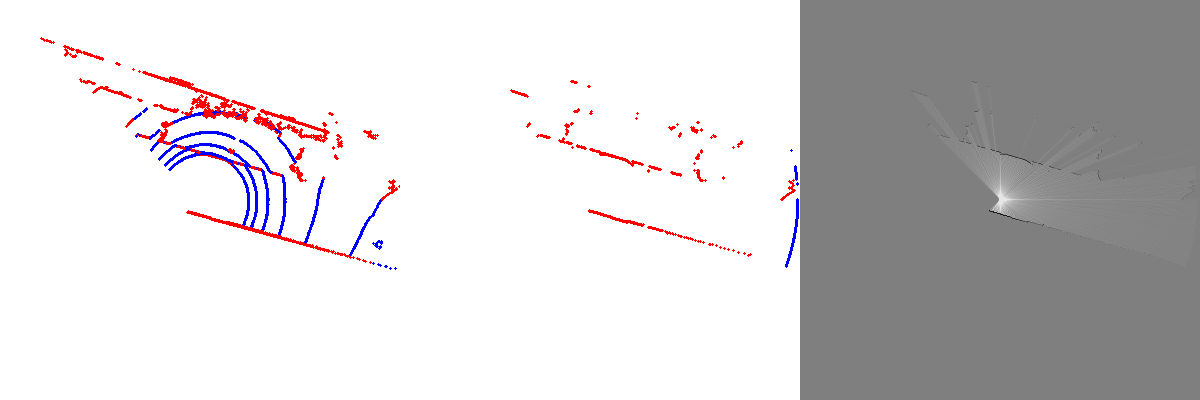}}
    \subfloat{ \includegraphics[width=0.12\textwidth]{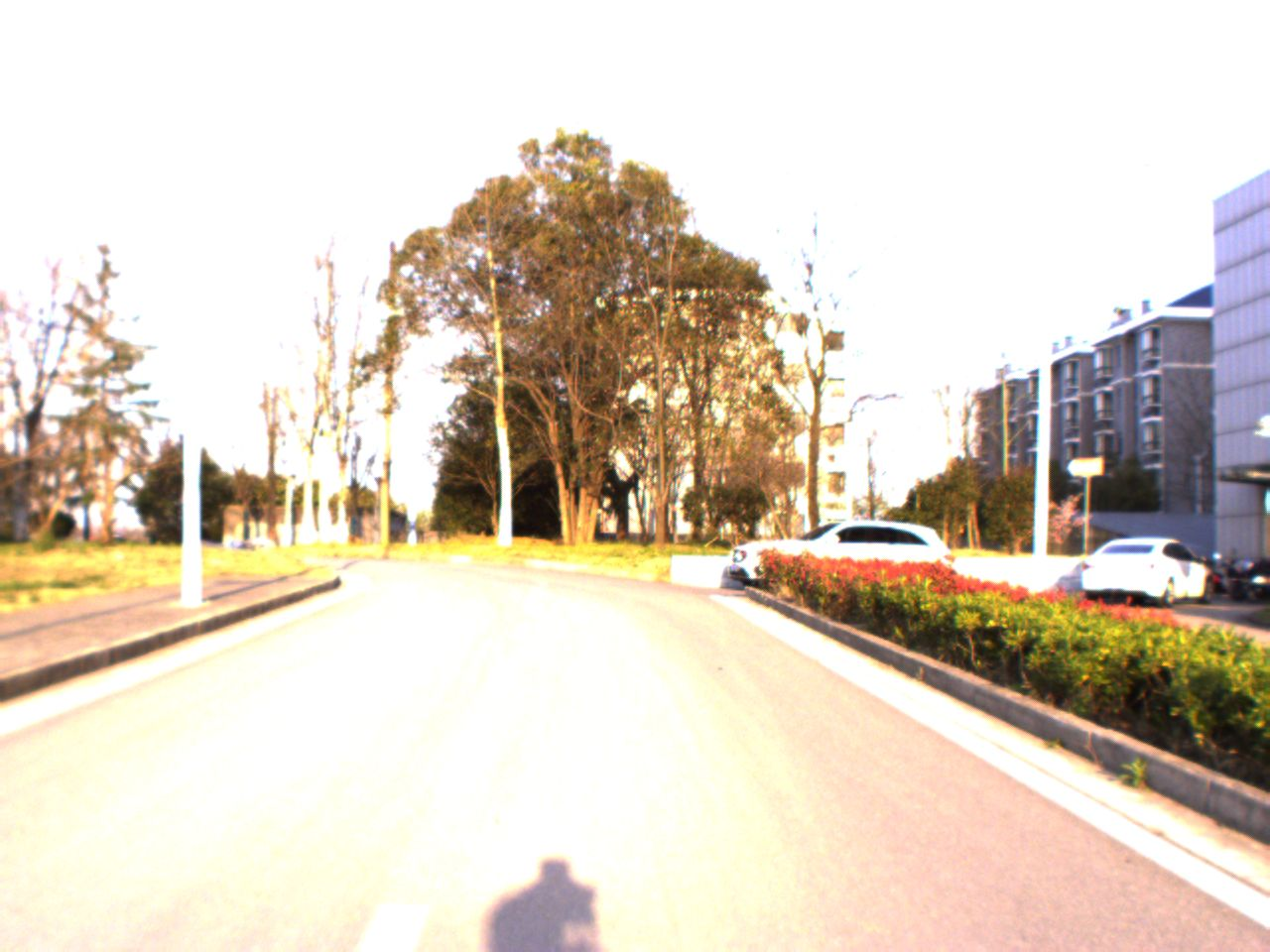}}
    \subfloat{ \includegraphics[width=0.36\textwidth]{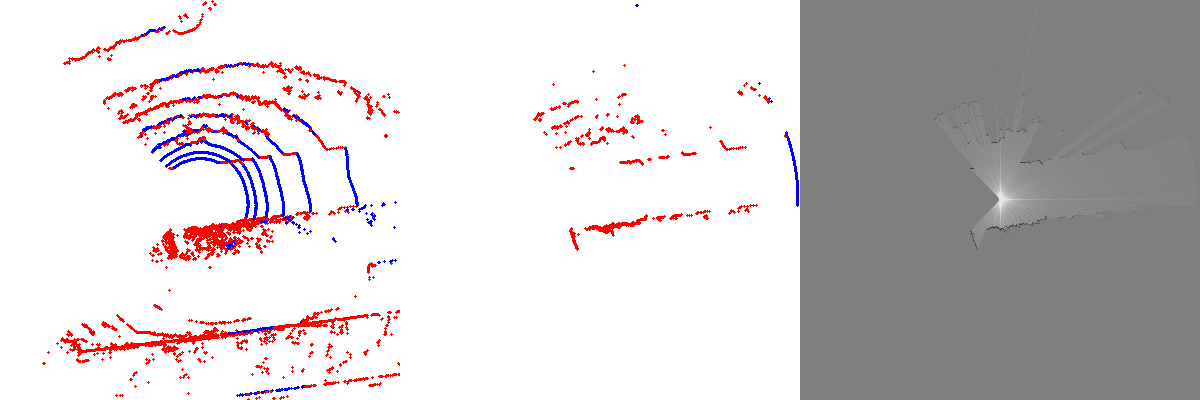}}
    \vskip -8pt
    \subfloat{ \includegraphics[width=0.12\textwidth]{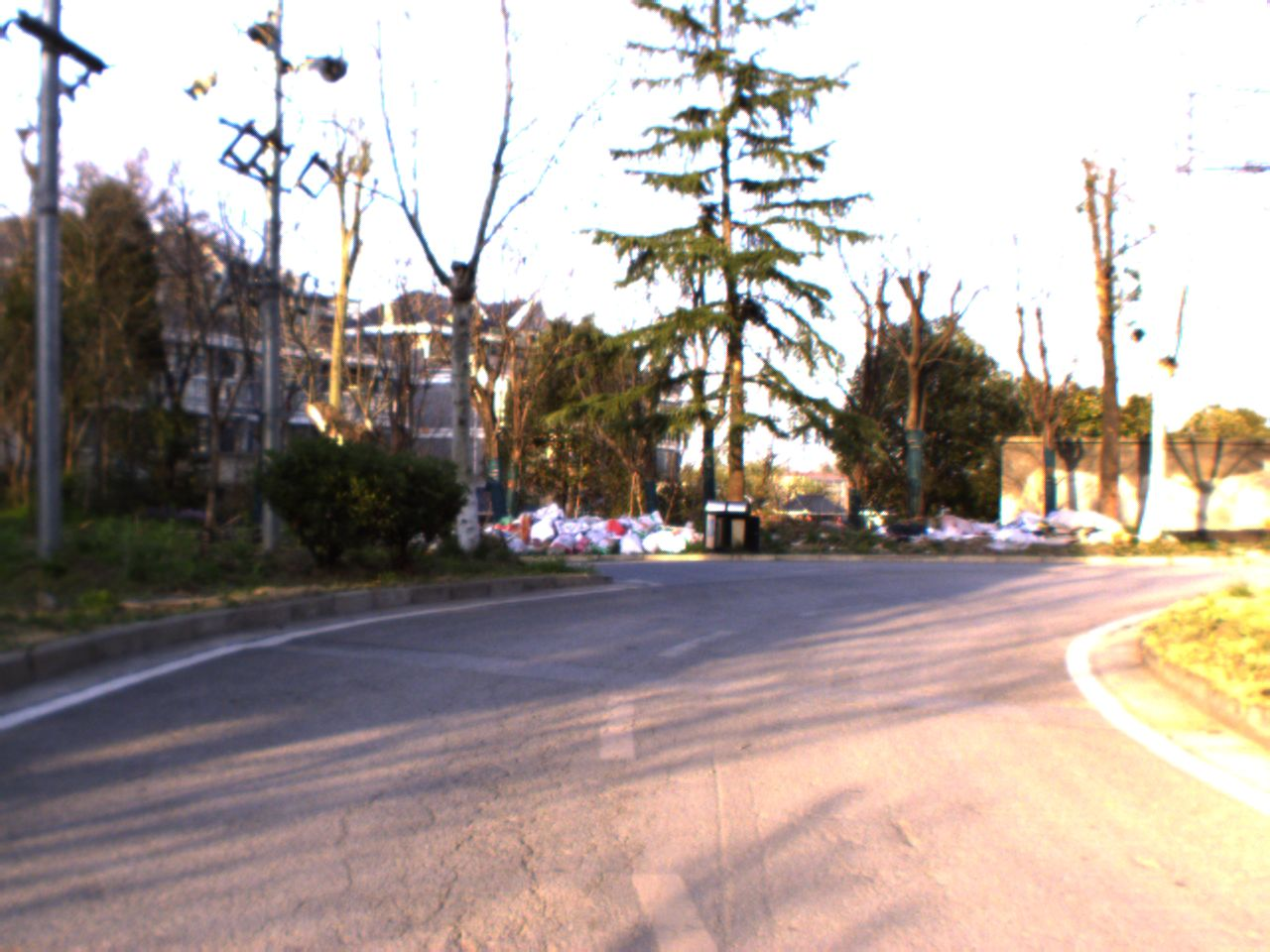}}
    \subfloat{ \includegraphics[width=0.36\textwidth]{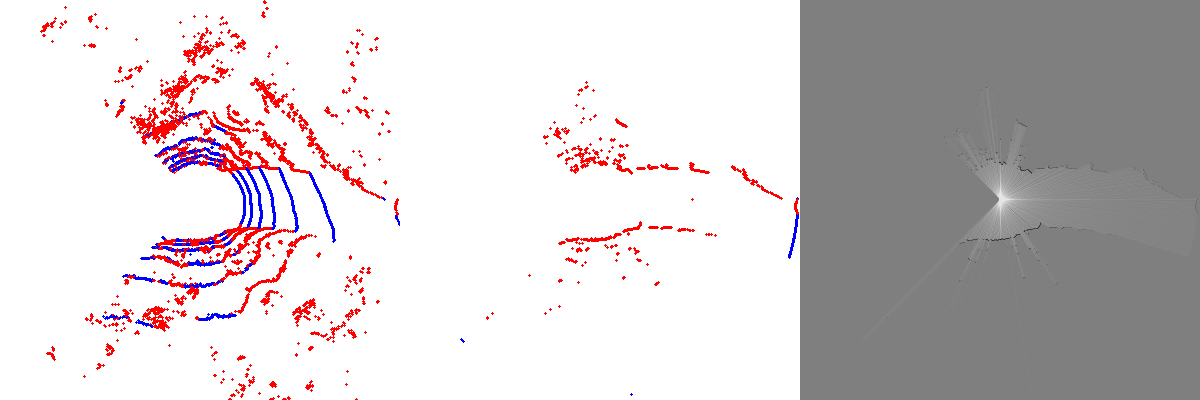}}
    \subfloat{ \includegraphics[width=0.12\textwidth]{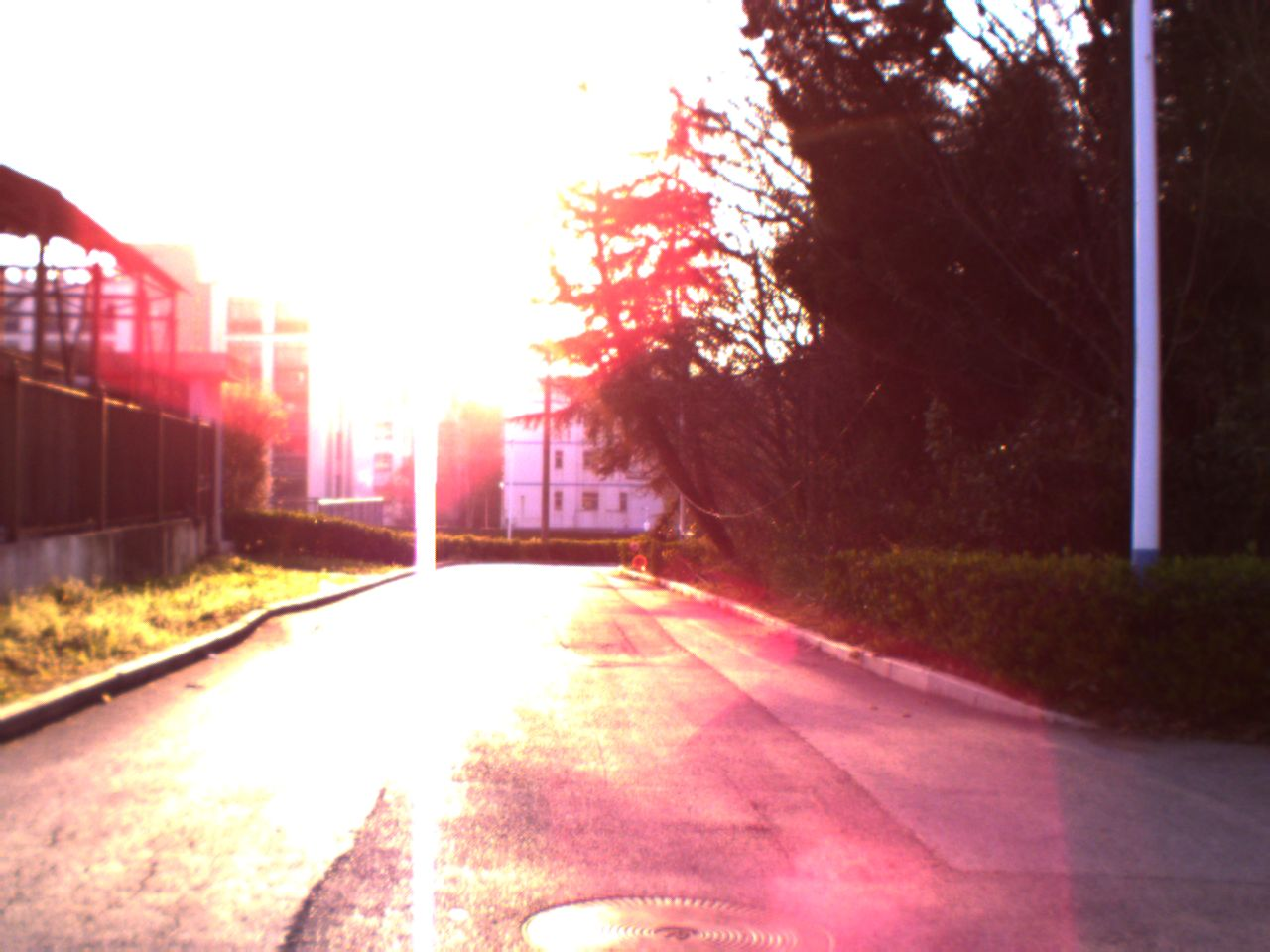}}
    \subfloat{ \includegraphics[width=0.36\textwidth]{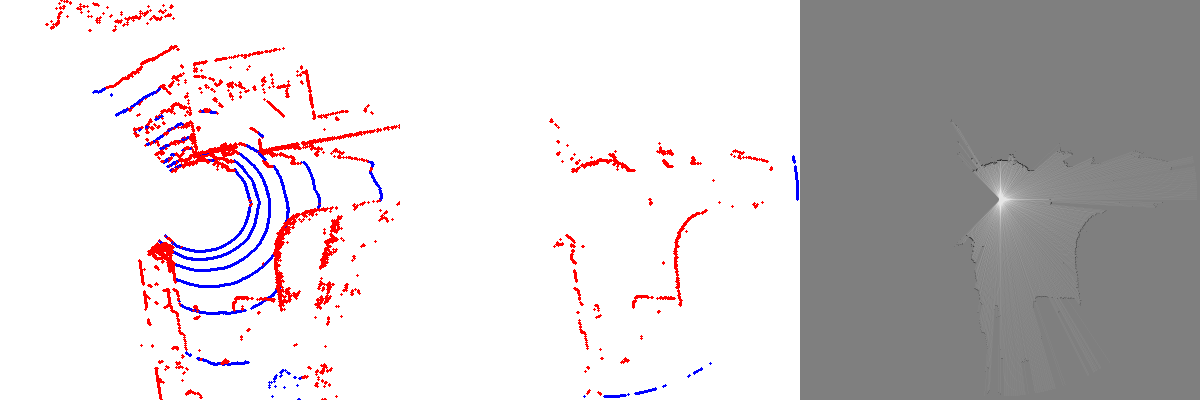}}
    \caption{Examples of creating local 2D occupancy grid maps (2D-OGM) in real environment test, with the columns corresponding to the images captured by a camera installed on the robot, the frames of point cloud by VLP-16, the detected curb regions, and the created 2D-OGMs based on the curbs, respectively.}
    \label{fig:locals}
\end{figure*}



\section{Implementation details and Experimental results}
We adopt the VLP-16 (16-channel) LiDAR as the only sensor to implement the proposed method, although we believe more advanced multi-channel LiDAR sensors (e.g., 32 or 64 channels) should work equivalently or even can achieve better results. 

\subsection{Traversable- and curb-region detection accuracy}
In general, our method is a road-based or traversable-region based map representation. 
Therefore, traversable- and curb-region detection performance is first evaluated 
on the KITTI-360 benchmark dataset \cite{Xie:CVPR16}. 
The reason we choose the KITTI-360 dataset is that this benchmark dataset provides ground-truth semantic segmentation of street scenes. 
Curb detection is crucial for autonomous driving in urban scenes. Actually, in most urban areas, the lowest positive obstacles no matter for autonomous cars or for robots are probably curbs, where positive obstacles are defined as the objects which stick out above the ground. The other positive obstacles such as vehicles or pedestrians are relatively easier to be detected. In addition, we also need curb-region detection results to build the 2D local grid map (Fig.\ref{fig:local_grid}). 

Therefore, both traversable- and curb-region detection performance is evaluated to ensure the accuracy of the created road map. Although our framework is implemented with a VLP-16 LiDAR sensor, it should also work with the other multi-channel LiDAR sensors. To evaluate with LiDAR sensors of different number of channels, we tried our method on the original point cloud data of the KITTI-360 benchmark and on the subsampled versions, respectively. Specifically, we sample one channel from every four channels of the KITTI-360 LIDAR data to simulate a 16-channel LiDAR sensor. We sample one channel from every two channels of the KITTI-360 LIDAR data to simulate a 32-channel LiDAR sensor.

\begin{figure}[h]
    \centering
    \includegraphics[width=0.6\textwidth]{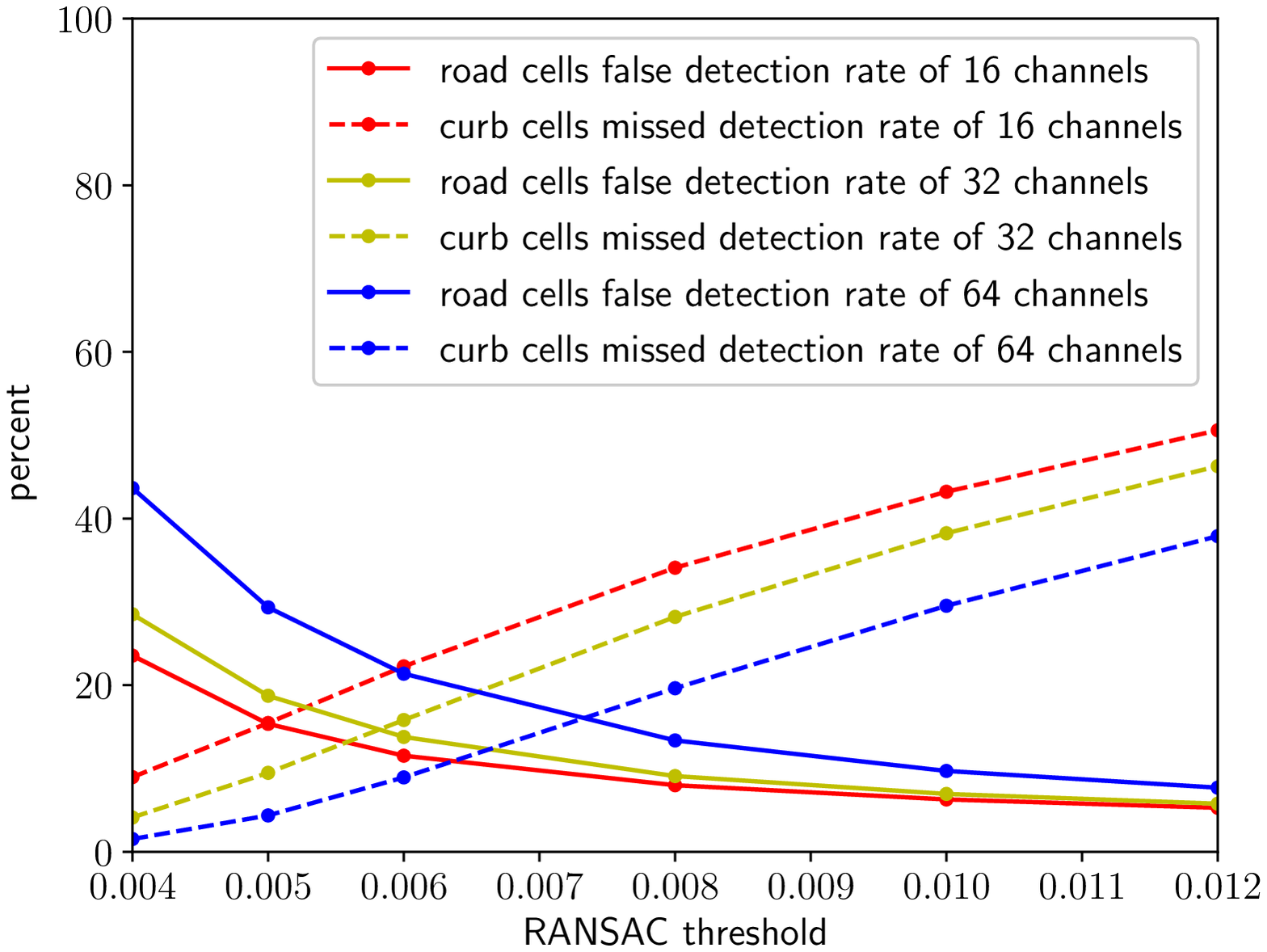}
    \caption{Performance of traversable- and curb-region detection using 16-channel, 32-channel and 64-channel LiDAR point-cloud data with different thresholds in RANSAC.}
    \label{fig:rate}
\end{figure}

The KITTI-360 benchmark provides multi-frame 3D semantic labels. We divide the point cloud into cells at the resolution of 0.1mx0.1m per cell, and label these cells as road-, curb-, or irrelevant-cells based on semantic information, where 
the irrelevant cells mostly contains non-trivial obstacles such as vehicles or trees. 
These labels are automatically generated based on the existing semantic labels of KITTI-360.
Then our curb and traversable detection results represented in the 2D cells are compared with the ground-truth. 

For curb region, we evaluate the $N_{missed}$ only, which is calculated as the number of curb cells that has been labeled as traversable. Additionally, we deem it as correct if a curb cell is labeled as an irrelevant-cell. For traversable region, we evaluate the $N_{false}$ only, which is the number of road cells that are mislabeled as obstacle. 
Given the total number of road cells $N_{road}$ and total number of curb cells $N_{curb}$, we define the missed detection rate of curb cells $rate_{missed}=\frac{N_{missed}}{N_{curb}}$ and false detection rate of road cells $rate_{false}=\frac{N_{false}}{N_{road}}$. Based on different threshold in RANSAC, we finally get the performance curves as shown in Figure \ref{fig:rate}. From the figure, we observe that the missed-detection rate deceases with more channels of LiDAR beams. Counter-intuitively, we have noticed that 
the false detection rate of traversable-region cells increases when using the LiDAR sensor of more channels. The main reason is that the LiDAR sensor with more channels has more points, so there are more false road points generated in general, and these points will fall into more cells. As a result, more cells are incorrectly marked as curb.

\subsection{The 3D LiDAR Road-Atlas Results}
To evaluate the whole 3D mapping process, we have evaluated our approach in three real-world urban scenes and two virtual urban scenes (garage and campus). The three real-world urban scenes include our own campus scene and two scenes from the KITTI Odometry datasets \cite{Geiger:2012CVPR}. The two virtual urban scenes 
were parts of the Autonomous Exploration Development Environment \cite{AEDEPA2021}
built by a CMU robotics team. All evaluations are based on a 16-channel LiDAR sensor.

\begin{figure*}[h]
    \centering
    \subfloat{ \includegraphics[width=0.19\textwidth]{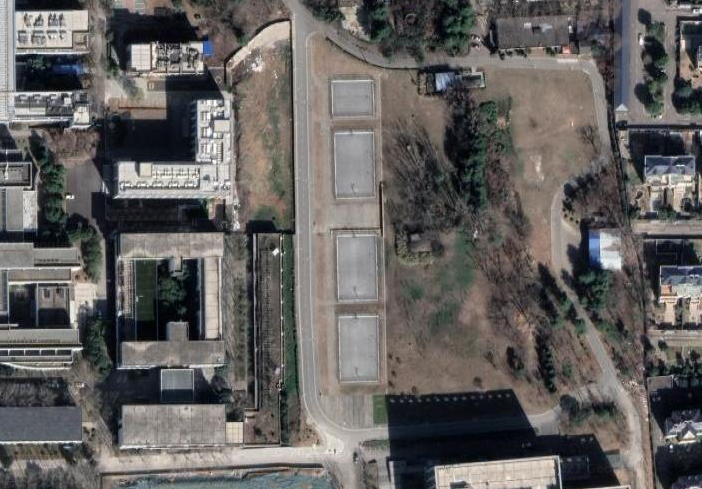} }
    \hskip -4pt
    \subfloat{ \includegraphics[width=0.38\textwidth]{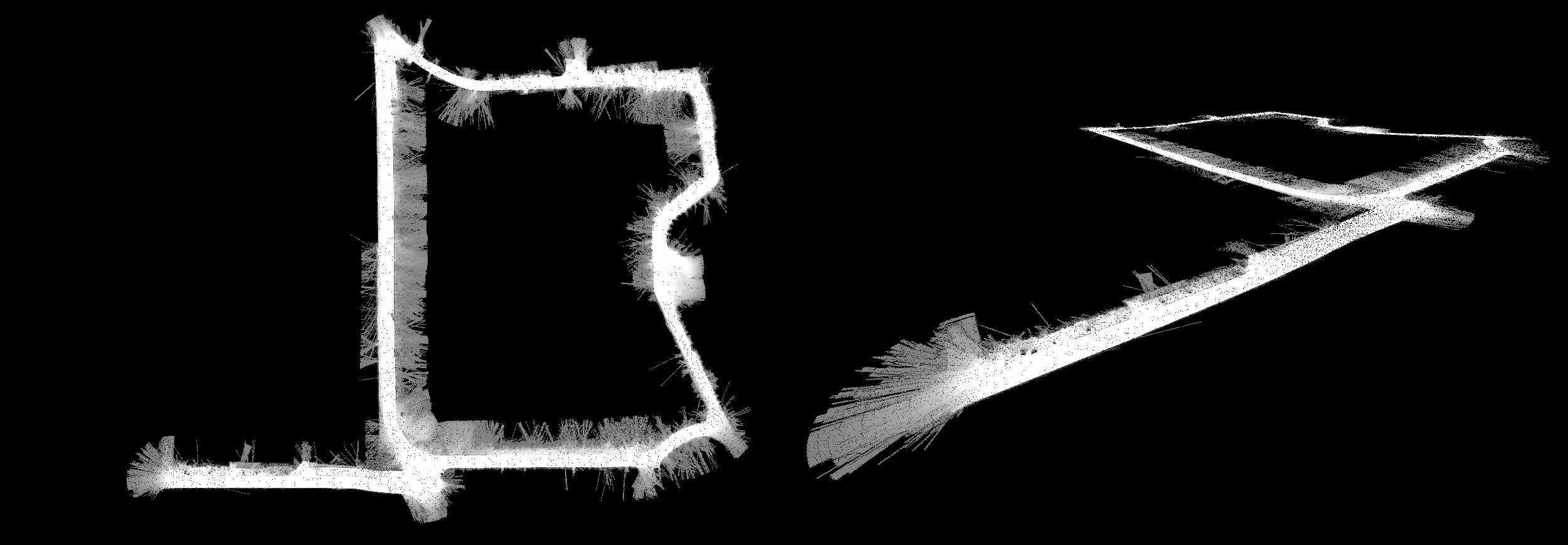} }
    \hskip -4pt
    \subfloat{ \includegraphics[width=0.38\textwidth]{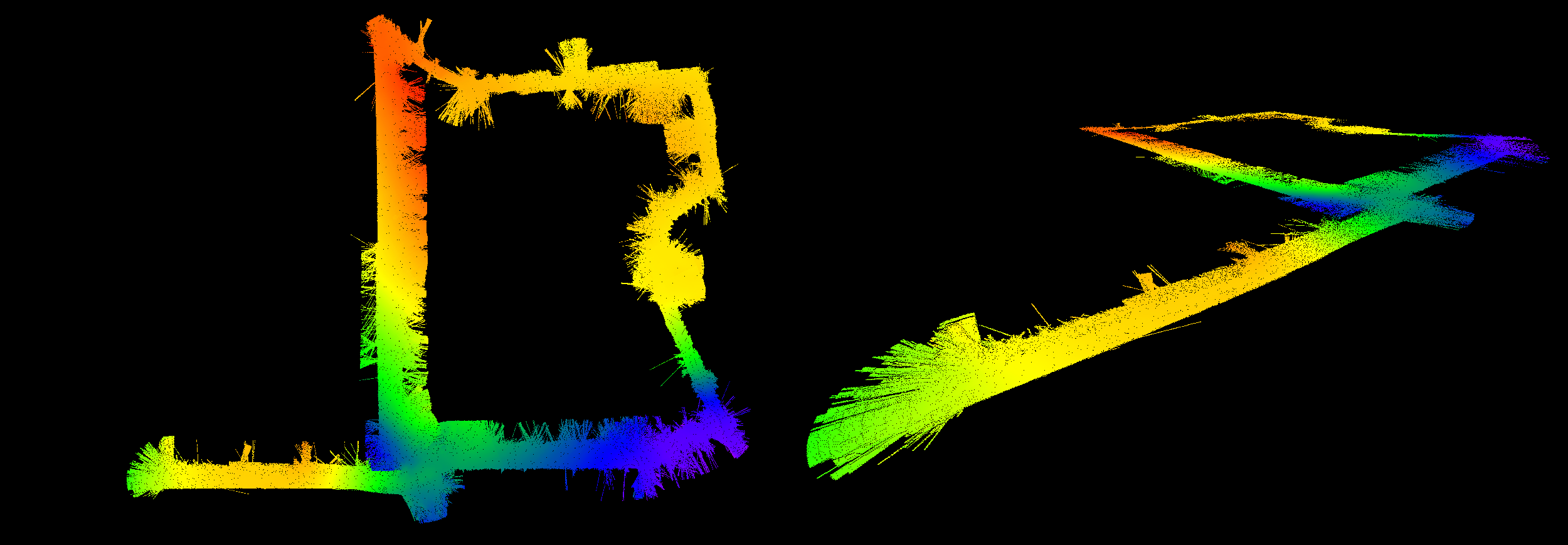} }
    \vskip 2pt
    \subfloat{ \includegraphics[width=0.19\textwidth]{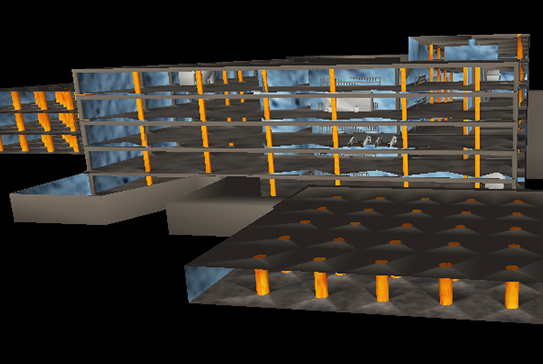} }
    \hskip -4pt
    \subfloat{ \includegraphics[width=0.38\textwidth]{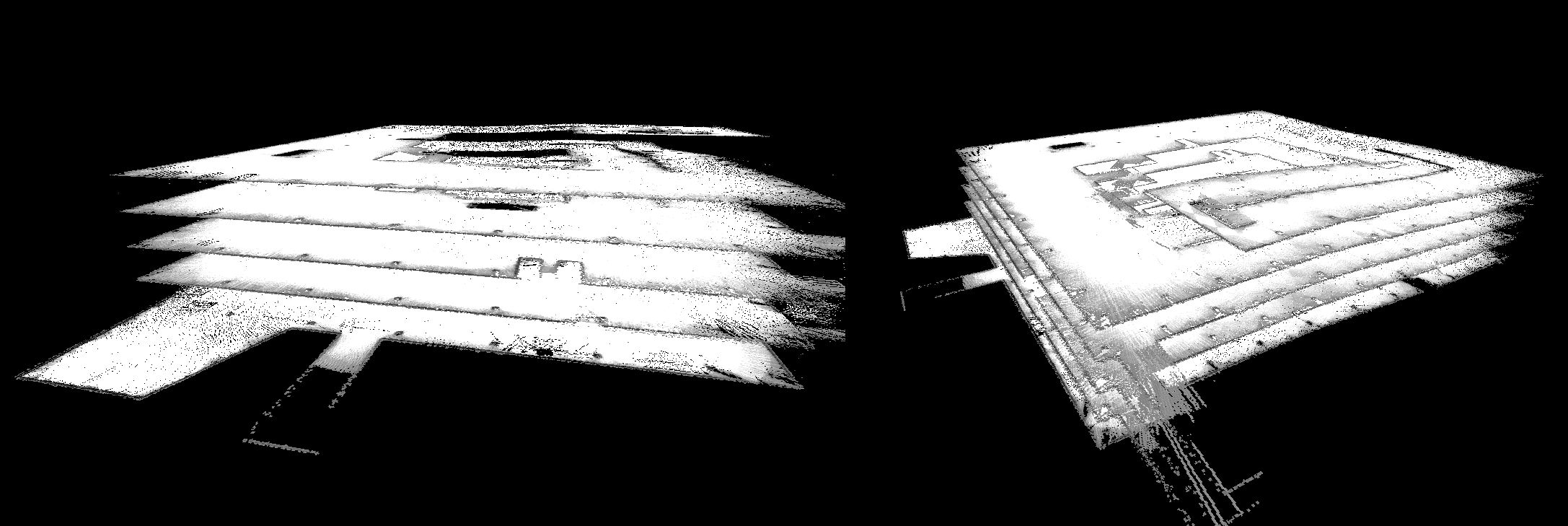} }
    \hskip -4pt
    \subfloat{ \includegraphics[width=0.38\textwidth]{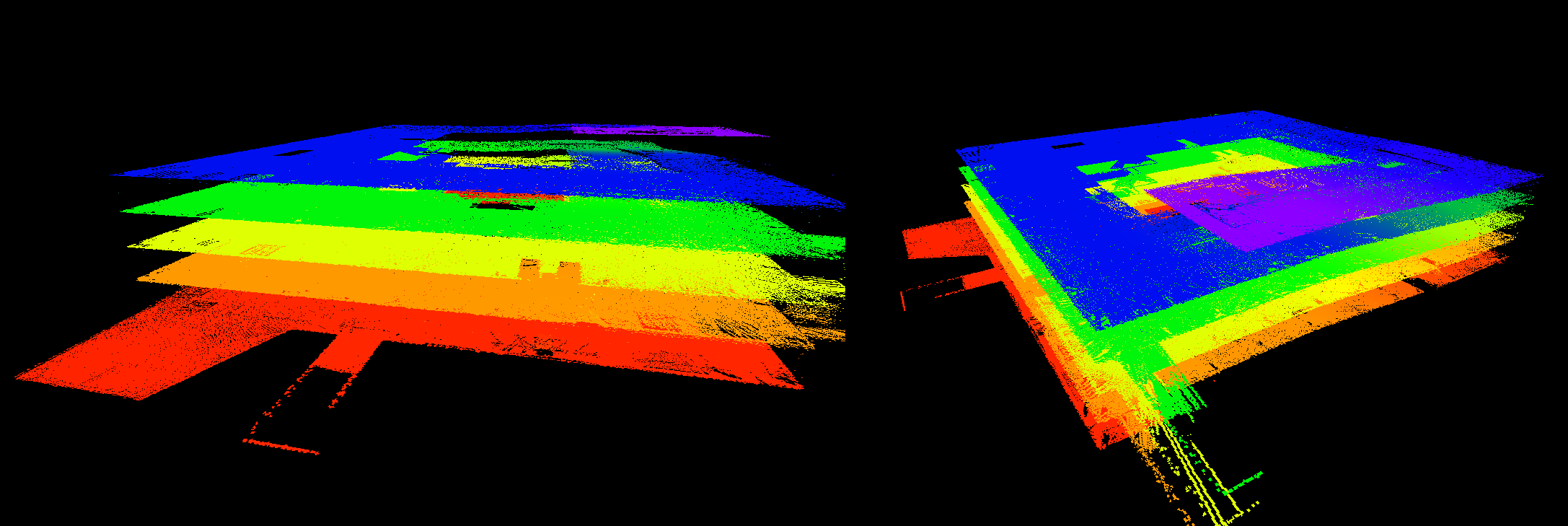} }
    \quad
    \vskip 2pt
    \subfloat{ \includegraphics[width=0.19\textwidth]{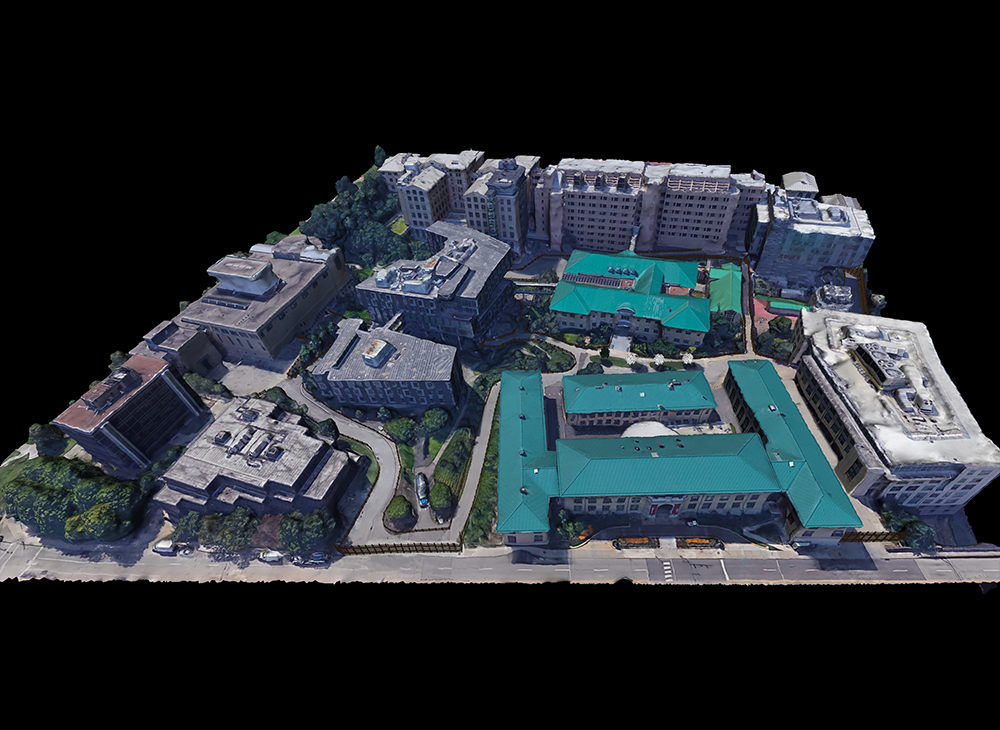} }
    \hskip -4pt
    \subfloat{ \includegraphics[width=0.38\textwidth]{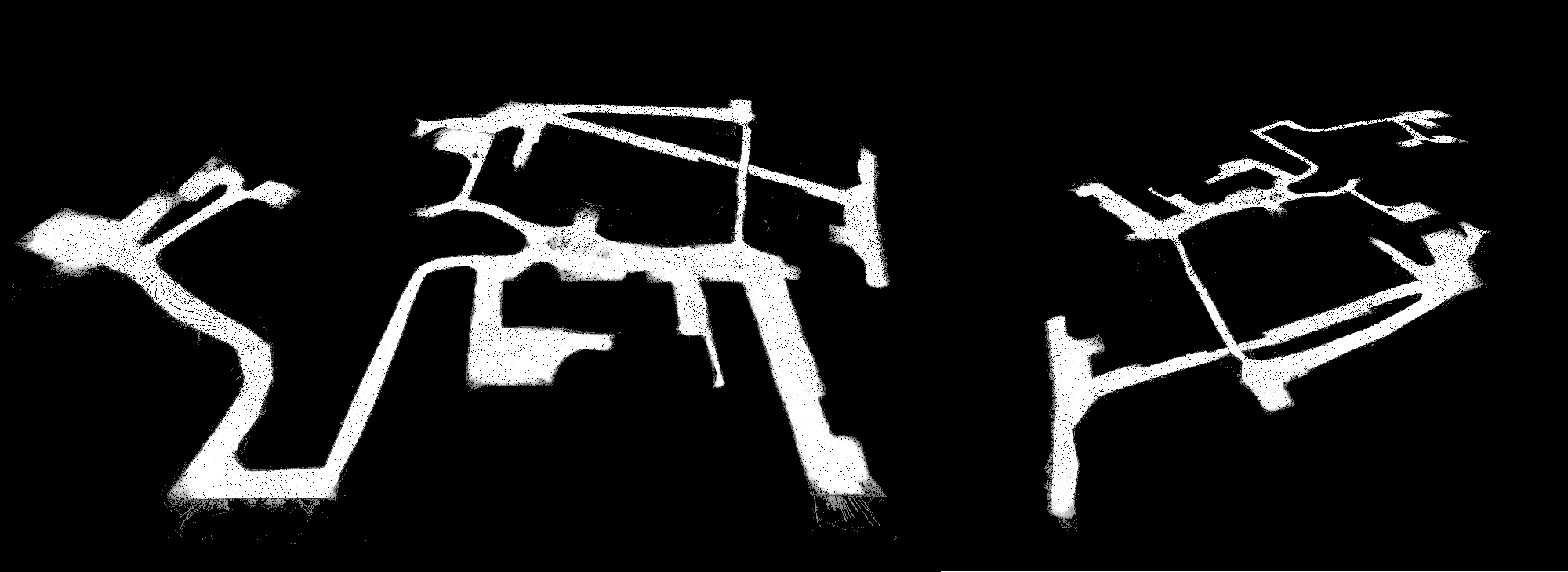} }
    \hskip -4pt
    \subfloat{ \includegraphics[width=0.38\textwidth]{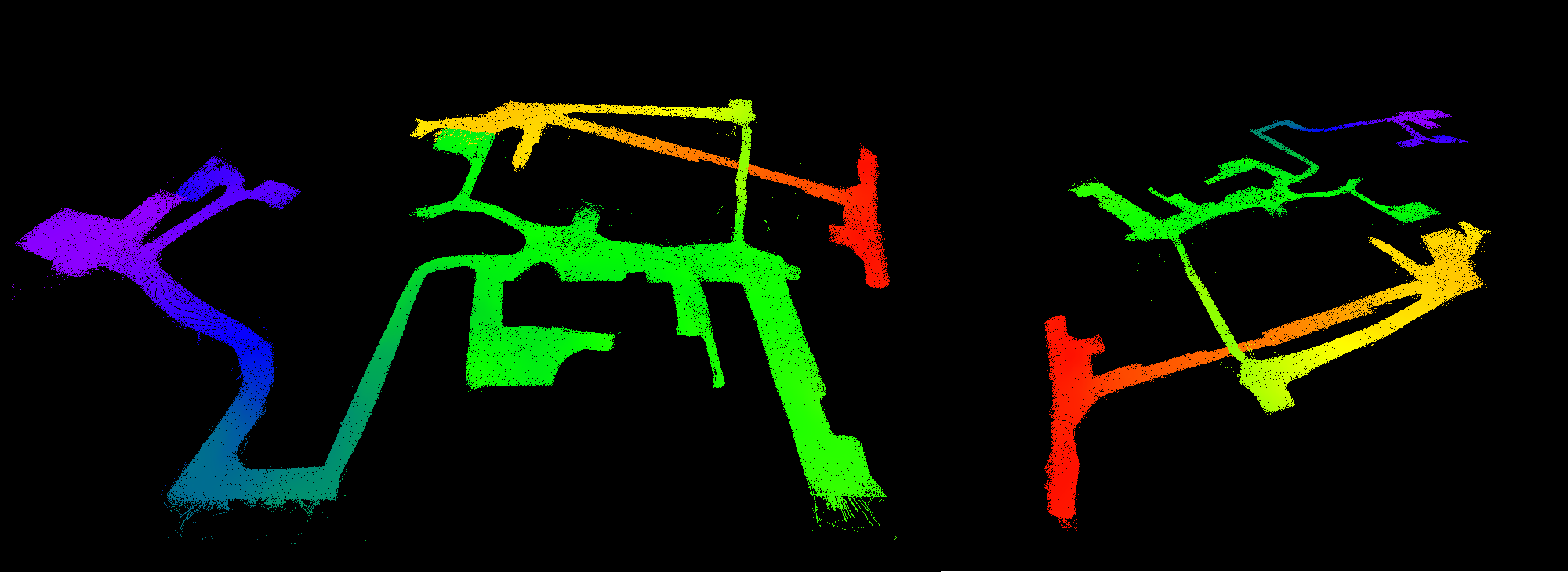} }
    \quad
    \vskip 2pt
    \subfloat{ \includegraphics[width=0.19\textwidth]{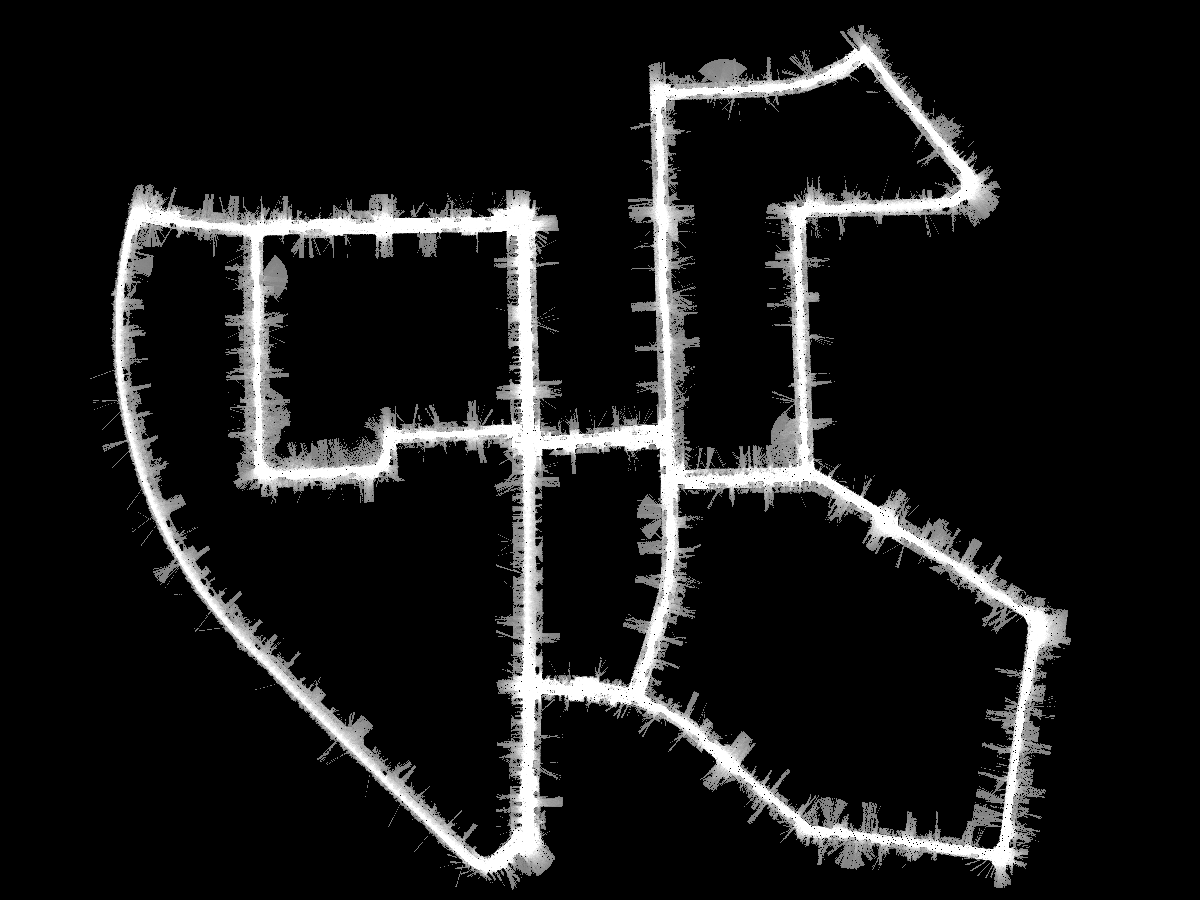} }
    \hskip -4pt
    \subfloat{ \includegraphics[width=0.19\textwidth]{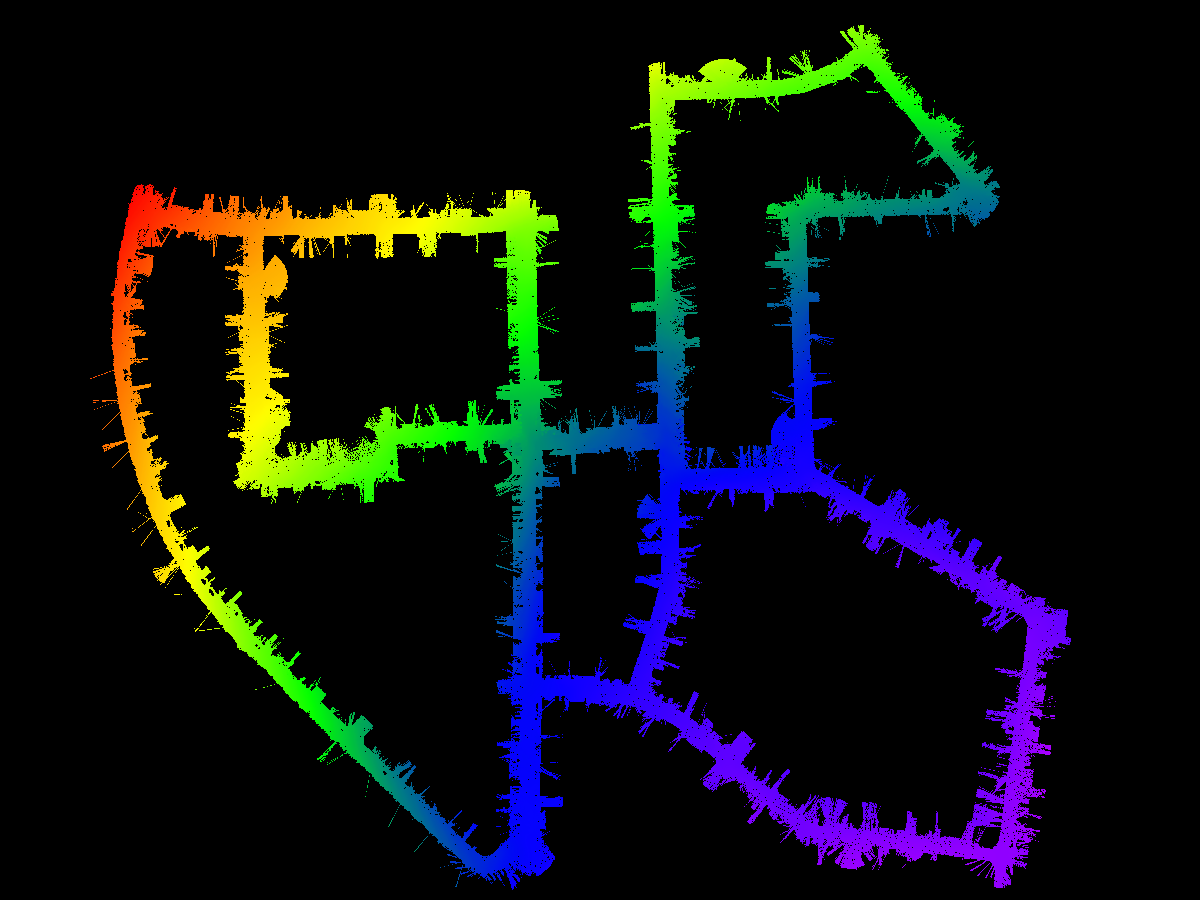} }
    \hskip -4pt
    \subfloat{ \includegraphics[width=0.19\textwidth]{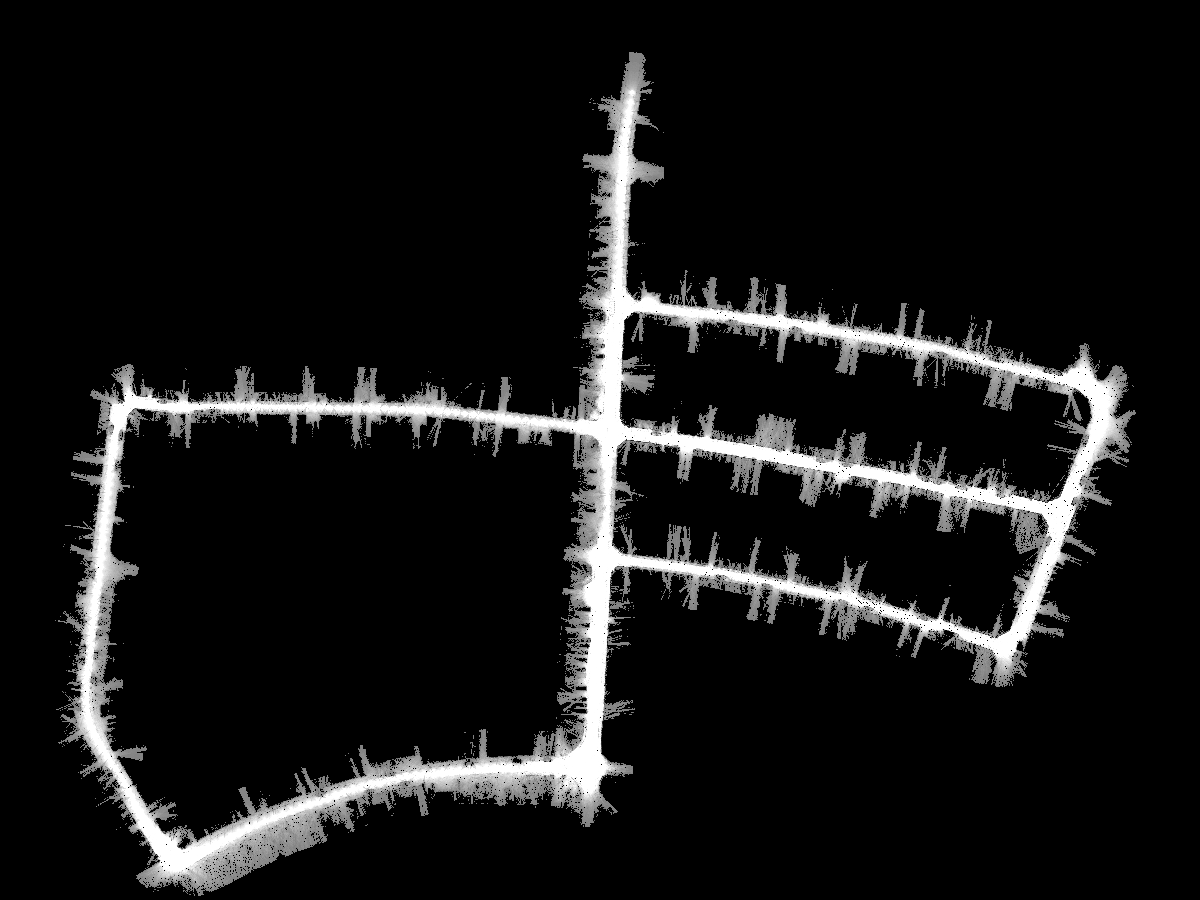} }
    \hskip -4pt
    \subfloat{ \includegraphics[width=0.19\textwidth]{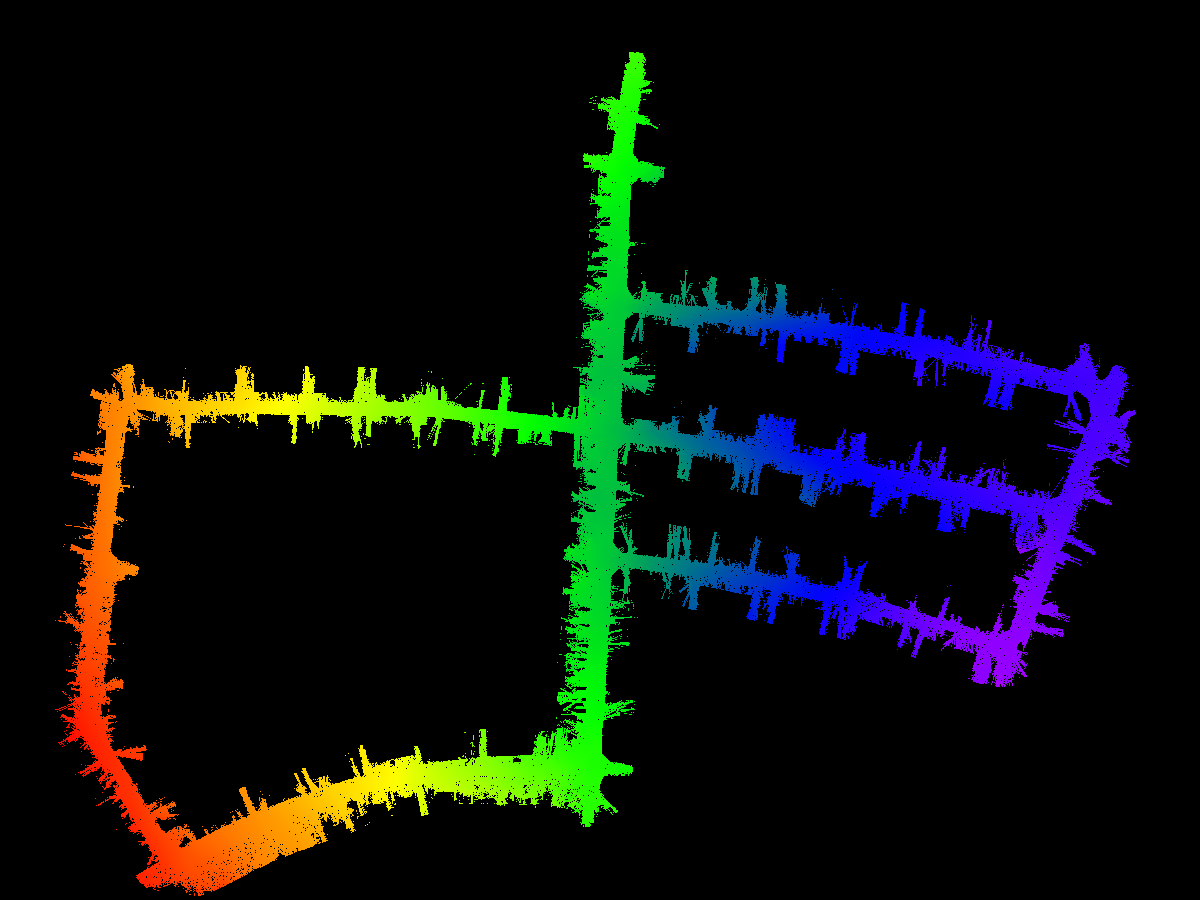} }
    \caption{Global 3D LiDAR-Atlas in five scenes. \textbf{The top three rows}: the first column shows a picture of the whole scene except the two KITTI odometry scenes, for which we do not have the global pictures. The second column shows two views of the created LiDAR road atlas for each scene. To perceive the altitude variation at each location of the created 3D road map, we have encoded the altitude of each cell in gradually changing colors, where red color represents the lowest in altitude, blue representing the highest in altitude, and green and yellow in the middle. The third column shows this effect. \textbf{The last row} shows correspondingly the LiDAR-Atlas resuts for the two KIITI-odometry scenes.}
    \label{fig:all3D}
\end{figure*}

\begin{figure*}
    \centering
    \subfloat{ \includegraphics[width=0.32\textwidth]{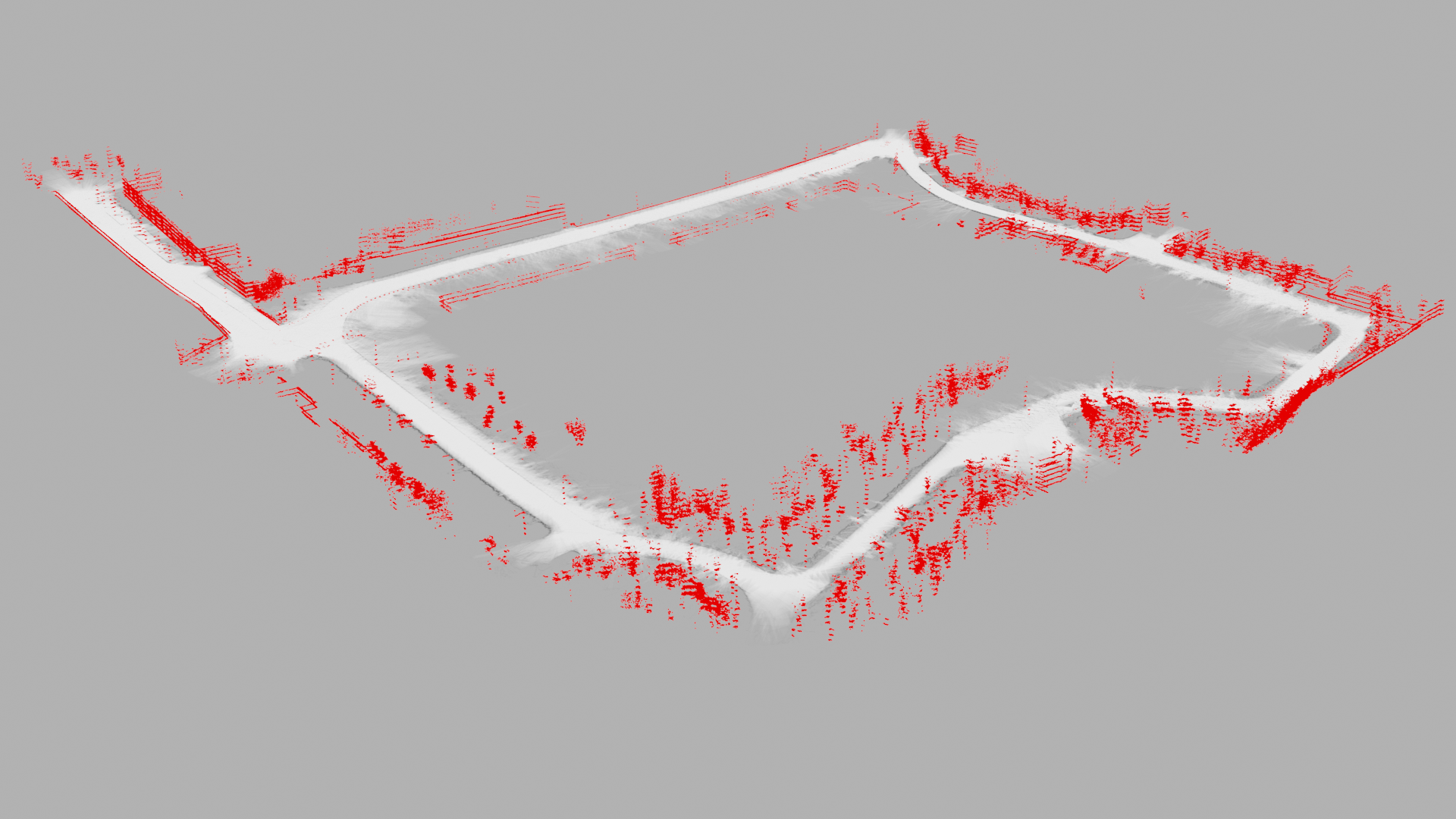} }
    \hskip -4pt
    \subfloat{ \includegraphics[width=0.32\textwidth]{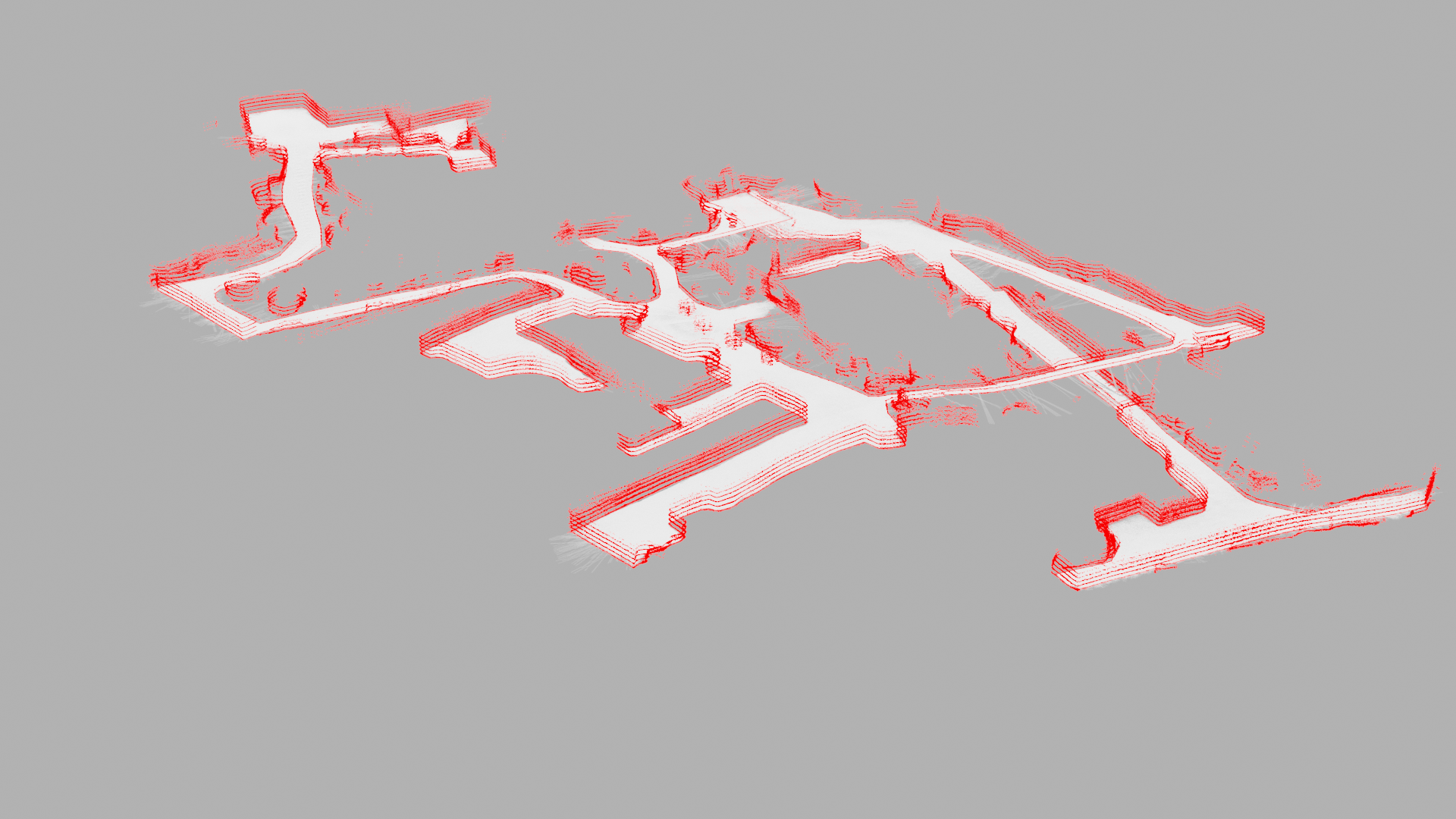} }
    \hskip -4pt
    \subfloat{ \includegraphics[width=0.32\textwidth]{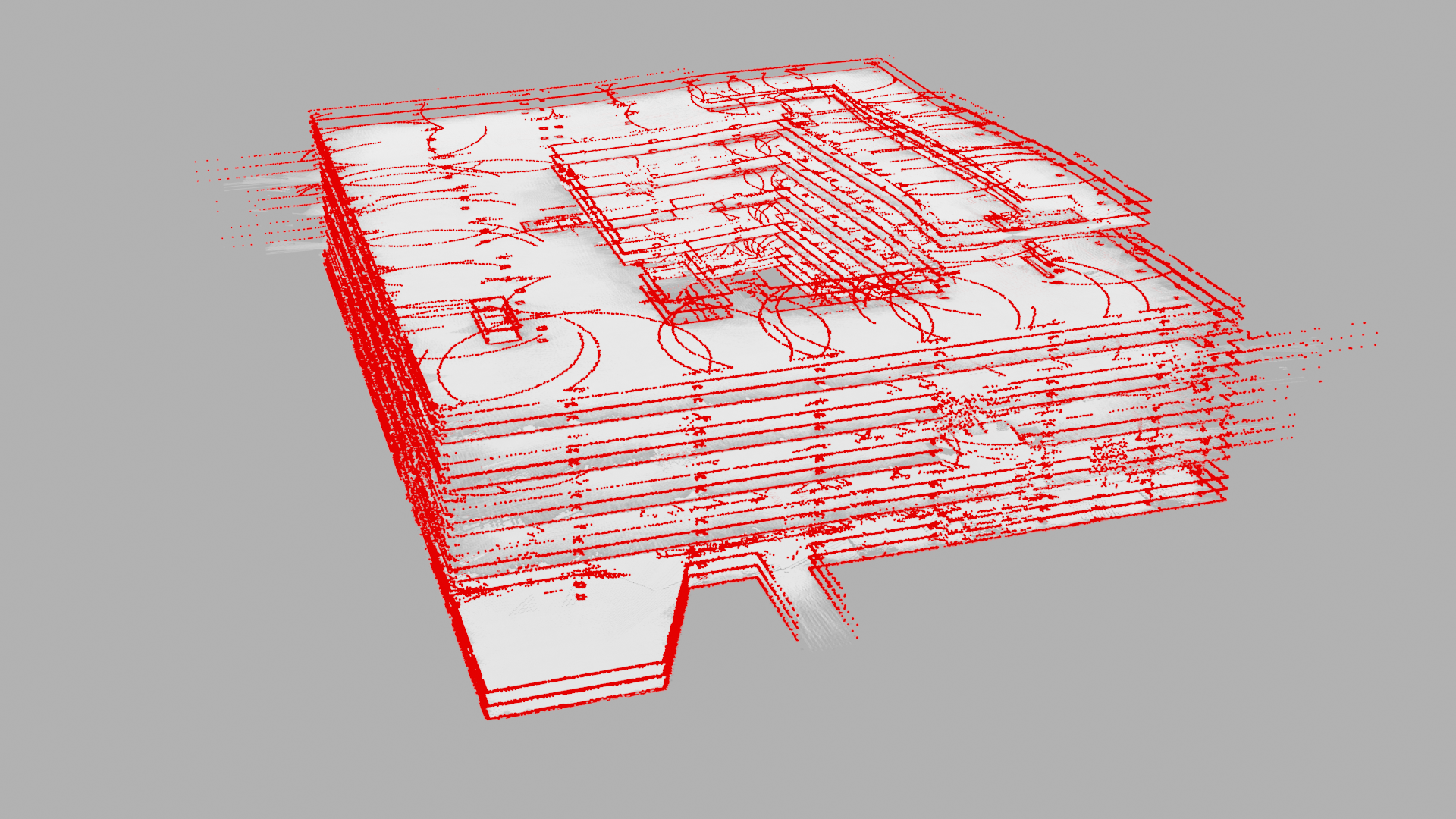} }
    \vskip 2pt
    \hskip -4pt
    \subfloat{ \includegraphics[width=0.32\textwidth]{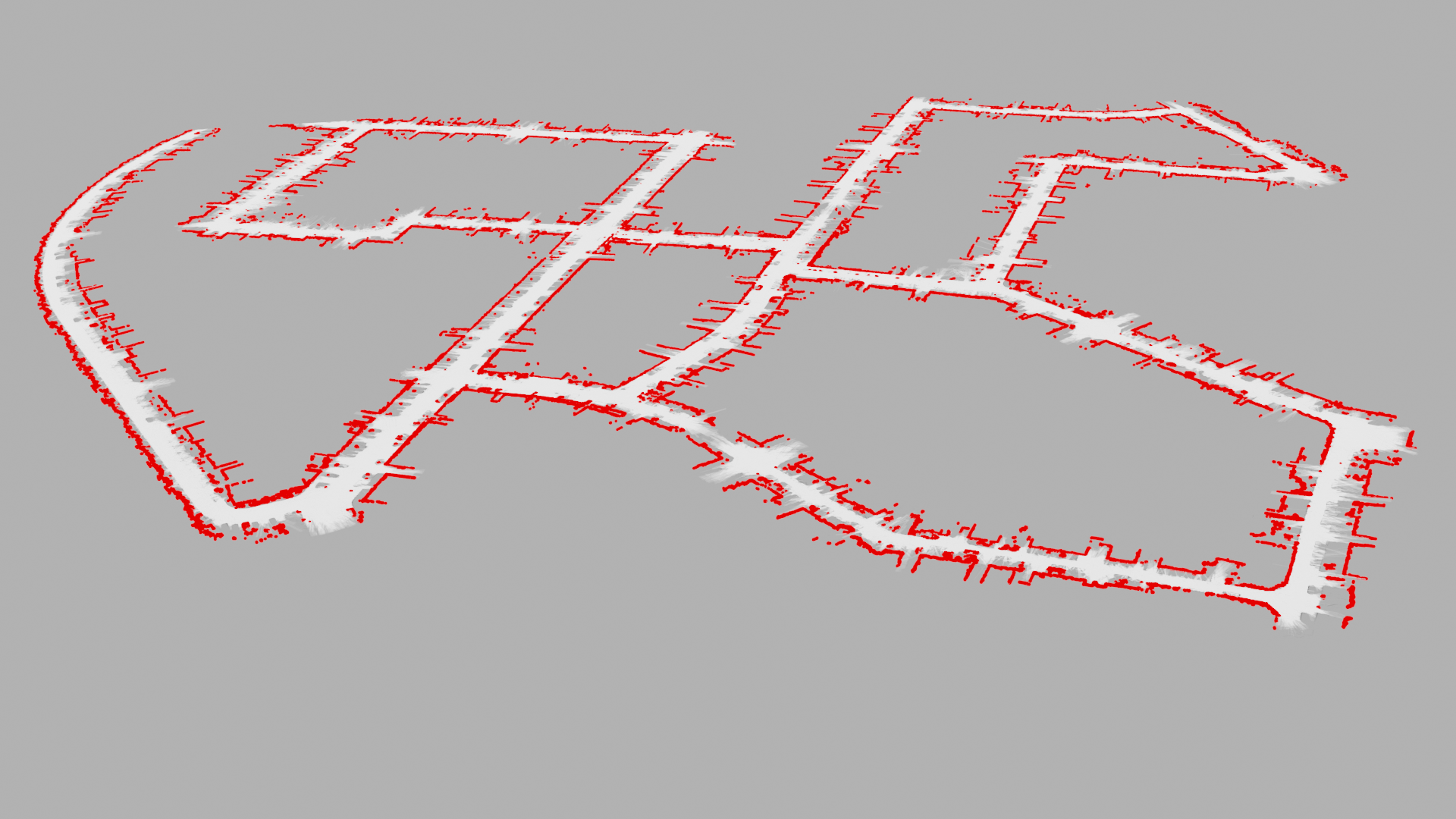} }
    \hskip -4pt
    \subfloat{ \includegraphics[width=0.32\textwidth]{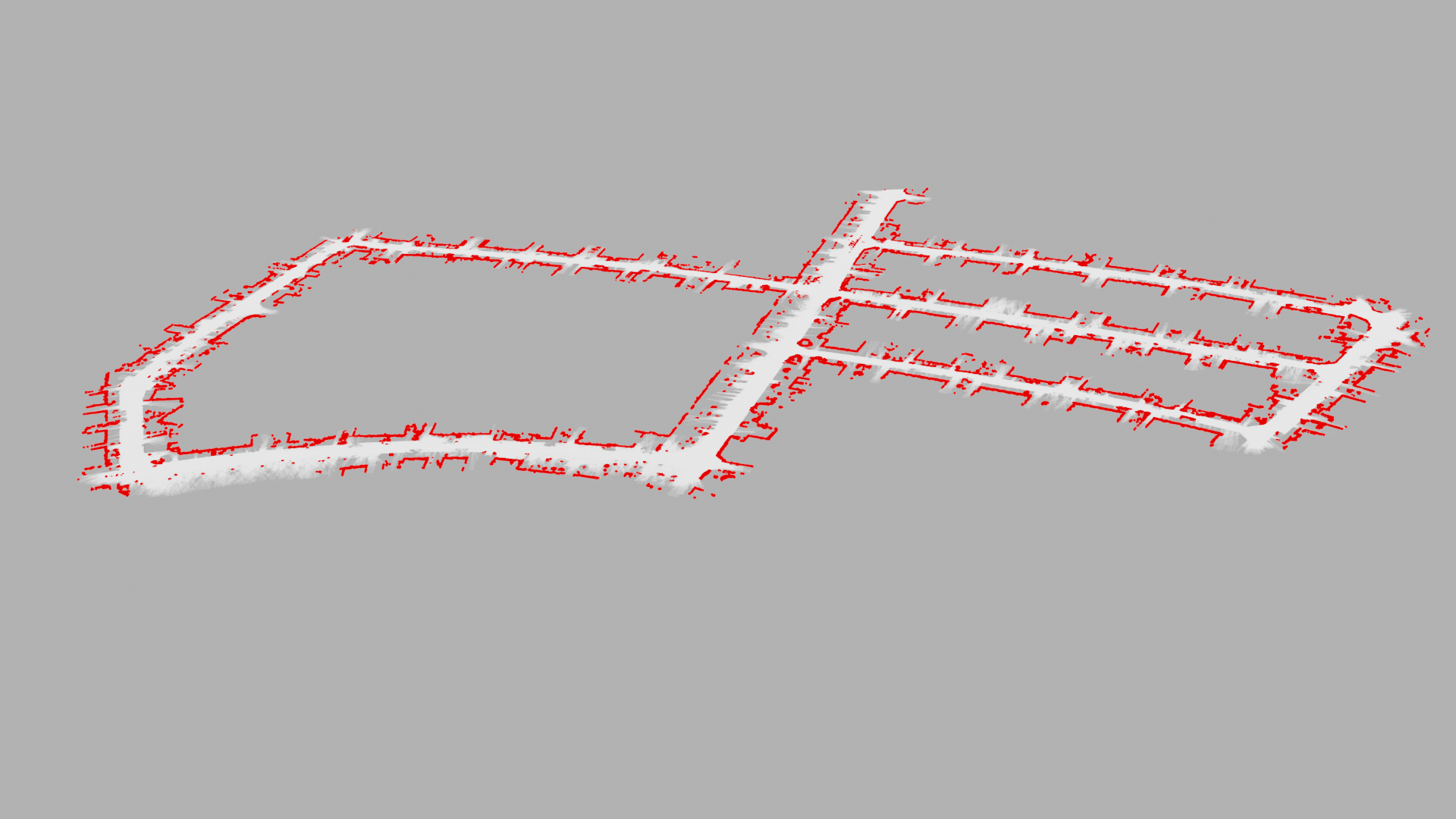} }
    \caption{Show embedded static-obstacle points for vehicle localization results in 5 scenes: our campus, v-campus, v-garage, KITTI-00, KITTI-05. Embedded static-obstacle points for vehicle localization are in red and traversable regions for vehicle navigation are in white.}
    \label{fig:totalmap}
\end{figure*}

In the three real-world urban scenes,  we use the LeGO-LOAM approach \cite{Shan:IROS18} to estimate a relative pose between two consecutive LiDAR frames, and use the LiDAR-Iris method \cite{Wang:IROS20} to provide a stable loop-closure detection. 
Part of the reason for choosing such a set of algorithms is that it can provide good poses in the height axis.
In the two virtual scenes, we use the ground-truth pose estimation provided by the datasets. Regarding the selection of key-frames, in the three real world scenes, we directly use the key-frames of SLAM as our key-frames (because only the key-frames of SLAM can get more accurate poses), in the virtual scene, we use 1Hz frequency to extract key-frames. In Fig.\ref{fig:all3D} and Fig.\ref{fig:totalmap}, we show the obtained 3D LiDAR road atlas without embedded vertical structures and the 3D LiDAR road atlas with embedded vertical structures, respectively. 

Note that although the tested real-world scenes contain non-flat road scenarios, there are no multiple-layer road infrastructures, such as underpass and overpass. 
In contrast, the simulated garage and campus scenes provided by the Autonomous Exploration Development Environment \cite{AEDEPA2021} contain such multiple-layer road or traversable-region scenes. The obtained LiDAR Road-Atlas maps from the two simulation environments have shown that our method is effective in dealing with such scenes. 
In implementation of traversable and curb detection in the simulation environment, due to the large horizontal resolution of the LiDAR in the simulation environment (350 points per circle), we adjusted the width of LiDAR imagery to 350, and the width of the local window for detecting traversable areas and obstacles to 9 (also about $10^{\circ}$). 

\subsection{Memory usage and map size}

The total memory consumption of our method includes two major parts, i.e., the cost for the created 3D road map (e.g., the results shown in Fig.\ref{fig:all3D}) and the space required for saving the encoded vertical structure of each cell.
As in the occupancy elevation map \cite{Souza:Robotica16}, our 3D road map also takes the advantage of the probabilistic fusion when merging the 2D local occupancy grid maps. 
Generally, it can save a lot of memory compared to the other 3D occupancy grid maps, e.g., the OctoMap method \cite{Hornung:AutoRobot13}. To make a comparison with the OctoMap method in terms of memory and hard-disk usage, we have recorded the file size and memory usage by our method and the OctoMap method for different scenes, shown in Table \ref{table:mem}. The input data of the two methods are exactly the same, and the cell or grid resolution equals to $0.1m$.

\begin{table*}[h]
\caption{Runtime memory and hard-drive (H-D) storage usage of our map}
\label{table:mem}
\centering
\begin{threeparttable}
\begin{tabular}{lrrrrrr}
\toprule
                & & \begin{tabular}[c]{@{}l@{}}our\\ campus\end{tabular} & v-campus & v-garage & KITTI00 & KITTI05 \\
\midrule

\multirowcell{2}{OctoMap radius=30m}  & runtime(GB\tnote{*}) & 4.25   & 3.53   & 2.43   & NA\tnote{*} & NA\\
                                                                   & H-D(MB\tnote{*})      & 371.71 & 295.53 & 222.46 &    & \\
\multirowcell{2}{OctoMap radius=20m}  & runtime(GB) & 3.55   & 3.09   & 2.36   & NA & NA\\
                                        & H-D(MB)      & 306.96 & 253.78 & 214.66 &    & \\
\midrule
                                                          
                                                          
                                                         

\multirowcell{2}{our method with radius=30m,\\ 32 bits obstacle descriptor\tnote{*}} & runtime(GB)   & 8.69  & 5.71   & 6.44  & 10.32  & 6.09   \\
                                                                            & H-D(MB)       & 49.77 & 52.51  & 41.50 & 178.18 & 120.59 \\
                                                          
\multirowcell{2}{our method with radius=20m,\\ 32 bits obstacle descriptor} & runtime(GB)   & 3.95  & 2.02   & 3.28  & 4.69   & 2.79   \\
                                                                            & H-D(MB)       & 39.57 & 47.96  & 40.93 & 148.27 & 96.37  \\
                                                          
\midrule
\multirowcell{2}{our method with radius=30m,\\ 8 bits obstacle descriptor\tnote{*}}  & runtime(GB)   & 6.06  & 4.34   & 4.84  & 7.10   & 4.51   \\
                                                                            & H-D(MB)       & 41.06 & 43.27  & 34.44 & 146.97 & 99.46  \\
                                                         
\multirowcell{2}{our method with radius=20m,\\ 8 bits obstacle descriptor}  & runtime(GB)   & 3.05  & 1.51   & 2.70  & 3.81   & 2.19   \\
                                                                            & H-D(MB)       & 32.64 & 39.56  & 33.83 & 122.30 & 79.49  \\
                                                         
\midrule
LeGO-LOAM\tnote{*} & H-D(MB) & 20.08 & & & 72.78 & 45.67 \\
\bottomrule
\end{tabular}
\begin{tablenotes}
\footnotesize
\item[*] Due to immense memory usage and very long update time, OctoMap cannot complete the mapping of the KITTI dataset.
\item[*] The file size of LeGO-LOAM refers to the sum of corner point cloud and surface point cloud.
\item[*] GB: Giga Bytes. MB: Mega Bytes. 
\item[*] The number of bits for obstacle (vertical-structure) descriptor here is equal to the product of the number of segments and the number of bits per segment.
\end{tablenotes}
\end{threeparttable}
\end{table*}

In the table, the ``runtime'' represents the total runtime memory cost till the end of the whole online mapping process. The ``H-D'' represents the size of the file after the map is saved as a file in hard drive. In our implementation, the runtime memory
is dependent on several major factors, i.e., the scanning range of LiDAR sensor, the cell resolution (representing the actual size in 3D world per cell),  the number of vertical segments which are used for encoding the vertical structure within each cell (determining the height limit of the point cloud within each cell), the number of bits per segment and the number of collected keyframes during traveling unit distance (e.g., 1 meter in our work). 

To make a fair comparison with the OctoMap representation, the cell resolution and the number of collected keyframes across each whole dataset are set to the same for both our method and the OctoMap algorithm \cite{Hornung:AutoRobot13}. Therefore, the rest parameters that can be adjusted for determining the ``runtime'' and ``H-D'' values of our method is the scanning range of LiDAR sensor and the numbers of bits used to encode the vertical structure  within each cell. 


In terms of the scan range of LiDAR sensor, it can affect the ``runtime'' memory of our method significantly. In our 3D road map, we 
temporarily stores the local 2D occupancy grid maps as the intermediate results in order to allow for updating poses of frames (such as the pose optimization when loop closure happens). In general, the size of a local 2D-OGM depends mostly on the LiDAR's sensing range and more memory consumption is necessary as the sensing range becomes larger.
Normally this value can be set as the range limit of a LiDAR senor. For example, the VLP-16's scan-range limit can be up to 100 meters. In this case, the size of a local 2D-OGM is 2000x2000 given the cell resolution of 0.1m/cell, which  means that there are about 4x10$^{6}$ cells in each local 2D-OGM. 

Empirically, the point cloud 60 meters away is generally quite sparse, meaning that it is not necessary to use the whole frame of point cloud for building map. 
Considering these factors, it is preferable to reduce the size of local 2D-OGM by only utilizing point cloud within a certain small range. For example, we set the radius of LiDAR's scan range to be 20m and 30m, respectively, in our experiments. Additionally, we can further reduce the memory cost on the descriptor of encoding vertical structure of each cell from four bytes to one byte only. Accordingly, the runtime memory usage of our method with different settings of mapping parameters are shown in Table \ref{table:mem}. 

When only using point cloud within a scan range of 20m for building 2D-OGM and use 1-byte descriptors, the runtime memory usage in all environments is smaller than that of OctoMap except for the virtual garage environment. 
We can also observe a significant reduction of runtime memory by using 1-byte descriptors in contrast to using 4-byte descriptors, where we observe a reduction of about 23$\%$, 25$\%$, and 17$\%$ on runtime memory cost in our campus, the v-campus and v-garage scenes, respectively. 
It deserves pointing out that the garage environment is a multi-layer parking-lot environment (as shown in the Figure \ref{fig:all3D}) and the trajectory is more concentrated within a small space. For such kind of scenarios, the memory usage of OctoMap is smaller. In contrast, our method needs to build the 3D road atlas by tracing a quite long trajectory within the same area repeatedly. Thus, the intermediate results (the local 2D-OGM) can occupy lots of memory, which results in a larger runtime memory consumption than the OctoMap. 
When using point cloud within the scan range of 30m for building 2D-OGM, the runtime memory usage of our method is generally more than that of the OctoMap method. This is still caused by the memory coverage of the intermediate results which are prepared for pose correction during loop-closure process. 
But we also notice that the OctoMap method will crash when applied to very large scenes, such as KITTI00 and KITTI05 odometry benchmark datasets, due to immense memory cost and extremely long map update time. In contrast, our method can still be applicable to such kind of large scenes.

In terms of the hard-drive storage space, it can be seen from Table \ref{table:mem} that our method saves a lot of more storage space than the OctoMap, no matter whether in the simulated environments or in real scenes. We also observe that the hard-drive storage of the map by our method does not change much when using either 4-byte or 1-byte descriptors for representing vertical structure geometry within each map cell. For all the results shown in Table \ref{table:mem}, a 16-channel LiDAR (or simulated 16-channel LiDAR) sensor is used, where the point cloud within a radius of 30m is used for building local 2D-OGM with only 1-byte descriptor for encoding vertical structure of each cell. 
Because it is the completed map that will be used in the ``repeat'' stage of the ``teach and repeat'' paradigm, considering the small hard-drive storage space of our method, our map representation is more intriguing than the existing 3D LiDAR maps, such as the OctoMap, when applied to a large-scale environment.

We also compare our method with the LeGO-LOAM approach \cite{Shan:IROS18} in terms of the hard-drive storage size of the resulting map file. In the LeGO-LOAM method, only points with high curvature or points within large plannar regions (corresponding to walls or road) are reserved in the resulting map. On three compared datasets, the ``H-D'' values of the LeGO-LOAM method is smaller than ours, but the point-cloud map obtained by the LeGO-LOAM method alone cannot be used for navigation because traversable region is not provided in the map. In contrast, the LiDAR Road-Atlas obtained by our method is a complete map which can favorably for navigation purposes alone.



\subsection{Online map update}
\begin{figure}[h]
    \centering
    
    \subfloat{ \includegraphics[width=0.28\textwidth]{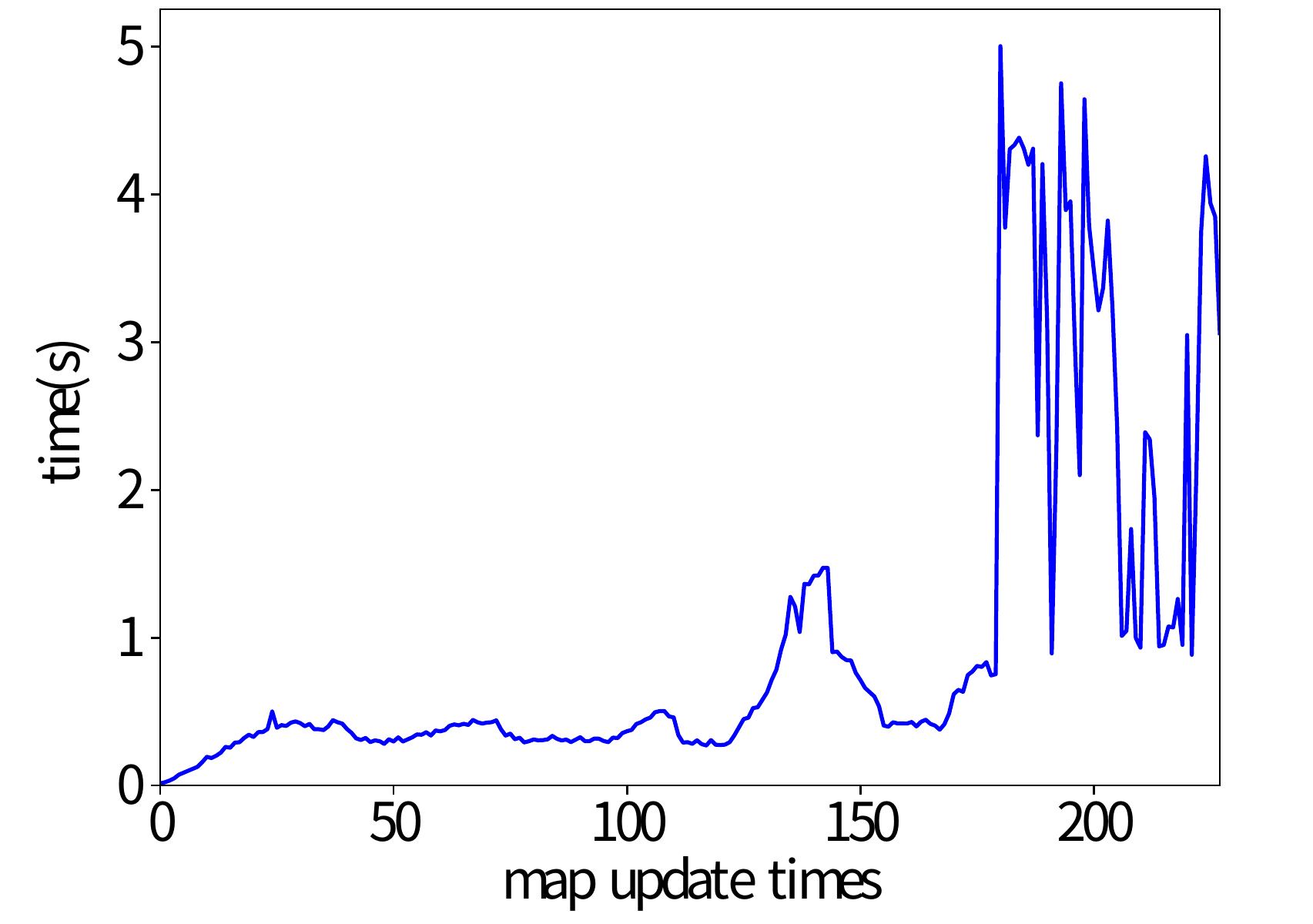}
    \label{update-real-30}}
    \subfloat{ \includegraphics[width=0.28\textwidth]{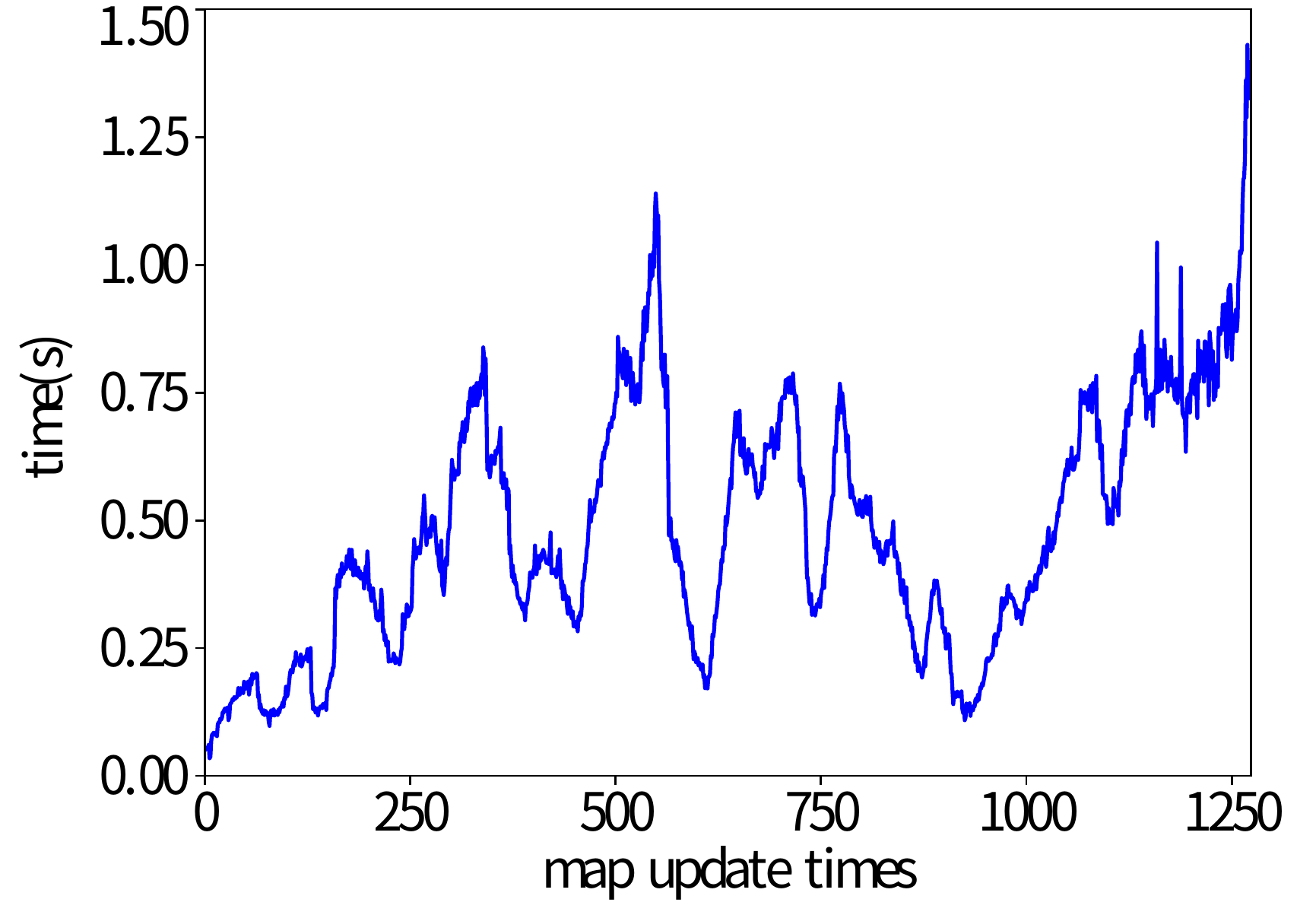} \label{update-campus-30}}
    \subfloat{ \includegraphics[width=0.28\textwidth]{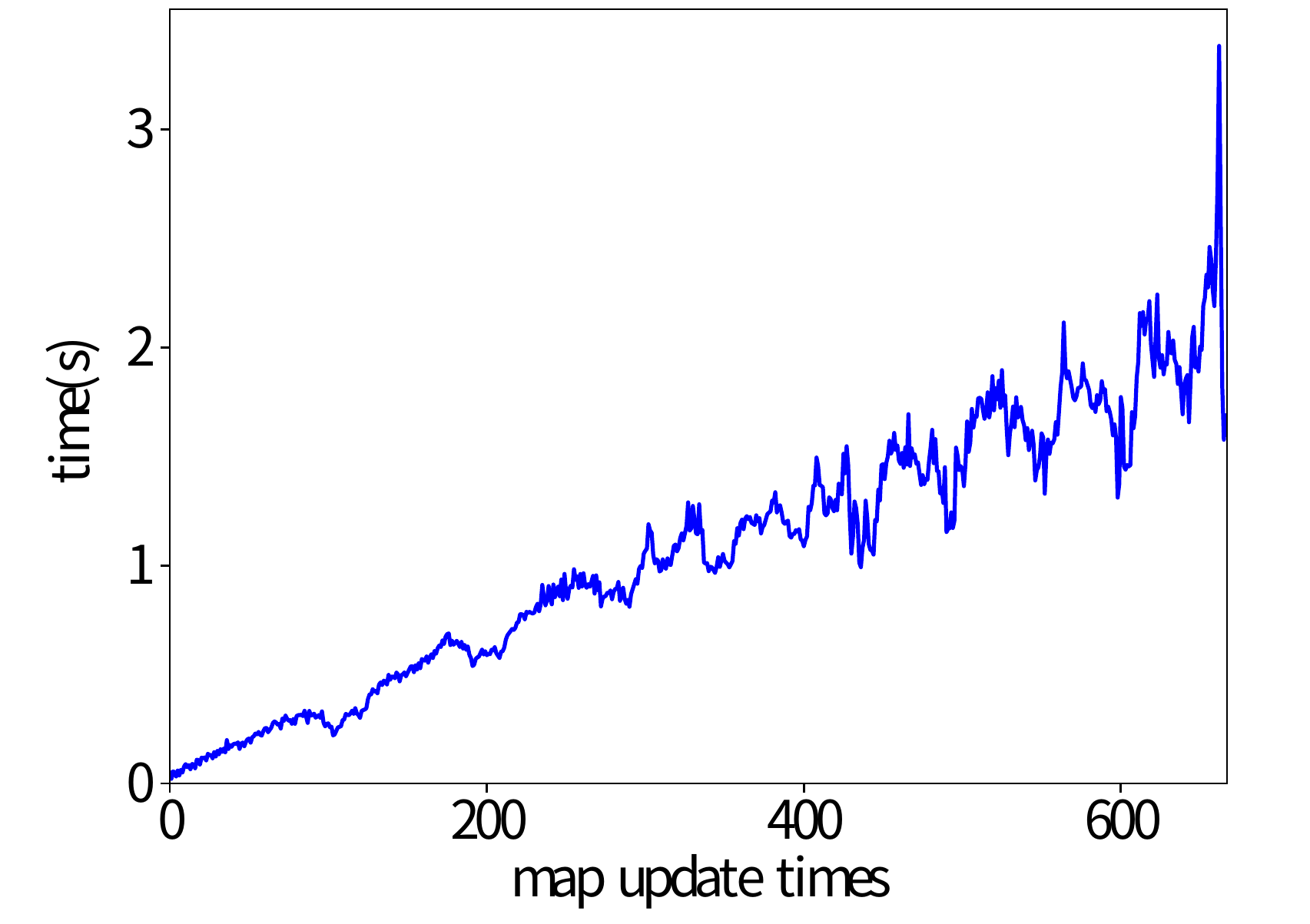} \label{update-garage-30}}
    \quad
    \subfloat{ \includegraphics[width=0.28\textwidth]{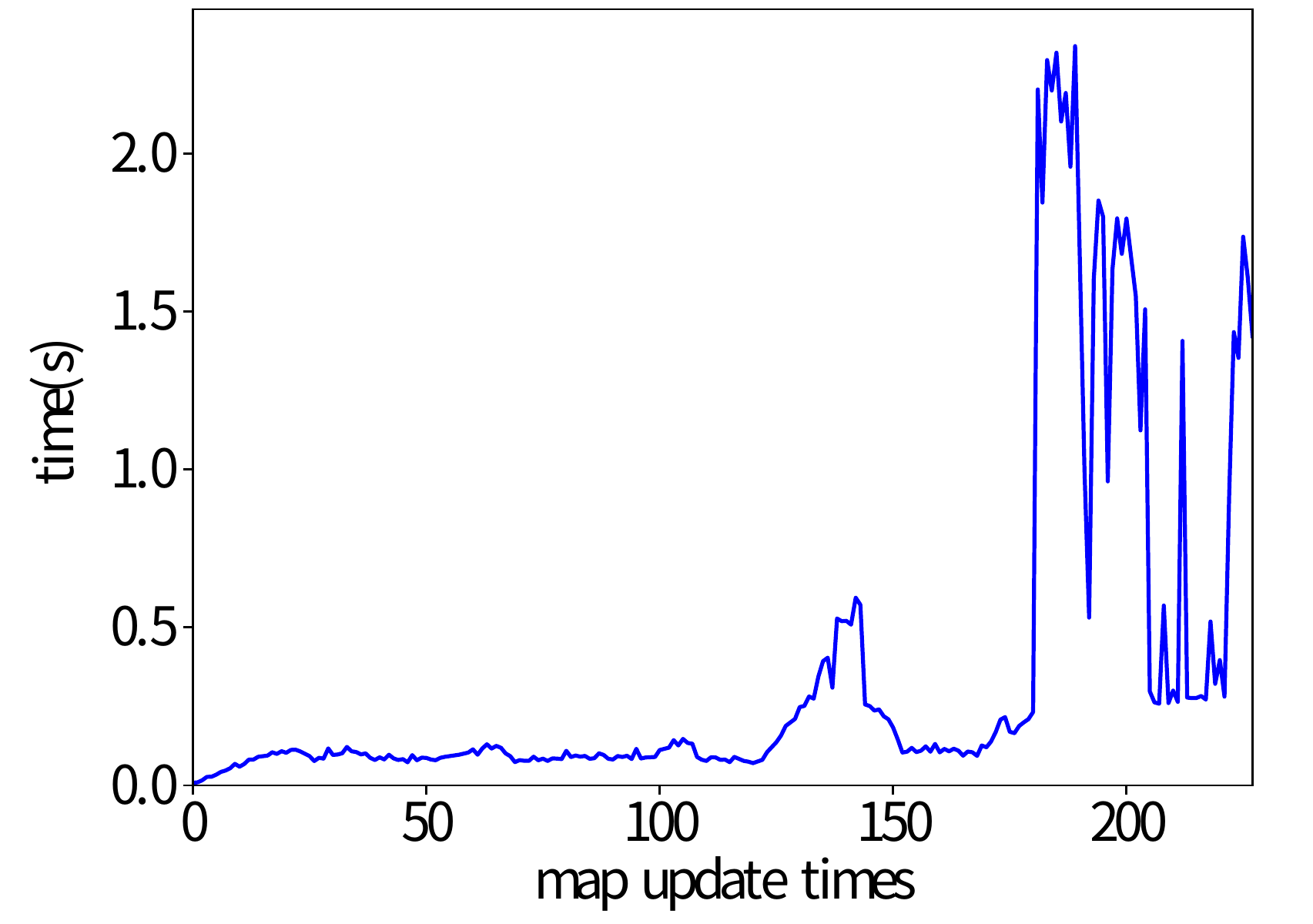} \label{update-real-20}}
    \subfloat{ \includegraphics[width=0.28\textwidth]{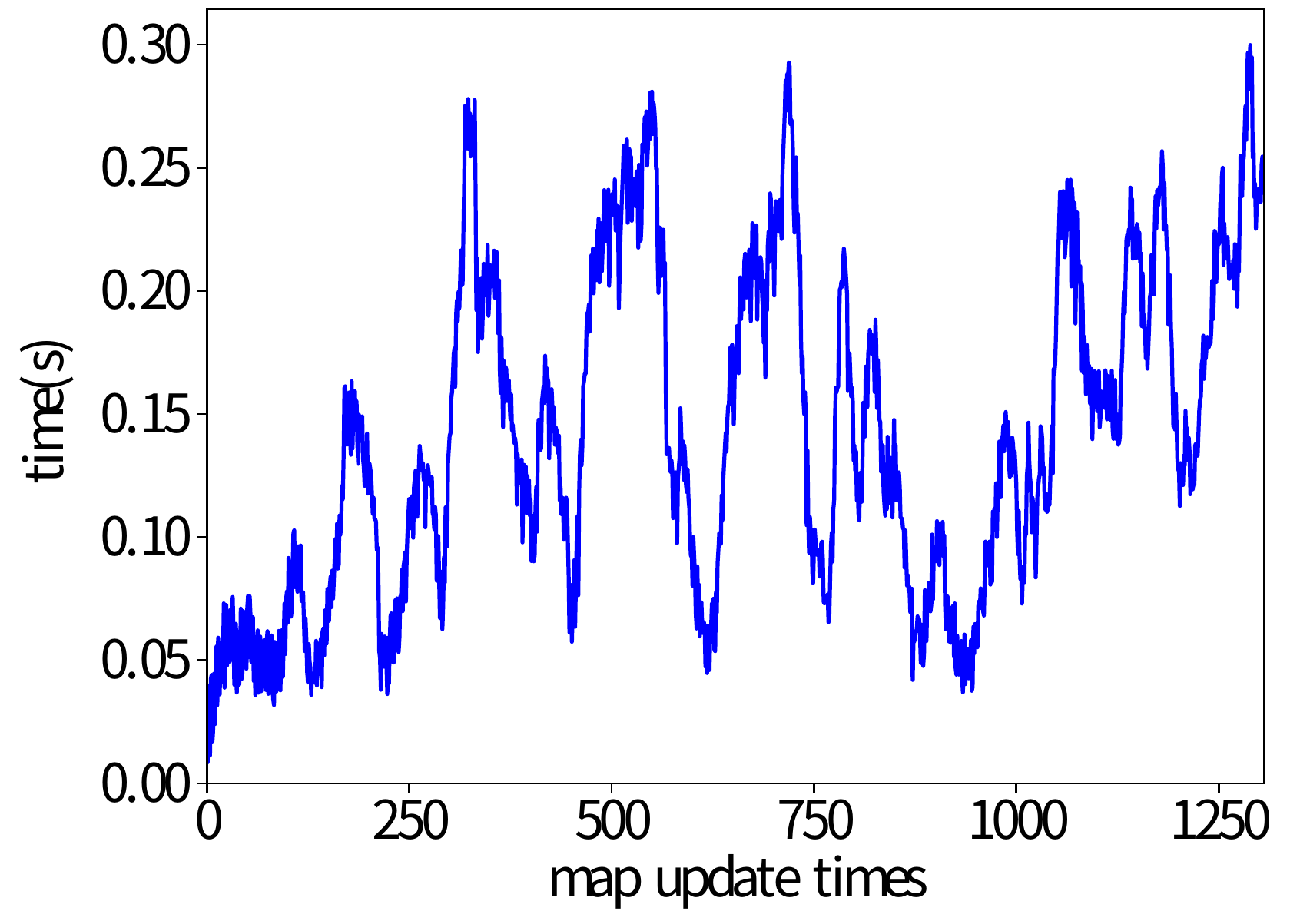} \label{update-campus-20}}
    \subfloat{ \includegraphics[width=0.28\textwidth]{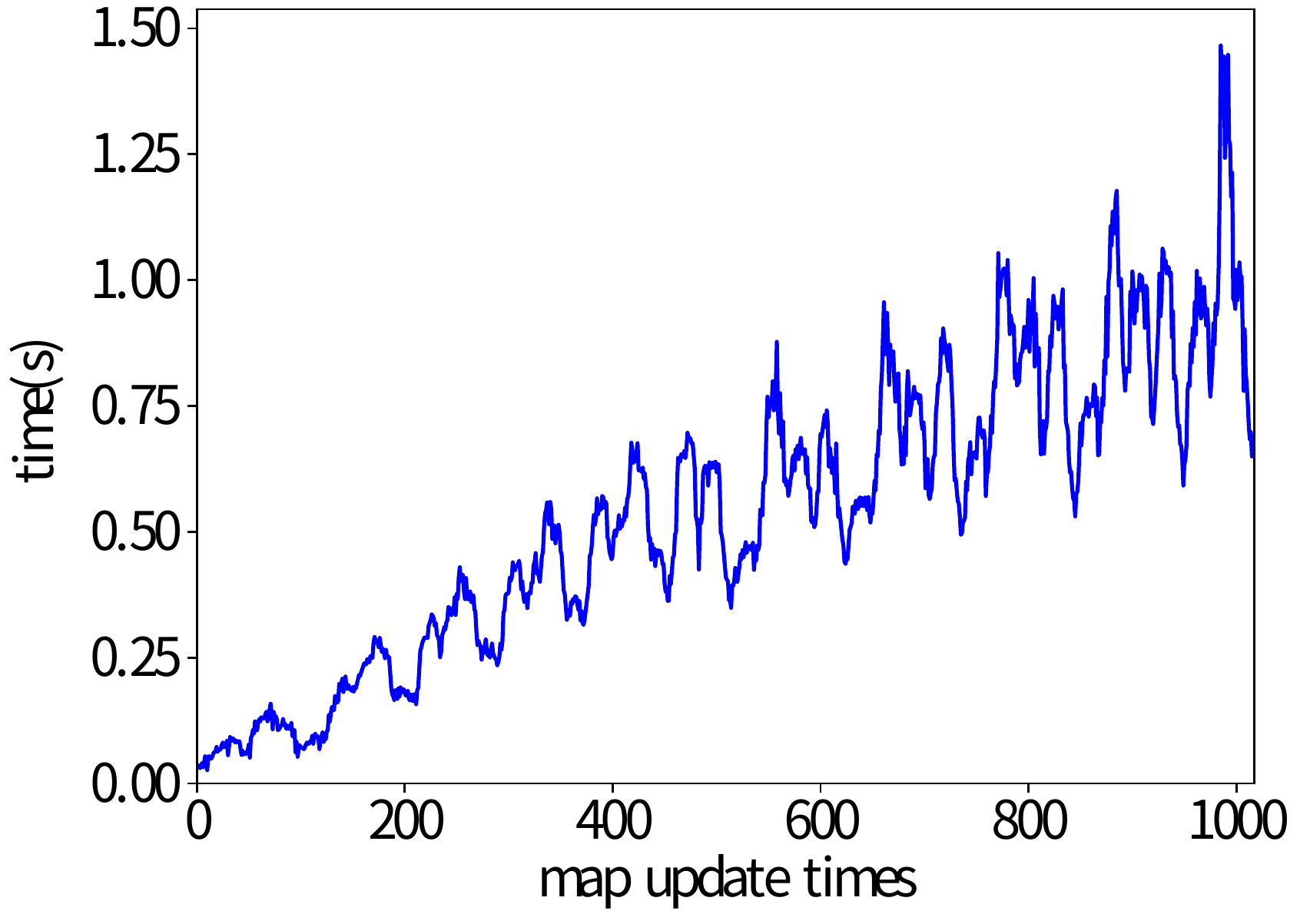} \label{update-garage-20}}
    \caption{Time consumption for map update in three scenes. The \textbf{top row} corresponds to the map update results when using point-cloud data within the radius of 30m for mapping, and the \textbf{bottom row} corresponds to the results of using point-cloud data within the radius of 20m for mapping. 
    The \textbf{left column} shows the results of our campus scene, where there are a short reverse loop-closure in the middle and a long loop-closure in the last part of the course. Therefore, we can observe an obvious increase of time cost in the middle, and a sharp increase in time cost in the last part. 
    The \textbf{middle column} shows the map update results of the virtual campus scene, where there is no loop-closure in the whole course. But there exist fusions of multiple-layer road in the middle, therefore, occasionally we can see increase in time consumption.
    The \textbf{right column} shows the map update results of virtual garage scene, where there is no loop closure detected. Due to the repeated fusion of multiple-layer driving-way in the garage, we can observe an increase of time cost globally.}
    \label{fig:update_time}
\end{figure}

Our map representation can be applied to an online SLAM process, for which 
it is important to update the map instantly. Thus, we tested the time consumption on map update. There is no need to worry about the time to build a local map from the point cloud: the time-consuming of this part is significantly smaller than the LiDAR data interval, and the specific results are listed in the Table \ref{table:local-time}. In the implementation, the map is updated once per second. When there is no pose update, the update will be skipped. The frequency of the LeGO-LOAM back-end optimization for pose update is less than 1Hz. The code of the map update part is paralleled in eight threads. We record the time of map update for three scenes and get the results as shown in Figure \ref{fig:update_time}. 

\begin{table}[h]
\caption{Time consumption for building local map from point cloud data, tested on our campus dataset.}
\label{table:local-time}
\centering
\begin{tabular}{rrrrr}
\toprule
Resolution(m) & Range(m) & \begin{tabular}[c]{@{}l@{}}RANSAC detection\\ time consumption(s)\end{tabular} & \begin{tabular}[c]{@{}l@{}}Building a local map from detection\\ results time consumption(s)\end{tabular} \\
\midrule
0.1 & 20 & 0.0137 & 0.0325 \\
0.2 & 20 & 0.0139 & 0.0189 \\
0.1 & 40 & 0.0130 & 0.0595 \\
0.2 & 40 & 0.0140 & 0.0303 \\
\bottomrule
\end{tabular}
\end{table}

Specifically, for the three scenes that used for evaluating the map-update time cost, the first scene is part of our campus (the first row of Fig.\ref{fig:all3D}), to which our full online mapping procedure is applied, including front-end pose estimation, loop-closure detection, back-end optimization, traversable-region (curb) detection, 2D-OGM generation, and multiple-layer road map fusion. The second and third scenes are the virtual campus and garage scenarios, where poses between consecutive LiDAR frames are known beforehand. Thus, we do not need the front-end pose estimation, loop-closure detection and pose optimization steps.

We can observe that most map update can be finished within one second when point-cloud data within the radius of 20m are used for creating the local 2D-OGM, shown as the bottom row of Fig.\ref{fig:update_time}. In contrast, it generally takes more time to update map when using point-cloud data within the radius of 30m, shown as the top row of Fig.\ref{fig:update_time}, which makes many senses in that a larger local 2D-OGM needs more time to be incorporated into the global map.

The first column of Fig. \ref{fig:update_time} shows the results of running SLAM online and updating the map in our campus environment. It can be seen that there are two obvious peaks near the 141$^{th}$ and 181$^{th}$ updates. This is due to the fact that more local map poses are updated after finding two loop closures. The other results shown in Fig. \ref{fig:update_time} corresponds to the map update in the two virtual scenes, where relative poses between two LiDAR frames are given in the simulation environment and thus no significant time-cost variations are present. The last column shows the results of the simulated garage scene. It can be clearly seen that the update time increases mostly linearly as the number of map updates increases. This is because the garage itself covers a relative small area, but it has more floor layers. The number of local maps covered in the same 2D-OGM cell continues to increase, and the time used to update a cell is proportional to the number of local maps that need to be fused.




\subsection{Vehicle localization in the map}
\begin{figure*}
    \centering
    \subfloat[Apollo]{ \includegraphics[width=0.45\textwidth]{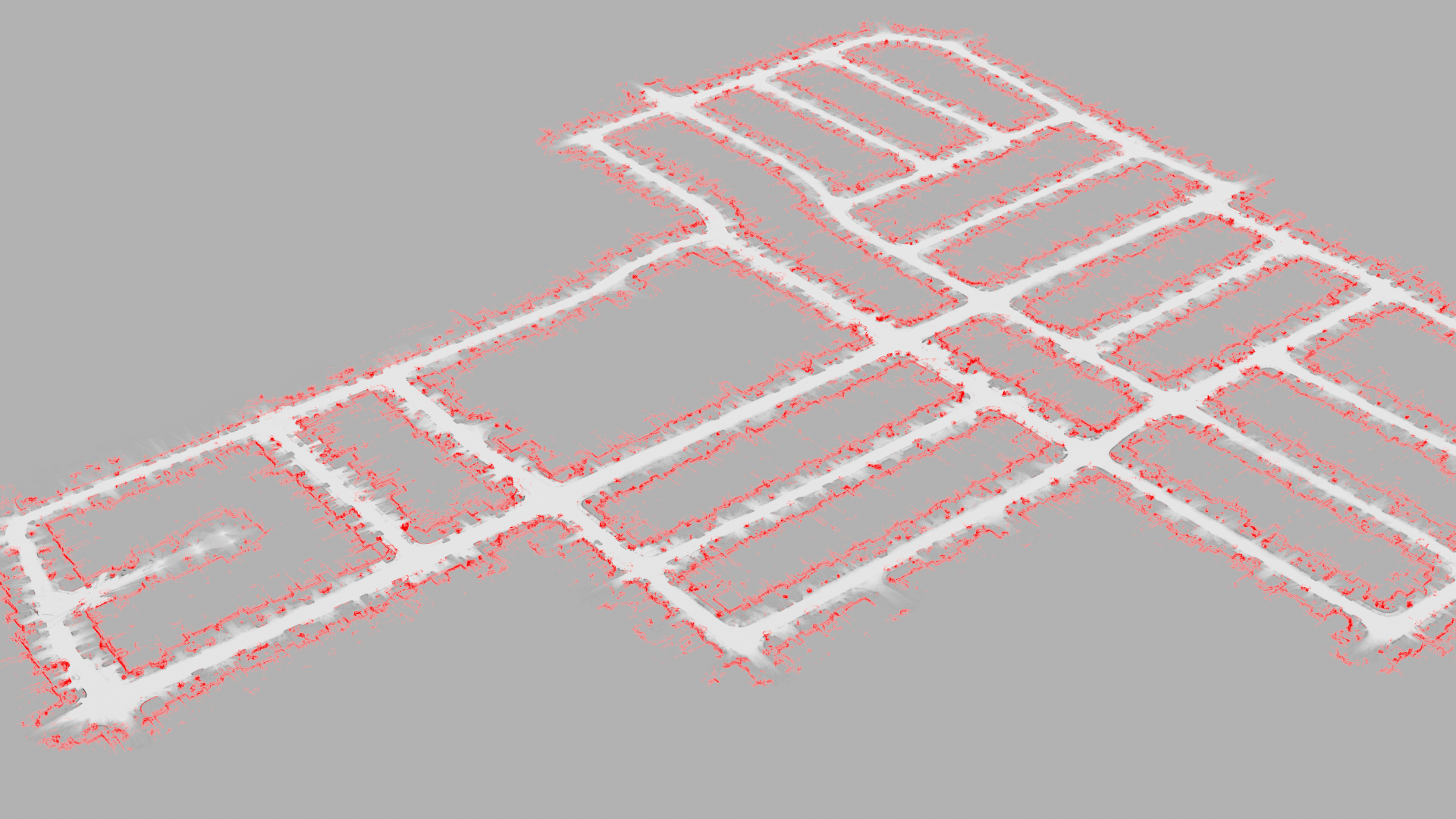} }
    \subfloat[MulRan]{ \includegraphics[width=0.45\textwidth]{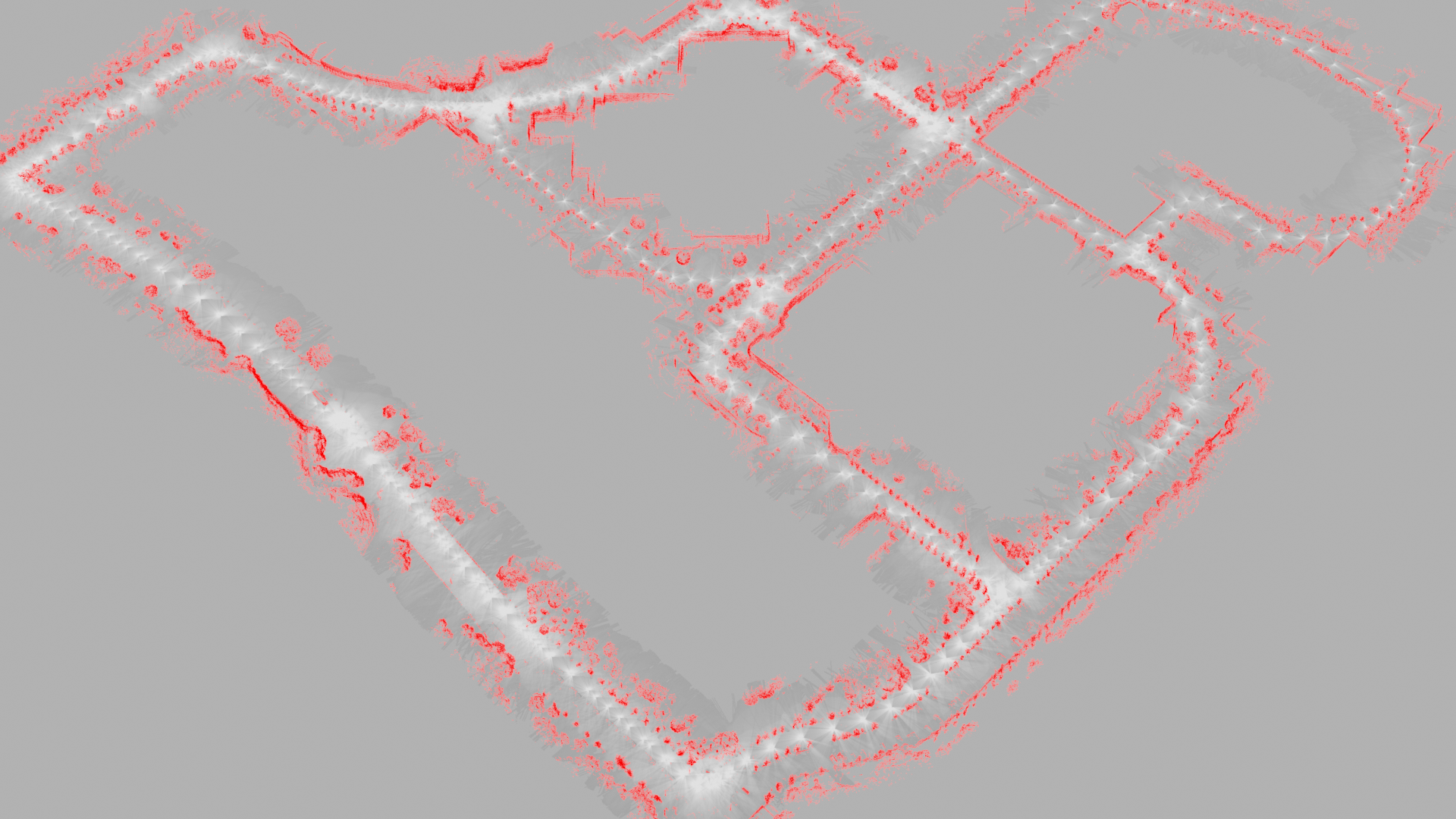} }    \caption{The created LiDAR Atlases from the Apollo and MulRan datasets based on our method, and they are used for localization when vehicle is running a second time in the maps.}
    \label{fig:locateScene}
\end{figure*}

Once our map is ready, we first test the performance of vehicle localization in it given the LiDAR point cloud data. 
Our test of localization is performed based on iterative closest point method (ICP) by directly matching current LiDAR point cloud with the obstacle point cloud which is decoded from our map representation. Since ICP is sensitive to the initial value, we use the localization result of the previous frame as the initial value of the new frame. At the same time, in order to accurately describe the localization error, we use the ground-truth pose in mapping to exclude the pose error of the map itself caused by SLAM in the results. The way of our localization experiment is similar to that in Monte Carlo localization methods. The Range-MCL\cite{chen2021icra} is a method with better experimental results in the class of Monte Carlo localization methods. It provides test results on multiple public datasets including Apollo\cite{ApolloDB} and MulRan\cite{kim2020mulran} in the paper, so we compare our results with the Range-MCL. We test on Apollo and MulRan datasets, which were collected when vehicles ran more than once in the same real scenes. Therefore, both datasets are suitable for testing localization performance given the map representation. We take the same trajectories as used in Range-MCL for ease of comparison. 
For the Apollo dataset, since the beginning of the test trajectory is not included in the map, it is difficult to locate it effectively. We start the test trajectory on this dataset from the 200th frame. For mapping data, since dense continuous frames are not necessary for mapping, we take one frame every 20 frames. Testing is done on every frame. For both mapping and test data, we do not directly use the original 64-channel data, but simulate 16-channel LiDAR data by taking one channel from every four channels to validate the performance of localization using a 16-channel LiDAR and our LiDAR road-atlas representation. The maps constructed from the mapping data of both datasets are shown in the Fig.\ref{fig:locateScene}, respectively. 

For the test results, we obtain the translation and rotation errors, and the average time cost for each frame. In Table \ref{table:compare}, we compare the results with those given in Range-MCL. In this test, our method adopts the following parameters: resolution=$0.2m$, radius=$40m$ (because some scenarios are relatively open and small radius can result in inaccurate positioning). The obstacle point (static vertical structures) takes the range of $-1m$ to $7m$ with respect to the sensor in vertical dimension, which is enough to cover most effective vertical structures. The number of segments in vertical dimension is $8$, and each segment uses 4 bits to represent the probability. This means that the length of the descriptor for encoding vertical structures in each cell is 32-bit. 

\begin{table}[h]
\caption{Comparison between our method and the Range-MCL \cite{chen2021icra} in terms of localization error and time cost.}
\label{table:compare}
\centering
\begin{tabular}{llrrr}
\toprule
         Dataset & method & \begin{tabular}[c]{@{}l@{}}translation\\ RMSE(m)\end{tabular} & \begin{tabular}[c]{@{}l@{}}rotation\\ RMSE(deg)\end{tabular} & \begin{tabular}[c]{@{}l@{}}average\\ time(s)\end{tabular} \\
\midrule
\begin{tabular}[c]{@{}l@{}}Apollo\end{tabular}  & our method & 0.26 & 1.07 & 0.1011 \\
                                                & Range-MCL  & 0.57 & 3.40 & \\
\midrule
\begin{tabular}[c]{@{}l@{}}MulRan\end{tabular}  & our method & 0.89 & 1.29 & 0.1058 \\
                                                & Range-MCL  & 0.83 & 3.14 & \\
\bottomrule
\end{tabular}
\end{table}

To demonstrate the effect of different parameters on localization accuracy, Table \ref{table:seg-effect} shows the effect of various number of segments in splitting the vertical dimension and Table \ref{table:range-effect} shows the effect of different LiDAR range radius and map resolutions. 
We observe that more segments can slightly improve the accuracy of localization, but at the same time, the time cost will increase due to the increase in the number of points.

In order to ensure that the occupied space does not increase significantly, our experiments simultaneously change the resolution and range while keeping the number of cells in the local map unchanged. Increasing the cell side length will reduce the number of points and increasing the range will bring a lesser increase in the number of points. 
It can be seen that the implementation shows that as these two increase, the time consumed decreases.

\begin{table}[h]
\caption{Effects of various number of vertical segments on localization}
\label{table:seg-effect}
\centering
\begin{tabular}{llrrr}
\toprule
         Dataset & segments & \begin{tabular}[c]{@{}l@{}}translation\\ RMSE(m)\end{tabular} & \begin{tabular}[c]{@{}l@{}}rotation\\ RMSE(deg)\end{tabular} & \begin{tabular}[c]{@{}l@{}}average\\ time(s)\end{tabular} \\
\midrule
\begin{tabular}[c]{@{}l@{}}Apollo\end{tabular}  & 2  & 0.288 & 1.106 & 0.0792 \\
                                                & 4  & 0.281 & 1.110 & 0.0867 \\
                                                & 8  & 0.265 & 1.073 & 0.1011 \\
                                                & 16 & 0.257 & 1.002 & 0.1305 \\
\bottomrule
\end{tabular}
\end{table}

\begin{table}[h]
\caption{Effects of different LiDAR range radius and map resolutions on localization}
\label{table:range-effect}
\centering
\begin{tabular}{llrrr}
\toprule
         Dataset & (resolution, range) & \begin{tabular}[c]{@{}l@{}}translation\\ RMSE(m)\end{tabular} & \begin{tabular}[c]{@{}l@{}}rotation\\ RMSE(deg)\end{tabular} & \begin{tabular}[c]{@{}l@{}}average\\ time(s)\end{tabular} \\
\midrule
\begin{tabular}[c]{@{}l@{}}Apollo\end{tabular}  & (0.2, 40) & 0.265 & 1.073 & 0.1011 \\
                                                & (0.3, 60) & 0.512 & 0.976 & 0.0908 \\
                                                & (0.4, 80) & 0.364 & 0.886 & 0.0817 \\
                                                & (0.2, 60) & 0.267 & 0.963 & 0.1506 \\
                                                & (0.4, 40) & 0.370 & 1.078 & 0.0807 \\
\midrule
\begin{tabular}[c]{@{}l@{}}MulRan\end{tabular}  & (0.2, 40) & 0.892 & 1.286 & 0.1058 \\
                                                & (0.3, 60) & 1.126 & 1.184 & 0.0946 \\
                                                & (0.4, 80) & 0.875 & 1.159 & 0.0844 \\
                                                & (0.2, 60) & 0.867 & 1.214 & 0.1330 \\
                                                & (0.4, 40) & 0.905 & 1.250 & 0.0699 \\
\bottomrule
\end{tabular}
\end{table}




\begin{figure}[h]
    \centering
    \includegraphics[width=0.7\textwidth]{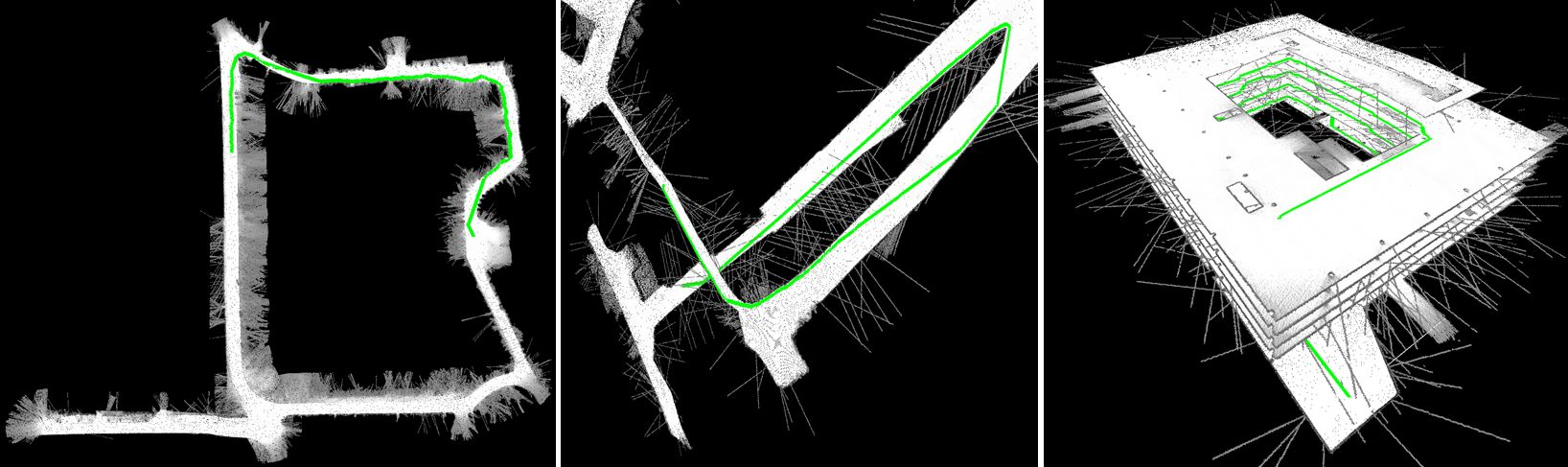}
    \caption{Three examples of path planning based on A* algorithm in the created 3D maps by our method.}
    \label{fig:path-planning}
\end{figure}

\subsection{Path planning in the map}
Since our map belongs to traversable- or road-based 3D map representation, where traversable- or road-region has been explicitly represented in the map. Therefore, another function that our map can provide is the path-planning capability. Given a start location and a destination in the map, a global path-planning algorithm can be applied to find a route for vehicle navigation given accurate localization. Figure \ref{fig:path-planning} shows three planned paths in three created road maps by our methods, respectively, where A* algorithm is applied to find the optimal path in each road map. It can be observed that the planned routes actually extend at different layer of the road map of the virtual campus and virtual garage scenes.

\section{Discussion}
The proposed map representation is built in an online mode by a LiDAR-based full SLAM framework. Combined with an exploratory planning algorithm, the proposed method can be used for autonomous robotic exploration in unknown environment. Certainly, the proposed method can also be used in an offline mode where the collected data are processed in a workstation to obtain the environment maps. In all, the created map representation can be scalable, extendable, and applicable in various ways. 

\textbf{Scalable}. 
Our map representation can be readily for scaling up whenever 
necessary to expand the current map size. Since the hard-drive storage size of our map representation is small and compact enough,
one preferable way to expand the map scale is to utilize the existing map that has been saved to hard-drive, and one only need to load the map file into memory, and to re-localize the current position of vehicle in the existing map, and do the same online mapping as this work. This way is better than continuously online building a large map in terms of the runtime memory usage.

\textbf{Extendable} 
Although we only use a single 16-channel LiDAR sensor to build the map, one can use LiDAR sensor with more channels, such as 32- or 64-channel LiDAR sensors. One advantage of our map representation when only using multiple-channel LiDAR is that the runtime memory consumption and hard-drive storage should approximately be invariant to the number of LiDAR's channels. 
Additionally, one can add heterogeneous information to the map by incorporating the other types of sensor data, such as 
visual features from cameras of different spectrum. To do that, one can estimate the external parameters between LiDAR sensor and camera before extracting visual feature along with corresponding point-cloud descriptor. Generally, richer information can be obtained from camera and stored in the map. Thus it can result in more accurate vehicle localization if vehicles are equipped with cameras as well.

\textbf{Applicable} 
Our map representation can also be used for detecting dynamic objects given the map, and also be useful for local motion planning. 
Because our map's hard-drive storage is small, it can be applied to distributed multiple-robot online mapping, and robots can share already saved maps with each other. 
Additionally, since our map representation is compactable in terms of the cell resolution, where robots can use a coarse-to-fine manner to exchange intermediate map results under bandwidth limitation whenever necessary. 

\section{Conclusion}

We propose a compactable and efficient 3D map representation for autonomous robot or vehicle navigation, called the LiDAR Road-Atlas. It can be generated by an online mapping framework  based on incrementally merging 2D local occupancy grid maps, where automatically labeled traversable regions, 
and 3D landmarks are encoded and embedded into the LiDAR Road-Atlas. 
Given the LiDAR Road-Atlas, one can achieve accurate vehicle localization in a coarse-to-fine manner, path planning in the labeled traversable regions and some other tasks.
We compare our map representation with a couple of popular map representation methods in robotics and autonomous driving societies, and our map representation is more favorable in terms of efficiency, scalablity and compactness. 

\section{Acknowledgement}
This work is supported by the Science
and Technology Development Fund of Macau SAR (File no. AGJ-2021-0046, SKL-IOTSC(UM)-2021-2023, and 0015/2019/AKP), Guangdong-Hong Kong-Macao Joint Laboratory of
Human-Machine Intelligence-Synergy Systems (No. 2019B121205007), and the startup project of Macau University (SRG2021-00022-IOTSC). 



%

\bibliographystyle{apalike}
\bibliography{reference}

\end{document}